\documentclass{article}
\PassOptionsToPackage{numbers, compress}{natbib}
\usepackage[preprint]{neurips_2026}
\usepackage[utf8]{inputenc} 
\usepackage[T1]{fontenc}    
\usepackage{hyperref}       
\usepackage{url}            
\usepackage{booktabs}       
\usepackage{amsfonts}       
\usepackage{nicefrac}       
\usepackage{microtype}      
\usepackage{xcolor}         
\usepackage{graphicx}       
\usepackage{amsmath}
\usepackage{multirow}
\usepackage{longtable}
\usepackage{subfigure}
\usepackage{caption}

\usepackage[most]{tcolorbox}

\usepackage{adjustbox}
\usepackage{colortbl}

\usepackage{tabularx,array,dcolumn}

\usepackage{algorithm}
\usepackage{algpseudocode}

\definecolor{PaperInk}{HTML}{222222}
\definecolor{PaperMuted}{HTML}{5F6670}
\definecolor{PaperRule}{HTML}{D8DDE3}
\definecolor{PaperBg}{HTML}{F9FAFB}
\definecolor{PaperAccent}{HTML}{3F5F73}
\definecolor{CautionAccent}{HTML}{8A4A4A}
\definecolor{CautionBg}{HTML}{FBF8F8}
\definecolor{jumpstartblue}{HTML}{17365D}

\newcommand{\takeawaybox}[1]{%
  \begin{tcolorbox}[
    enhanced,
    breakable,
    colback=PaperBg,
    colframe=PaperRule,
    coltext=PaperInk,
    boxrule=0.35pt,
    borderline west={2pt}{0pt}{PaperAccent},
    arc=0.7mm,
    outer arc=0.7mm,
    left=8pt,
    right=8pt,
    top=5pt,
    bottom=5pt,
    boxsep=0pt,
    before skip=0.75\baselineskip,
    after skip=0.55\baselineskip,
    fontupper={\fontsize{10pt}{12pt}\selectfont}
  ]
    \textcolor{PaperAccent}{\textbf{Takeaway.}}~#1
  \end{tcolorbox}%
}

\RequirePackage{libertinus}
\RequirePackage[scale=0.92]{tgheros}
\RequirePackage[libertine]{newtxmath}

\title{JumpStart Your Policy Learning with Lessons from 160,000 Training Runs}

\author{%
  \textbf{Nabil Omi}\textsuperscript{1}\thanks{Corresponding author: nabilomi@cs.uw.edu}
  \quad
  \textbf{Eric Bae}\textsuperscript{1}
  \quad
  \textbf{Chung Yik Edward Yeung}\textsuperscript{1}
  \quad
  \textbf{Siddhartha Sen}\textsuperscript{2}
  \quad
  \textbf{Ali Farhadi}\textsuperscript{1}\\
  \textsuperscript{1}University of Washington \quad
  \textsuperscript{2}Microsoft
}

\begin{document}

\maketitle

\begin{abstract}
    Reliable progress in offline policy learning depends on careful reporting, well-tuned baselines, and evaluation across diverse conditions. Prior work has shown that results can be sensitive to reporting choices, hyperparameter tuning, and dataset properties, but these sources of variability have not been systematically investigated together at the scale needed to understand how they shape conclusions. To address this gap, we present a large-scale empirical study of offline reinforcement and imitation learning, training over 160,000 policies across 114 datasets. At this scale, no algorithm dominates: aggregate performance among the strongest methods is often close, but the leaders differ substantially across environments. We find that proper hyperparameter tuning frequently reshuffles perceived algorithm rankings and that benchmark composition can produce conflicting conclusions. We also study hyperparameter sensitivity and transfer across environments, identifying a simple strategy for deriving strong default configurations. We use our findings to develop a dataset-conditioned recommender that provides task-specific algorithm recommendations for practitioners. Finally, we release JumpStart: a resource suite containing every trained policy, per-model scores and hyperparameters, strong baselines across all environments, training and evaluation code, and an extensible website for retrieving, analyzing, and contributing results. Together, these resources aim to make offline policy-learning research more reliable and enable future work beyond the scope of this study.
\end{abstract}

\begin{center}
    \bfseries\color{jumpstartblue}
    \href{https://github.com/RAIVNLab/jumpstart}{Code}
    \enspace\textbullet\enspace
    \href{https://jumpstart.cs.washington.edu/models}{Models and Data}
    \enspace\textbullet\enspace
    \href{https://jumpstart.cs.washington.edu/}{Website}
\end{center}
\vspace{-0.5em}

\begin{figure}[h]
    \centering
    \includegraphics[width=\linewidth]{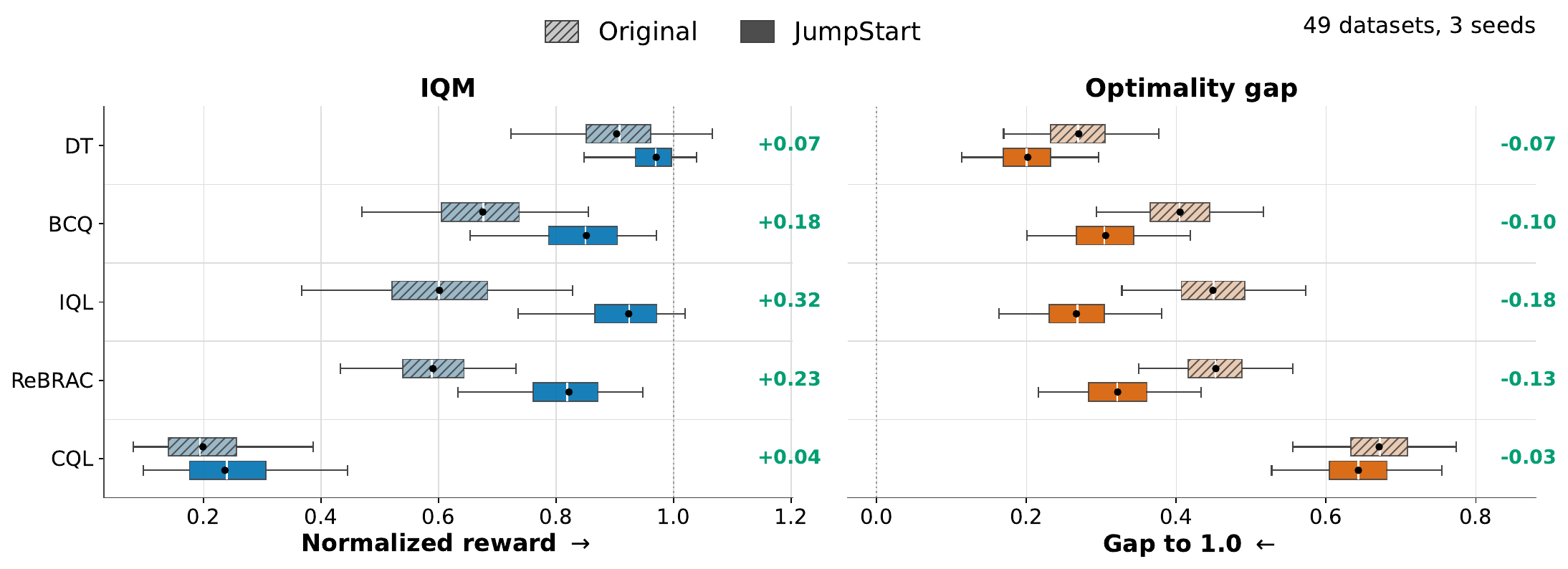}
    \caption{Our tuned models substantially outperform published baselines on Minari and change the relative ordering of algorithms. Original paper implementations use the authors' recommended or closest available hyperparameters. Further details are provided in Appendix~\ref{app:og-baselines}.}
    \label{fig:header-figure}
\end{figure}
\newpage

\section{Introduction}
\label{sec:introduction}

Learning policies from fixed datasets offers an attractive alternative when online interaction is expensive, slow, or unsafe. Offline Reinforcement Learning (ORL) uses reward-labeled data to optimize a policy without further interaction, while Imitation Learning (IL) learns to reproduce behavior from demonstrations. We use the term \emph{offline policy learning} to encompass both settings, since they share the central empirical challenge studied here: reliably training and comparing policies using pre-collected data.

This comparison is harder than standard benchmark results often suggest. Performance can change substantially with the choice of dataset, environment, hyperparameters, evaluation protocol, and aggregation rule. When these factors vary across studies, it becomes difficult to determine whether an apparent improvement comes from the algorithm itself or from the comparison surrounding it. Small evaluation suites and limited tuning budgets make these sensitivities particularly difficult to see.

Prior work has exposed individual parts of this problem. Offline RL performance can depend strongly on reward normalization, initialization, policy parameterization, network architecture, learning-rate schedules, and other implementation choices \citep{kang2023improving}. Hidden layer size \citep{cetin2024simple}, dataset coverage and trajectory quality \citep{schweighofer2021dataset}, reward sparsity, horizon, and stochasticity \citep{bhargava2023when}, and the use of human rather than synthetic demonstrations \citep{mandlekar2021matterslearningofflinehuman} can all change which methods appear strongest. Aggregation and uncertainty reporting introduce further ambiguity \citep{agarwal2021deep}. Cross-paper comparisons compound these problems: studies use different environment subsets, evaluation budgets, and benchmark versions, and numerical results are sometimes transferred between settings where they are not directly comparable, such as different environment versions \citep{kostrikov2021offline}.

We present a large-scale empirical study designed to make these sensitivities visible. Using nearly one million GPU-hours, we train more than $160{,}000$ policies across 114 datasets spanning continuous control, robotics, navigation, and Atari. Figure~\ref{fig:header-figure} shows that our tuned models not only substantially outperform reference runs constructed from the original implementations and published hyperparameters, but also change the relative ordering of the algorithms. Across the 49 shared datasets, IQL overtakes BCQ under both IQM and optimality gap after tuning. Hyperparameter selection can therefore change the conclusion of a comparison, not simply the absolute scores it reports. The reference implementations are modified only as needed to load the standardized datasets we use, we otherwise follow the authors' tuning guidance or use the configuration from the most similar reported dataset. Full matching and tuning details are provided in Appendix~\ref{app:og-baselines}. We use the resulting corpus to study how tuning, benchmark composition, aggregation, and algorithm design affect empirical conclusions. 

Our primary contributions are:

\begin{itemize}
    \item We conduct an extensively tuned comparison of offline policy-learning algorithms across 114 datasets, producing more than $160{,}000$ trained policies and a controlled view of performance across continuous- and discrete-control domains.
    \item We characterize how hyperparameter choice, tuning budget, and random seed affect absolute performance and algorithm rankings, identify the most influential hyperparameters and practical tuning budgets, and derive strong default configurations that transfer across datasets.
    \item We show that benchmark composition can produce conflicting algorithm rankings even when the selected subsets are comparable in size to established evaluation suites, and we propose reporting practices that make this sensitivity visible.
    \item We systematically analyze how dataset characteristics relate to algorithm performance and use these relationships to recommend promising algorithms for new tasks.
    \item We release the JumpStart corpus, including trained policies, hyperparameters, training code, and a website for retrieving, comparing, analyzing, and extending results.
\end{itemize}

\section{Background and Related Work}
\label{sec:background}

    In this section, we briefly review the background needed to place our study in context. We first outline the main algorithm families we study, then summarize the datasets and benchmarks used in offline policy learning, and finally discuss prior empirical studies on what makes algorithms work or fail, along with related issues in evaluation, reporting, and tuning.

    \textbf{Algorithms.} We study a mix of popular imitation learning and offline RL methods. Behavior Cloning (BC) fits actions directly through supervised learning. Action Chunking with Transformers (ACT; \citealp{zhao2023learning}) predicts short action sequences to reduce compounding error in manipulation settings. VQ-BeT~\citep{lee2024vqbet} uses vector-quantized latent actions to capture multimodal behavior. Diffusion Policy~\citep{chi2023diffusion} models action sequences through conditional denoising, allowing it to represent multimodal action distributions and perform receding-horizon control. Decision Transformer (DT; \citealp{chen2021decision}) models trajectories autoregressively and conditions its predictions on a desired return-to-go. Implicit Q-Learning (IQL; \citealp{kostrikov2021offline}) improves a policy through implicit value estimation without querying out-of-distribution actions. Conservative Q-Learning (CQL; \citealp{kumar2020conservative}) regularizes value estimates conservatively to stabilize offline Q-learning. ReBRAC~\citep{tarasov2023revisiting} builds on TD3+BC with a set of specific design choices, including layer normalization, a deeper critic, a higher discount factor, and decoupled actor and critic penalties. Batch-Constrained deep Q-learning (BCQ; \citealp{fujimoto2019offpolicydeepreinforcementlearning}) constrains candidate actions to remain close to the support of the dataset.

    \textbf{Datasets.} The data used for offline policy learning can vary widely in quality, coverage, and domain, and these properties can strongly influence which algorithms perform well. The most widely used benchmark is D4RL~\citep{fu2021d4rldatasetsdeepdatadriven}, which introduced a standardized suite for evaluating offline, or data-driven, reinforcement learning. The original suite includes Maze2D and AntMaze~\citep{fu2021d4rldatasetsdeepdatadriven}, Gym-MuJoCo~\citep{brockman2016openaigym,todorov2012mujoco}, Adroit~\citep{rajeswaran2018learning}, FrankaKitchen~\citep{gupta2019relay}, Flow~\citep{wu2017flow}, and CARLA~\citep{dosovitskiy2017carla}. In practice, most subsequent evaluations focus on the first five groups and omit Flow and CARLA.

    Prior to D4RL, studies generally generated their own non-standardized evaluation datasets~\citep{fujimoto2019offpolicydeepreinforcementlearning,kumar2019stabilizingoffpolicyqlearningbootstrapping,wu2019behaviorregularizedofflinereinforcement}. These datasets were commonly produced by training an expert through online reinforcement learning and then retaining either its training replay buffer or trajectories generated by selected checkpoints. Later, \citet{mandlekar2021matterslearningofflinehuman} introduced RoboMimic, a collection of human teleoperation trajectories for the RoboSuite framework~\citep{zhu2020robosuite}. They found that conclusions obtained from machine-generated datasets did not necessarily transfer to human demonstrations, a distinction that we examine more broadly in Section~\ref{subsec:aggregation}.

    More recent benchmarks have aimed to address limitations of D4RL. \citet{lu2023challenges} introduce V-D4RL, which provides pixel-based analogues of D4RL tasks for evaluating offline RL from visual observations. \citet{liu2023libero} release LIBERO, a procedurally generated suite of 130 language-conditioned manipulation tasks built around RoboSuite and designed to study lifelong knowledge transfer in imitation learning. \citet{rafailov2024d5rl} propose D5RL as a successor to D4RL, motivated by saturation on the original tasks, with more realistic manipulation and locomotion environments and heterogeneous data sources spanning scripted, teleoperated, and play-style data. Finally, \citet{park2025ogbench} introduce OGBench, an offline goal-conditioned RL benchmark spanning eight environment types and 85 datasets designed to probe capabilities such as stitching, long-horizon reasoning, and robustness to stochasticity.

    In parallel, the Farama Foundation released Minari~\citep{minari}, a standardized dataset format and Python API that hosts reimplementations of D4RL and other reference offline RL datasets. Minari has partially supplanted the original D4RL codebase, which is no longer actively maintained. For reproducibility and ease of maintenance, as discussed further in Section~\ref{subsec:datasets}, we use a 114-dataset subset of Minari. These datasets were collected from several sources, including humans, RL agents, and scripted planners. This diversity allows us to study not only transfer between human- and machine-generated data, but transfer across expert sources more generally in Section~\ref{subsec:aggregation}.

    \textbf{Reporting practices and reproducibility.}\label{subsec:reporting-practices} Inconsistent reporting has long been an issue in deep RL. \citet{agarwal2021deep} recommend robust aggregate metrics accompanied by uncertainty estimates. We adopt two such metrics throughout: the interquartile mean (IQM), which discards the top and bottom $25\%$ of normalized scores before averaging, and the optimality gap, which measures the average shortfall from a target performance level; we use expert performance as the target.

    Canonical offline policy-learning papers differ in ways that make direct comparison difficult. For example, BCQ~\citep{fujimoto2019offpolicydeepreinforcementlearning}, BEAR~\citep{kumar2019stabilizingoffpolicyqlearningbootstrapping}, and BRAC~\citep{wu2019behaviorregularizedofflinereinforcement} primarily present reward-over-training curves rather than consolidated benchmark tables, while Decision Transformer~\citep{chen2021decision} reports results on a comparatively restricted subset of D4RL. In Section~\ref{sec:aggregate-evals}, we show that benchmark subsets comparable in size to established evaluation suites can nevertheless produce sharply different algorithm rankings. Results are also sometimes copied across papers despite differences in the underlying environment versions, such as v0 versus v2, as noted by \citet{kostrikov2021offline}.

    \citet{tarasov2023revisiting} address part of this problem by providing single-file implementations of several algorithms and evaluating them on a unified set of D4RL dataset versions. Their primary goal, however, is implementation standardization rather than extensive per-dataset tuning. Our results show why this distinction matters: tuning can substantially change both absolute performance and the relative ordering of algorithms. For example, BC frequently appears near the bottom of their leaderboards, whereas after tuning we find it substantially more competitive.

    \textbf{Empirical studies.} \citet{andrychowicz2020mattersonpolicyreinforcementlearning} train tens of thousands of on-policy RL agents while varying more than 50 implementation choices, finding that low-level details such as policy initialization, observation normalization, and action-distribution parameterization can matter as much as the high-level algorithm. \citet{orsini2021mattersadversarialimitationlearning} reach a similar conclusion for adversarial imitation learning and additionally show that algorithm rankings shift substantially between synthetic and human demonstrations.

    \citet{paine2020hyperparameterselectionofflinereinforcement} conduct an early study of hyperparameter selection in offline RL. Across three algorithms and ten continuous-control tasks, they train 2,560 policies and find that performance varies substantially with hyperparameter choice. They also show that Fitted Q Evaluation can often rank the resulting policies without additional environment interaction. This addresses a different problem from ours: their policies have already been trained, and the goal is to select a strong candidate when the budget for evaluating those policies is limited.

    This setup treats training many candidate policies as feasible while evaluation is the primary bottleneck. Although this remains relevant when deployment is expensive or risky, the balance is less clear for modern policy learning, where models may contain billions of parameters and be trained on large datasets~\citep{black2026pi0visionlanguageactionflowmodel}. Our goal is instead to compare algorithms after adequate tuning while reducing the cost of reaching that point. Across 162,864 policies from ten algorithms and 114 datasets, we identify which hyperparameters warrant attention and how their importance changes across datasets (Section~\ref{subsec:sensitivity}). We then use the same sweeps to derive stronger default parameters, improving the performance of an initial run and reducing the amount of target-specific tuning required (Section~\ref{subsec:transfer}).

    \citet{kang2023improving} ablate 20 low-level implementation choices across CQL, CRR, and IQL on D4RL and RL Unplugged. They show that many reported gains arise from these choices rather than the central algorithm and that their effects depend strongly on the data distribution. Some of their recommendations do not transfer uniformly across our broader suite; for example, we find that the importance of learning rate varies substantially across datasets. \citet{cetin2024simple} similarly argue that scale can matter more than novelty in offline RL, showing that appropriately scaled simple methods can match or outperform more elaborate algorithms on D4RL. To control for model scale, we either include larger hidden sizes in our search spaces or use networks comparable in scale to the larger configurations in their study while varying critic count and layer width.

    Dataset properties provide another explanation for inconsistent results. \citet{schweighofer2021dataset} find that exploration coverage and trajectory quality strongly predict which methods perform well, while \citet{bhargava2023when} show that the relative ordering of offline RL and imitation-style methods depends on properties such as reward sparsity, horizon, and demonstration source. We extend this line of work by automatically parsing environment and dataset descriptions into structured characteristics and using them in a practitioner-facing recommender that identifies promising algorithms for a given task. Prior studies therefore establish that implementations, hyperparameters, model scale, and dataset properties can all affect performance; our study brings these factors together and makes the resulting relationships directly actionable.

    Our study brings these lines of analysis together across ten algorithms, 114 datasets, continuous- and discrete-control domains, and more than $160{,}000$ trained policies. Because we extensively tune each algorithm on each dataset, we can analyze dataset effects without relying on potentially weak default configurations. In Section~\ref{sec:recommendation-system}, we study these effects systematically and use environment and dataset characteristics to recommend promising algorithms for a given task. We additionally release the complete corpus of trained policies, hyperparameters, and training code so that future studies can reproduce, inspect, and extend our analysis. Table~\ref{tab:related-work-comparison} summarizes the components combined in our study relative to the most closely related empirical work.

    \begin{table*}[t]
        \centering
        \caption{Comparison with selected large-scale empirical studies in offline policy learning. Prior-work counts describe the principal offline benchmarks; JumpStart counts include the full hyperparameter-tuning corpus.}
        \label{tab:related-work-comparison}
        \scriptsize
        \setlength{\tabcolsep}{3pt}
        \renewcommand{\arraystretch}{1.08}
        \resizebox{\textwidth}{!}{
            \begin{tabular}{@{}lrrrcccccc@{}}
                \toprule
                Study &
                \shortstack{Datasets\\or tasks} &
                \shortstack{Methods /\\variants} &
                \shortstack{Policies\\counted} &
                Continuous &
                Discrete &
                \shortstack{Trained\\models} &
                \shortstack{Policy\\metadata} &
                \shortstack{Training\\code} &
                \shortstack{Extensible\\leaderboard} \\
                \midrule
                \citet{agarwal2020optimistic} (Atari)
                    & 60 games & 6 & $1{,}800^\dagger$
                    & -- & \checkmark & -- & -- & \checkmark & -- \\
    
                \citet{paine2020hyperparameterselectionofflinereinforcement}
                    & 10 tasks & 3 & $2{,}560^\dagger$
                    & \checkmark & -- & -- & -- & -- & -- \\
    
                \citet{mandlekar2021matterslearningofflinehuman}
                    & 8 tasks & 6 & NR
                    & \checkmark & -- & $\triangle$ & $\triangle$ & \checkmark & -- \\
    
                \citet{schweighofer2021dataset}
                    & 150 (6 envs.) & 9 & 6,750
                    & -- & \checkmark & -- & -- & \checkmark & -- \\
    
                \citet{d3rlpy}
                    & 17 datasets/tasks & 12 & $1{,}400^\dagger$
                    & \checkmark & \checkmark & -- & \checkmark & \checkmark & -- \\
    
                \citet{tarasov2023revisiting}
                    & 30 datasets & 10 & $1{,}200^\dagger$
                    & \checkmark & -- & -- & \checkmark & \checkmark & -- \\
    
                \citet{kang2023improving}
                    & 26 tasks & 9 & $702^\dagger$
                    & \checkmark & -- & -- & -- & \checkmark & -- \\
    
                \citet{bhargava2023when}
                    & 3 principal suites & 3 & NR
                    & \checkmark & \checkmark & -- & -- & \checkmark & -- \\
    
                \midrule
                \textbf{JumpStart (ours)}
                    & 114 & 10 & 162,864
                    & \checkmark & \checkmark & \checkmark & \checkmark & \checkmark & \checkmark \\
                \bottomrule
            \end{tabular}
        }
    
        \vspace{2pt}
        \parbox{\textwidth}{\scriptsize
            \textit{Notes.}
            NR denotes an unreported total; $\triangle$ denotes a partial release.
            Method counts include separately evaluated variants.
            $^\dagger$Counts are reconstructed from reported experimental settings
            and exclude auxiliary ablations and offline-to-online fine-tuning.
            Policy metadata comprises policy-level scores paired with hyperparameters
            or experiment configurations.
            An extensible leaderboard accepts externally contributed results.
        }
    \end{table*}

\section{Training Methods}
\label{sec:methods}

    \subsection{Datasets}
    \label{subsec:datasets}
 
        \begin{figure}
            \centering
            \includegraphics[width=0.8\linewidth]{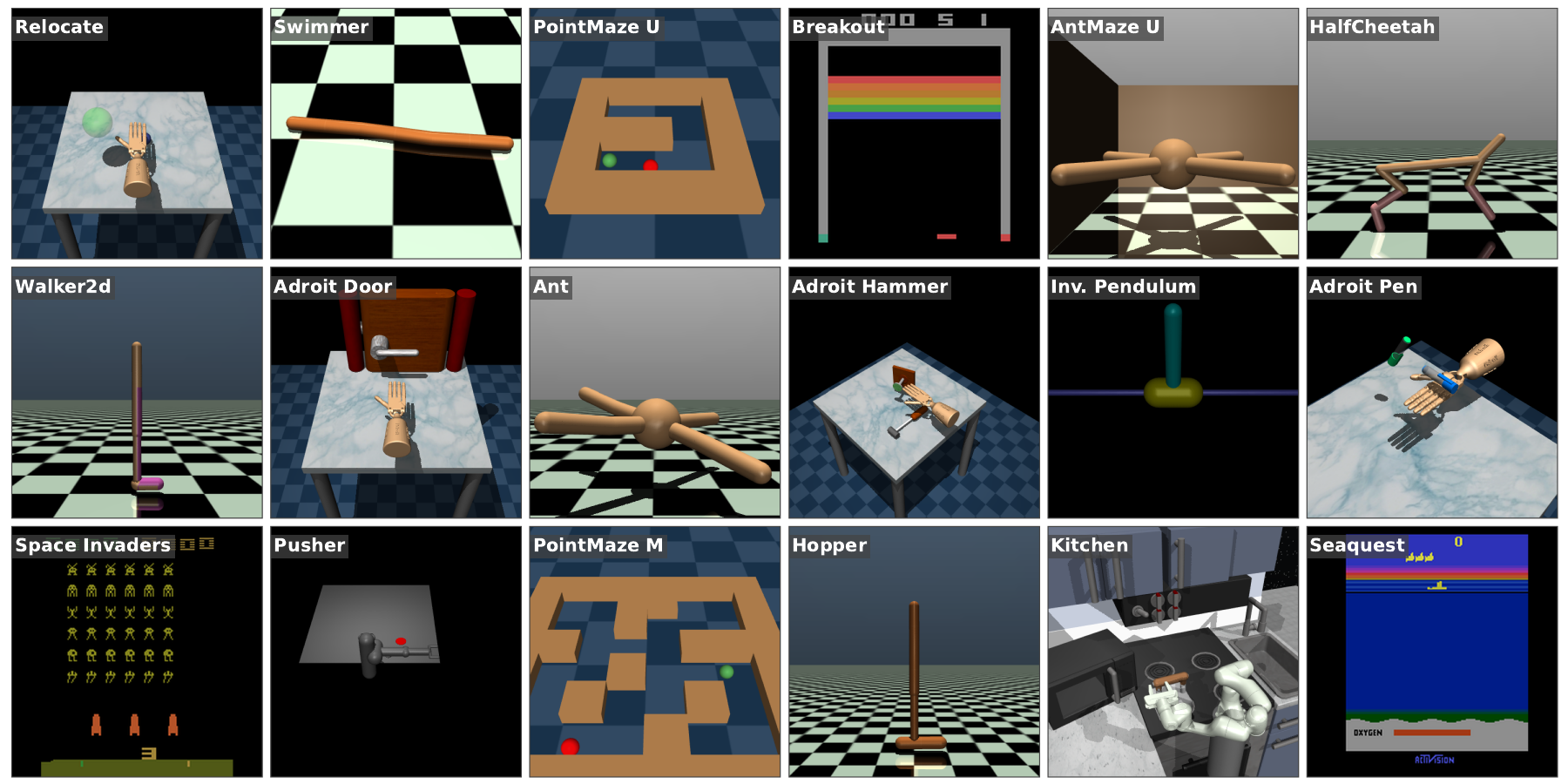}
            \caption{An 18-environment subset of the diverse tasks used in the study. A full list of environments and datasets can be found in Appendix~\ref{app:datasets}.}
            \label{fig:tasks-sample}
        \end{figure}
    
        Minari follows the same recipe as D4RL, generating offline datasets from trained policies~\citep{minari}, and additionally includes 7 human-collected datasets. We use Minari rather than the original D4RL release because it is the actively maintained successor, with newer environment versions, clearer documentation, and openly available dataset-generation code. Because the underlying datasets are not bit-identical to D4RL, some absolute numbers (e.g., CQL on AntMaze) may differ from prior reports. A subset of the tasks used in our study is visualized in Figure~\ref{fig:tasks-sample}. For Figure~\ref{fig:header-figure}, we  adapted the source code of prior work \citep{chen2021decision,tarasov2023revisiting,fujimoto2019offpolicydeepreinforcementlearning,kostrikov2021offline,kumar2020conservative} to support Minari and adapted hyper-parameters using the procedure from~\citet{tarasov2023revisiting} (additionally following tuning suggestions from their papers, if available), and re-evaluated their models against ours. 
        
        We made two additions. The Atari datasets in Minari were too small for the discrete-control comparisons we wanted, so we collected new ones with $100$k states per game using their dataset generation code, following the Atari$100$k setting~\citep{kaiser2019modelbased}. We also omit a number of environments whose default datasets are trivially solvable across all methods we tested. The remaining $114$ datasets ($57$ continuous control, $57$ discrete) cover a wide range of distributional shapes, including some near-zero-reward degenerate datasets that we filter from our analyses, though they are still included in artifact releases for future studies. After filters, we land at $52$ discrete and $57$ continuous datasets. We provide the full set of datasets and their descriptions in Appendix~\ref{app:datasets}.

        D4RL provides fixed expert and random reference returns for each environment, whereas Minari currently provides these values only for datasets imported from D4RL. In our work, we define "expert" as the policy that generated the dataset used for offline policy learning. To apply a consistent normalization procedure across all environments, we estimate the expert reference directly from each dataset and estimate the random reference by running a random policy for 100 trajectories in each unique environment.

        For a trajectory $\tau$, let
        \[
            R(\tau) = \sum_{t=0}^{T_\tau} r_t
        \]
        denote its undiscounted return. For dataset $\mathcal{D}$ from environment $e$, we define
        \[
            R_{\mathrm{expert}}(\mathcal{D})
            =
            \frac{1}{|\mathcal{D}|}
            \sum_{\tau \in \mathcal{D}} R(\tau),
            \qquad
            R_{\mathrm{random}}(e)
            =
            \frac{1}{100}
            \sum_{i=1}^{100} R(\tau_i^{\mathrm{rand}}).
        \]
        Given an evaluation return $R_{\mathrm{eval}}$, we report
        \[
            R_{\mathrm{norm}}
            =
            \frac{
                R_{\mathrm{eval}} - R_{\mathrm{random}}(e)
            }{
                R_{\mathrm{expert}}(\mathcal{D}) - R_{\mathrm{random}}(e)
            }.
        \]

    \subsection{Model training}
    \label{subsec:model-training}
    
        Our pipeline is built on d3rlpy~\citep{d3rlpy} for verified algorithm implementations and Ray Tune~\citep{liaw2018tuneresearchplatformdistributed} for distributed hyperparameter tuning. We completed full hyperparameter tuning runs with BC, BCP (BC with percentile-based data filtering, also called BC\%, a common baseline~\citep{tarasov2023revisiting,chen2021decision}), BCQ, CQL, and DT across all $114$ continuous and discrete datasets; VQ-BeT, ACT, IQL, and ReBRAC run on $57$ continuous control datasets. We start with the LeRobot\footnote{\url{https://github.com/huggingface/lerobot/tree/main/src/lerobot/policies/vqbet}, \url{https://github.com/huggingface/lerobot/tree/main/src/lerobot/policies/act}} implementations for ACT and VQ-BeT and modify them to work with our codebase. For Diffusion Policy, we used a modified implementation from the original paper's code release\footnote{\url{https://github.com/real-stanford/diffusion_policy}}. We validated the modified implementations on a subset of MuJoCo tasks in Minari by converting datasets to a format supported by the original code.
        
        \textbf{Training budgets.} Value-based methods (CQL, BCQ, IQL, ReBRAC) use $500{,}000$ gradient steps on continuous environments and $250{,}000$ on discrete. BC, BCP, and DT use $50$ epochs. Diffusion and ACT uses $100$ epochs, and VQ-BeT uses 300 total split among the stages. These budgets were chosen via an exponentially increasing sweep on a small set of testing datasets ($3$ MuJoCo HalfCheetah datasets and D4RL AntMaze) under default and four random hyperparameter configurations, picking the smallest budget at which rewards plateau. 

        \begin{table}[t]
            \centering
            \caption{Number of trained policies by algorithm and action space.}
            \label{tab:policy-counts}
            \footnotesize
            \setlength{\tabcolsep}{3pt}
            \begin{tabular}{@{}lrrrrrrrrrrr@{}}
                \toprule
                & BC & BCP & BCQ & CQL & DT & IQL & ReBRAC & ACT & VQ-BeT & Diffusion & Total \\
                \midrule
                Continuous & 11,400 & 11,400 & 11,400 & 11,400 & 11,400 & 11,302 & 11,401 & 10,830 & 8,833 & 10,590 & 109,956 \\
                Discrete   & 11,400 & 11,400 & 8,687  & 10,021 & 11,400 & -- & -- & -- & -- & -- & 52,908 \\
                \midrule
                \textbf{Total} & \textbf{22,800} & \textbf{22,800} & \textbf{20,087} & \textbf{21,421} & \textbf{22,800} & \textbf{11,302} & \textbf{11,401} & \textbf{10,830} & \textbf{8,833} & \textbf{10,590} & \textbf{162,864} \\
                \bottomrule
            \end{tabular}
        \end{table}
        
        \begin{table}[t]
            \centering
            \caption{Search-space boundary occupancy. We report the percentage of trials containing at least one hyperparameter within 5\% of either search-range boundary. Enrichment is the difference between the top 10\% of trials and all trials.}
            \label{tab:hp-boundaries}
            \small
            \begin{tabular}{@{}lrrr@{}}
                \toprule
                Algorithm & Top 10\% & All trials & Enrichment \\
                \midrule
                ACT    & 32.8\% & 35.3\% & $-2.5$ pp \\
                BC     & 33.5\% & 33.9\% & $-0.4$ pp \\
                BCP    & 41.6\% & 40.3\% & $+1.3$ pp \\
                BCQ    & 85.4\% & 85.2\% & $+0.2$ pp \\
                CQL    & 83.3\% & 78.5\% & $+4.8$ pp \\
                DT     & 70.1\% & 69.9\% & $+0.1$ pp \\
                IQL    & 71.2\% & 71.1\% & $+0.1$ pp \\
                ReBRAC & 74.6\% & 71.6\% & $+3.1$ pp \\
                VQ-BeT & 62.9\% & 64.6\% & $-1.7$ pp \\
                \bottomrule
            \end{tabular}
        \end{table}
        
        \textbf{Hyperparameter tuning.} We use random search~\citep{bergstra2012random} and the Tree-structured Parzen Estimator (TPE), which models better- and worse-performing configurations to bias subsequent trials toward promising regions~\citep{bergstra2011algorithms}. We run TPE through Ray Tune's \texttt{OptunaSearch} wrapper~\citep{akiba2019optuna}. The search spaces are reported in Tables~\ref{tab:hp-ranges-continuous} and~\ref{tab:hp-ranges-discrete}, and model counts are reported in Table~\ref{tab:policy-counts}. To test whether these ranges constrain performance, we compare how often the top 10\% of trials and the full sweep contain at least one hyperparameter within 5\% of a search-range boundary. If there is a significant difference in prevalence within the boundaries, the sweep may have been limited in width. As shown in Table~\ref{tab:hp-boundaries}, the median enrichment among top trials is only $+0.1$ percentage points, with no consistent increase across algorithms. Strong configurations therefore do not systematically accumulate near the edges of our search spaces, suggesting that the selected ranges are sufficiently broad.

        \textbf{Evaluation.} For all evaluations, we use the final checkpoint evaluated across 100 episodes. Minari provides a method to recover the environment used to collect each dataset, but sometimes omits wrappers (e.g., resizing or frame stacking), as in Atari. In these cases, we recover the missing wrappers from Minari's data-generation scripts.\footnote{\url{https://github.com/Farama-Foundation/minari-dataset-generation-scripts/tree/main}}

\section{Evaluation Sensitivity}
\label{sec:aggregate-evals}

    Aggregates can hide more than they show. The apparent best algorithm depends materially on the choice of \textbf{metric}, \textbf{data}, and \textbf{tuning budget}. We treat each as a distinct failure mode and recommend evaluation practices that make the resulting ambiguity legible.

    \begin{figure}
        \centering
        \includegraphics[width=0.6\linewidth]{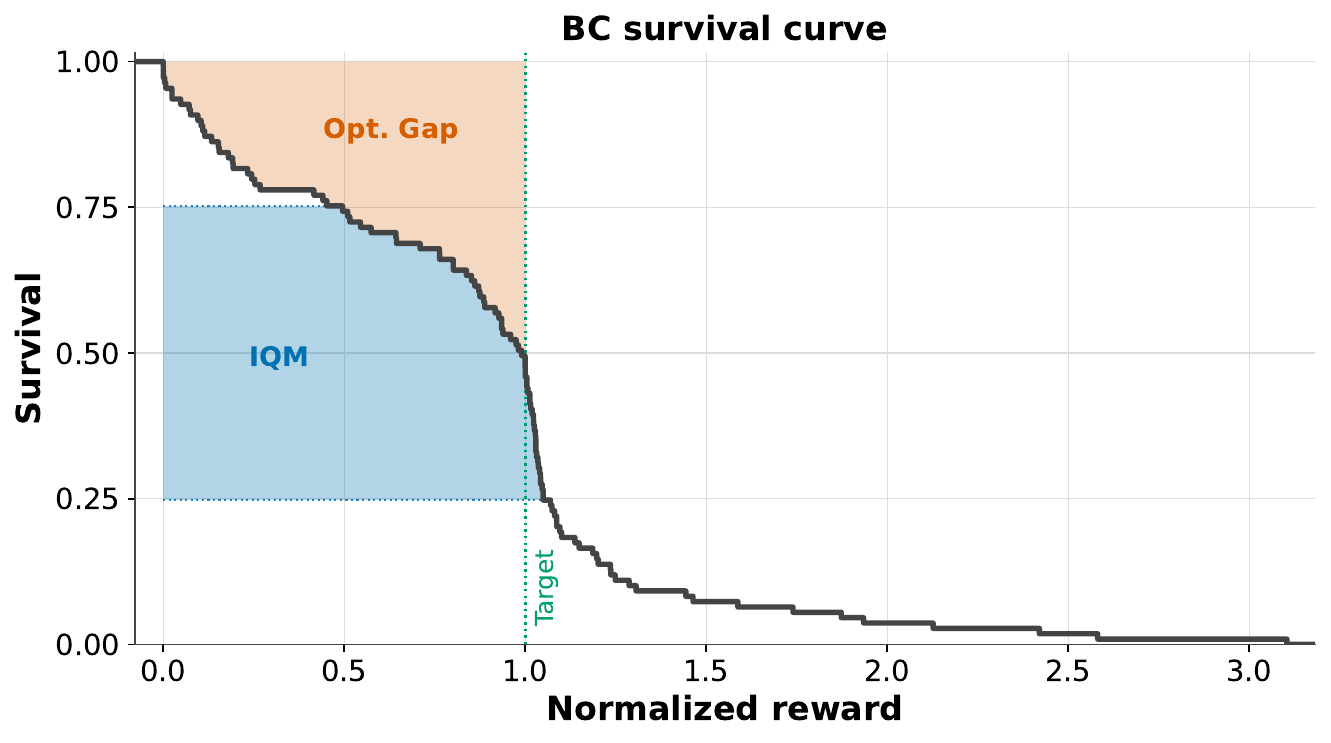}
        \caption{A visual example of optimality gap and the IQM overlaid on a survival chart for behavior cloning on our suite.}
        \label{fig:opt-gap-iqm}
    \end{figure}
    
    The choice of evaluation metric has been studied extensively in reinforcement learning. Following \citet{agarwal2021deep}, we avoid reporting only means or medians over many runs and datasets, and instead use the interquartile mean (IQM) and optimality gap as robust aggregate metrics. For normalized scores $x_1,\ldots,x_n$, let $q_{0.25}$ and $q_{0.75}$ denote the empirical first and third quartiles. The IQM is
    \[
        \operatorname{IQM}(x)
        =
        \frac{1}{|\mathcal{I}|}
        \sum_{i \in \mathcal{I}} x_i,
        \qquad
        \mathcal{I}
        =
        \{i : q_{0.25} \leq x_i \leq q_{0.75}\}.
    \]
    For an optimality threshold $\gamma$, the optimality gap is
    \[
        \operatorname{OG}_{\gamma}(x)
        =
        \frac{1}{n}
        \sum_{i=1}^{n}
        \max(\gamma - x_i, 0).
    \]
    In our setting, normalized expert performance corresponds to $\gamma=1$. Since metric aggregation is already well explored, we adopt these methodologies and focus our analysis on the remaining sources of uncertainty. We provide a visual explanation of IQM and optimality gap in Figure~\ref{fig:opt-gap-iqm}.

    \begin{figure}
        \centering
        \includegraphics[width=\linewidth]{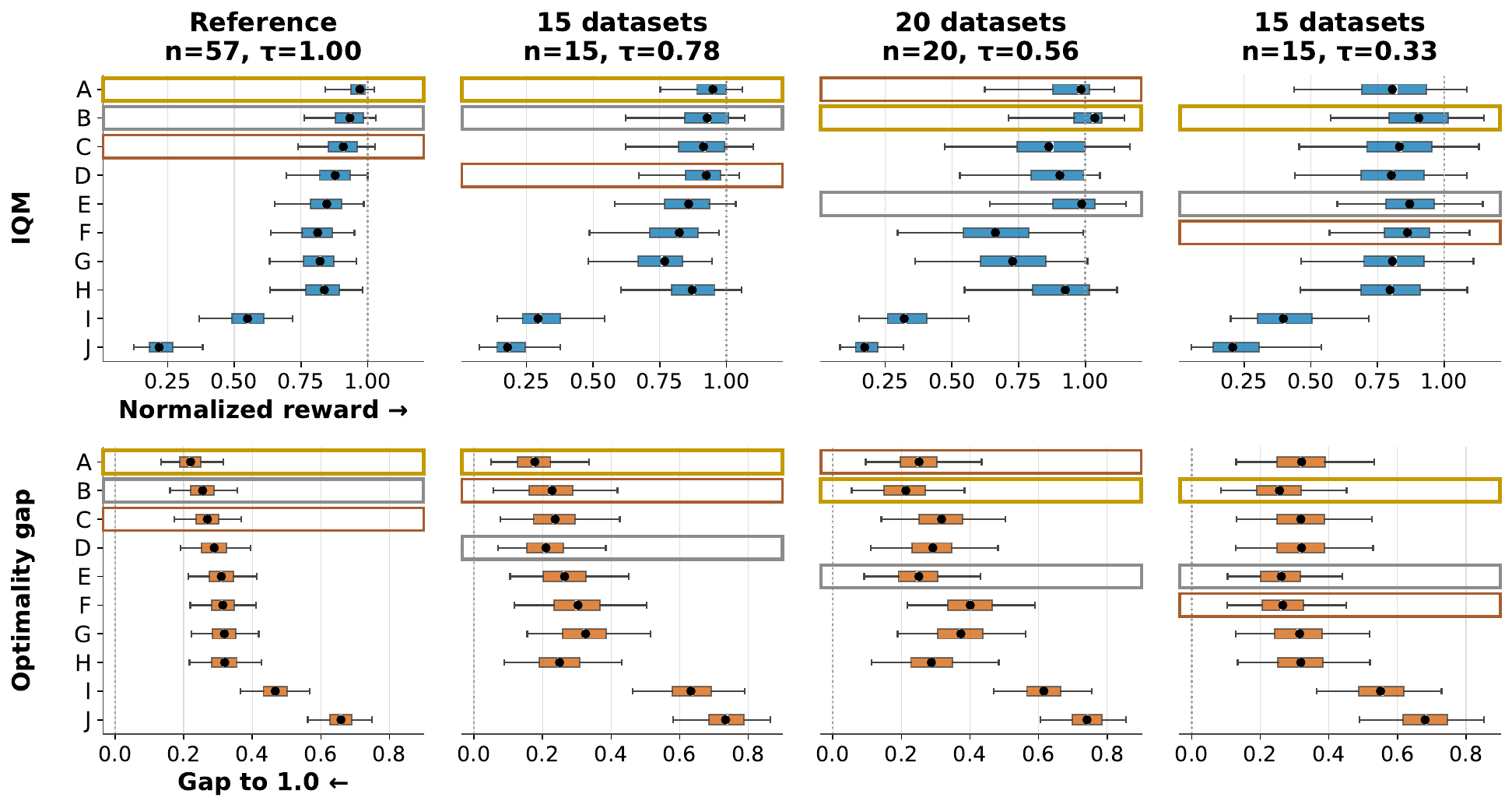}
        \caption{15-dataset subset leaderboard constructions with different Kendall's $\tau$ to the full-environment leaderboard. Even with 15-20 different datasets, it is possible to have rankings in different orders, with the top 3 (gold, silver, bronze outlined) varying substantially. Letters A-J indicate different algorithms.}
        \label{fig:leaderboard-tau}
    \end{figure}
    
    We compare algorithm orderings, or leaderboards, using Kendall's $\tau$. Let $r_i^{(a)}$ and $r_i^{(b)}$ denote the rank of algorithm $i$ under two leaderboards $a$ and $b$, respectively, with rank $1$ corresponding to the best-performing method. For each pair of algorithms $(i,j)$, we compute
    \[
        c_{ij}
        =
        \operatorname{sign}
        \left[
            \left(r_i^{(a)} - r_j^{(a)}\right)
            \left(r_i^{(b)} - r_j^{(b)}\right)
        \right],
    \]
    so that $c_{ij}=1$ if the two leaderboards order the pair consistently, $c_{ij}=-1$ if they reverse the pair, and $c_{ij}=0$ if the pair is tied in either ordering. To visualize the disagreement values, we show a reference leaderboard and a few alternatives with varying Kendall's $\tau$ values in Figure~\ref{fig:leaderboard-tau}. Letters A-J indicate different algorithms, and as ordering deviates from the reference leaderboard, the Kendall's $\tau$ drops. 

    We begin by analyzing how the tuning and budget can materially change the conclusions of a study in Section~\ref{subsec:tuning-budget}. We additionally compare the variance across different hyperparameter sets to the seed variance, and suggest an allocation of resources that removes more uncertainty than simply running more seeds.

    In Section~\ref{subsec:aggregation}, we analyze how the separation, choice and quantity of datasets impacts reliability, and generalize on prior work finding human-generated datasets behaving differently from machine-generated ones, finding that this can be applied to \textit{any} different expert source: planner-based solvers, reinforcement learning, cloned datasets, and human.

    Finally, in Section~\ref{subsec:leaderboard}, we construct leaderboards and summary tables showing final algorithm performance after hyperparameter tuning.
    
    \takeawaybox{We use boxes like this to close each analysis with its main takeaway.}

    \subsection{Tuning budget and seed variance}
    \label{subsec:tuning-budget}
        
        \begin{figure}
            \centering
            \includegraphics[width=0.8\linewidth]{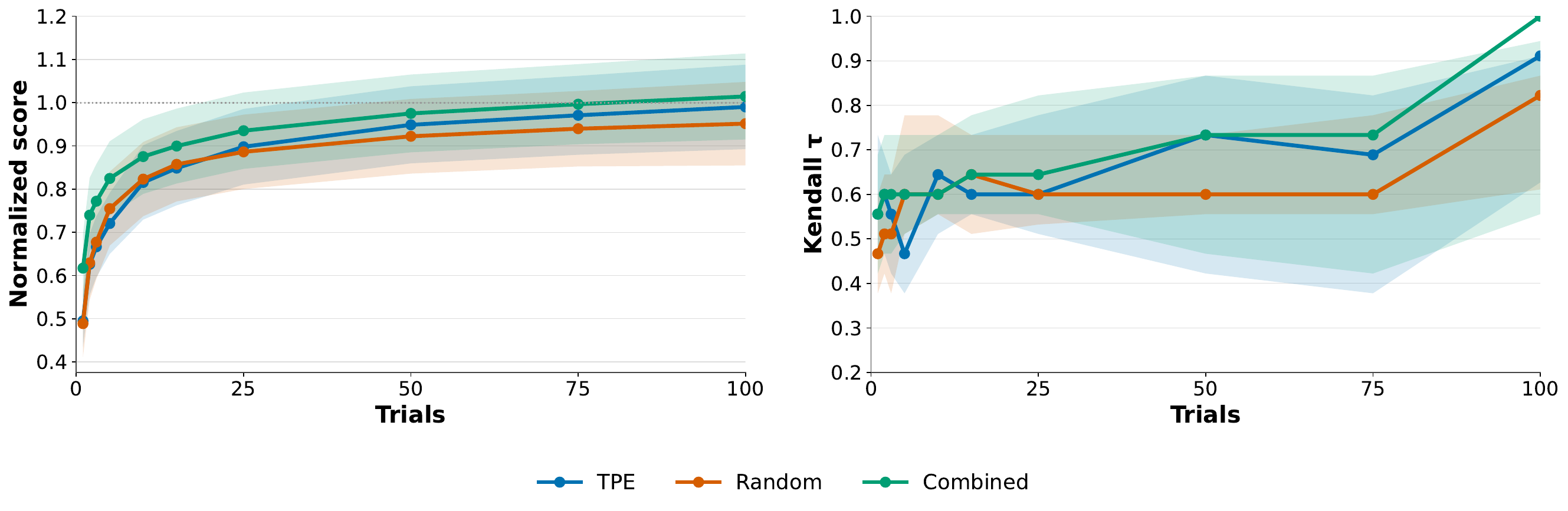}
            \caption{Tuning progress in continuous-control. Most of the improvement in best observed normalized score occurs within 25--50 trials, whereas agreement with the full-budget algorithm ranking improves more gradually. Confidence intervals are across datasets, where peak performance can vary substantially.}
            \label{fig:hp_score_over_time}
        \end{figure}
        
        \begin{figure}
            \centering
            \includegraphics[width=0.8\linewidth]{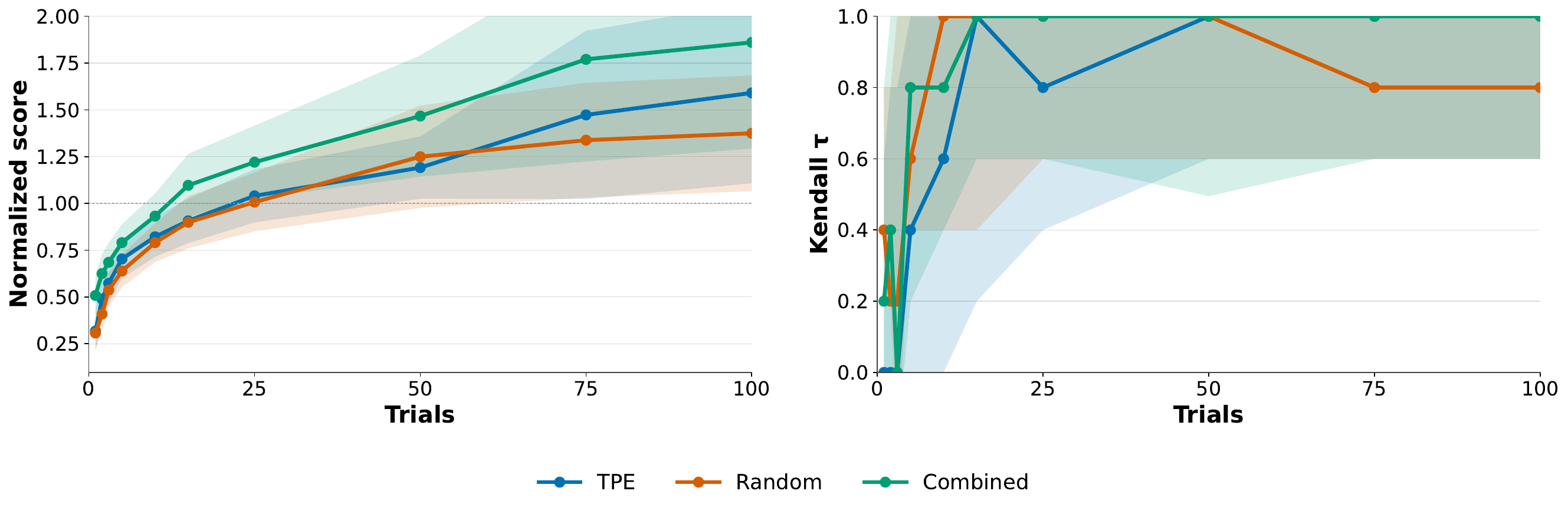}
            \caption{Tuning progress in discrete-control. Agreement with the full-budget ranking increases rapidly during the first 10--25 trials, while the best observed normalized score continues to improve with additional tuning. Confidence intervals are across datasets, where peak performance can vary substantially.}
            \label{fig:hp_tau_over_time}
        \end{figure}
    
        Running 200 hyperparameter trials for every algorithm--dataset pair is beyond the budget of many evaluations. We therefore use the complete sweep as a reference and ask how quickly smaller searches recover both its best observed performance and its algorithm ordering. 
        
        In continuous-control datasets, these two quantities converge at different rates. Figure~\ref{fig:hp_score_over_time} shows that most of the improvement in normalized score occurs within the first 25--50 trials, with smaller gains thereafter. Agreement with the full-budget ranking improves more slowly. A search may therefore find strong configurations before it provides a reliable comparison between algorithms: continuous-control leaderboards remain sensitive to tuning budget even after performance begins to plateau.
        
        Ranking agreement increases more quickly in discrete control. Figure~\ref{fig:hp_tau_over_time} shows a sharp improvement during the first 10--25 trials, after which Kendall's $\tau$ remains generally high despite some fluctuations. The best observed score continues to increase throughout the search.

        \begin{figure}
            \centering
            \includegraphics[width=\linewidth]{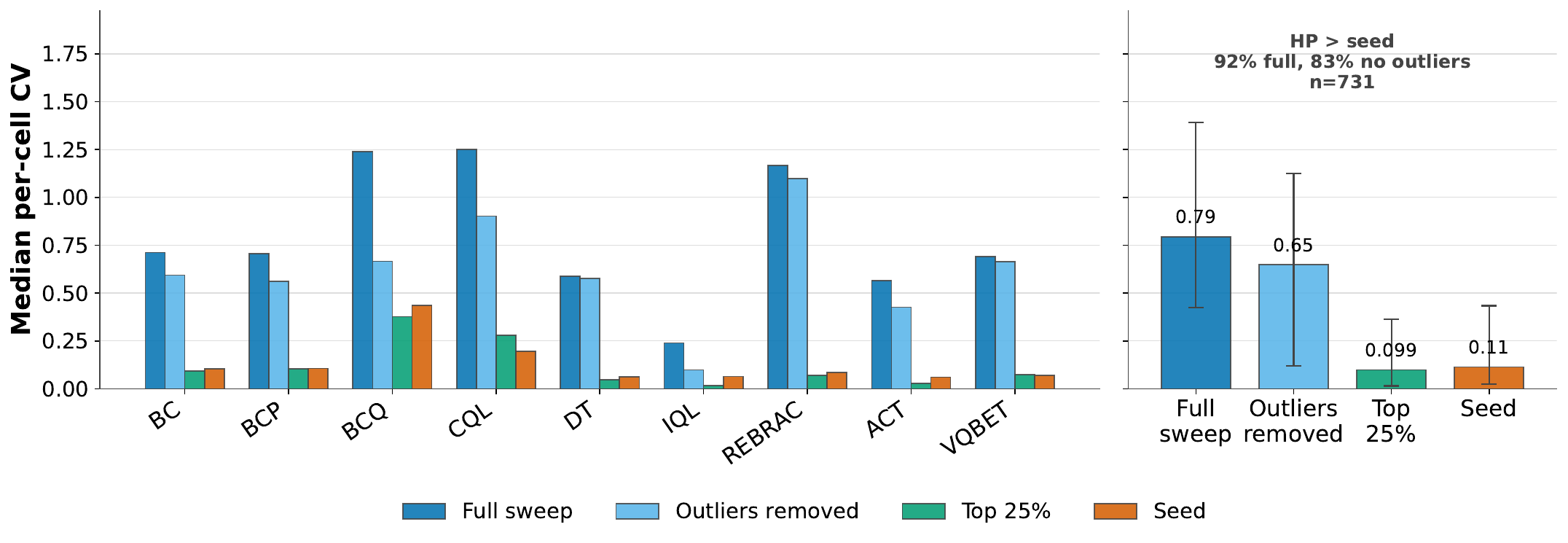}
            \caption{Variability across hyperparameter configurations and random seeds. The left panel reports the median per-cell coefficient of variation for each algorithm, and the right panel summarizes 731 algorithm--dataset cells. Hyperparameter variability exceeds seed-wise variability in 92\% of cells, or 83\% after removing outlier configurations. Within the top 25\% of configurations, however, variability is comparable to that observed across seeds.}
            \label{fig:hp_vs_seed_variance}
        \end{figure}
    
        \begin{figure}[t]
            \centering
            \includegraphics[width=0.9\linewidth]{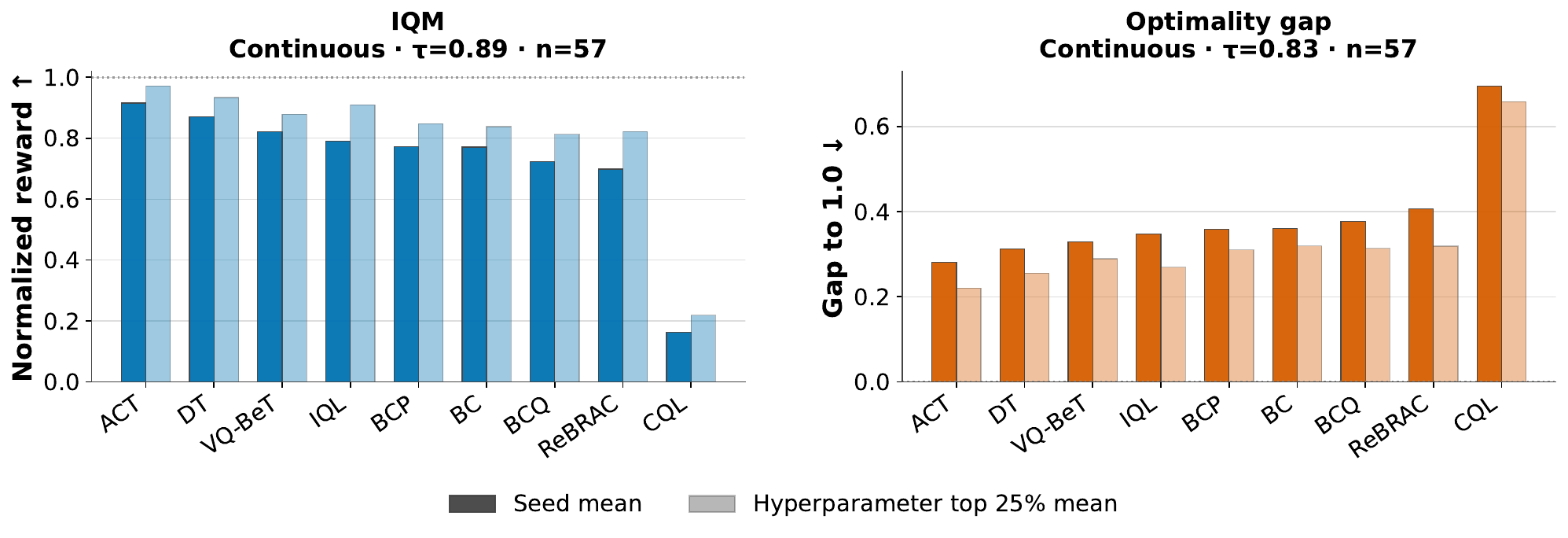}
            \caption{Comparison between continuous-control leaderboards constructed from the top-quartile mean of the hyperparameter sweep and from independently seeded runs. IQM and optimality-gap estimates are nearly identical, and the resulting algorithm rankings have Kendall correlations of $\tau=0.94$ for IQM and $\tau=0.83$ for optimality gap.}
            \label{fig:topquartile-vs-seeds}
        \end{figure}

        Tuning and seed replication answer different questions. Additional configurations explore sensitivity to hyperparameter choice and increase the chance of finding a strong policy, whereas repeated seeds estimate the stochastic variability of a fixed configuration. A finite evaluation budget must therefore balance search-space exploration against replication of the selected configurations.

        Across the complete sweep, hyperparameter choice is the larger source of variation. As shown in Figure~\ref{fig:hp_vs_seed_variance}, variation across configurations exceeds seed-wise variation in $92\%$ of algorithm--dataset cells and in $83\%$ after removing outlier configurations. Results based on only a few configurations can therefore depend more strongly on which hyperparameters were tried than on which random seeds were used. Seed replication is necessary for estimating uncertainty between runs, but additional seeds cannot compensate for an inadequately explored search space and leaves a much larger source of uncertainty unexplored. For our experiments we pick the 95th percentile hyperparameter for each dataset--algorithm pair tested, and train for 5 seeds.
        
        The upper-performing region is considerably more concentrated. Among the top $25\%$ of configurations, variation across hyperparameters is comparable to variation across seeds. We therefore average the normalized scores of the top $25\%$ of configurations within each algorithm--dataset pair, producing one robust near-optimal score per pair, and compute IQM and optimality gap across these dataset-level scores.
        
        Figure~\ref{fig:topquartile-vs-seeds} validates this approximation against independently seeded runs. The two procedures produce nearly identical IQM and optimality-gap estimates, while their continuous-control rankings agree strongly: Kendall's $\tau$ is $0.94$ for IQM and $0.83$ for optimality gap.

        \takeawaybox{Hyperparameter searches may find many strong configurations long before providing a stable comparison between algorithms.}
        
        \takeawaybox{Hyperparameter choice dominates variation across the full search, while the top of the search is much more concentrated. A budget of 25--50 trials provides a useful practical compromise.}

    \subsection{Datasets and Environments}
    \label{subsec:aggregation}

        The datasets and environments included in an evaluation shape both the stability and interpretation of its leaderboard. We examine how rankings change with benchmark size and composition, and whether they transfer across different expert sources.

        \subsubsection{Benchmark Size and Composition}

        \begin{figure}
            \centering
            \includegraphics[width=0.9\linewidth]{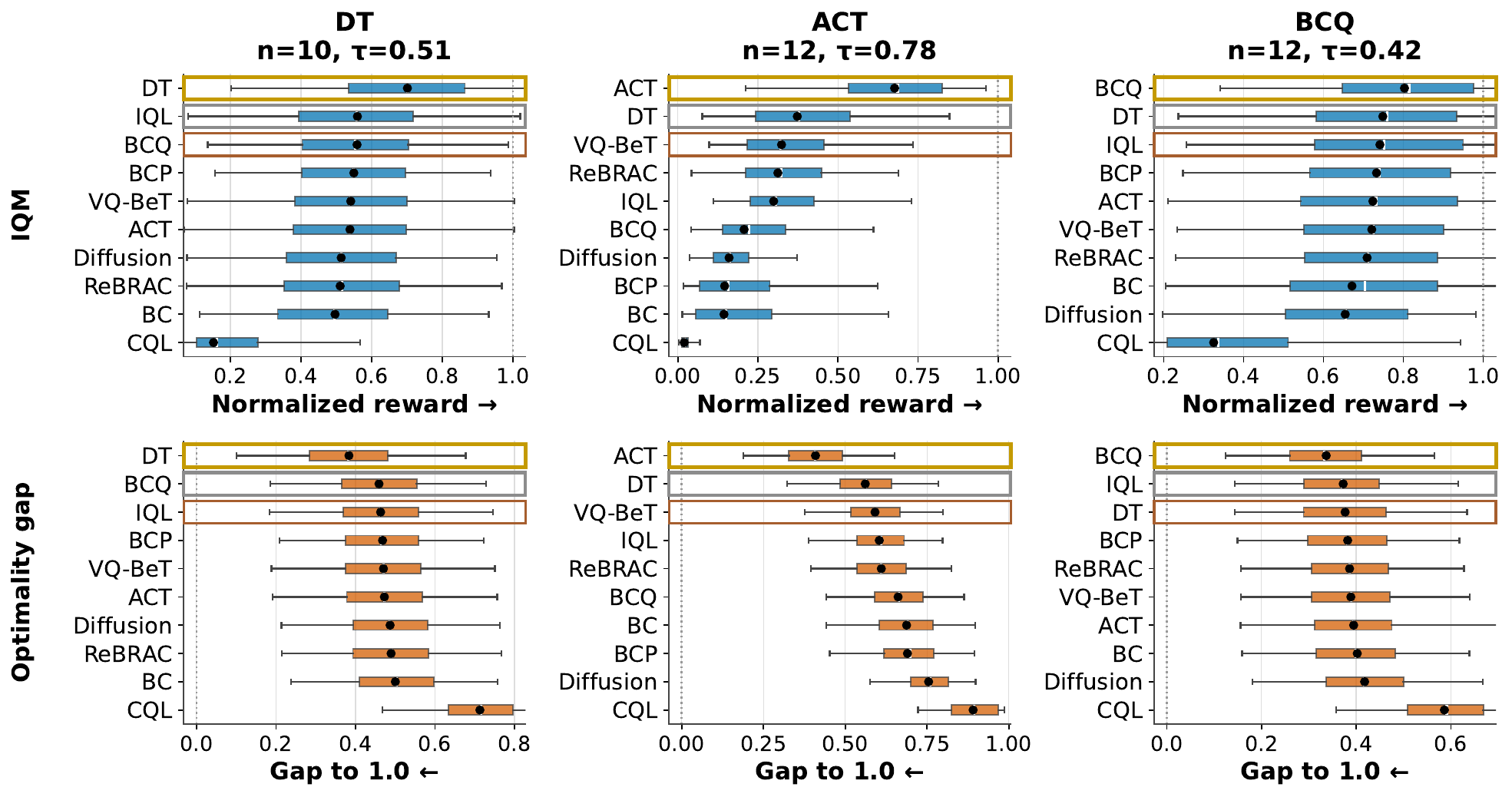}
            \caption{Selected benchmark subsets that favor different algorithms. We sample 2,000 subsets at each tested size ($n\in\{10,12,15,20\}$). Among these candidates, subsets containing 12, 10, and 12 environment--dataset pairs place ACT, DT, and BCQ first by a margin, respectively, under both IQM and optimality gap.}
            \label{fig:gaming-comparisons}
        \end{figure}
        
        Benchmark composition can materially change the conclusion of an evaluation. Small benchmarks are naturally vulnerable to unrepresentative sampling, but size alone may not solve the problem: we ask whether subsets comparable in size to established evaluation suites can still produce substantially different algorithm rankings.
        
        We draw 2,000 random subsets independently at each of four sizes, $n\in\{10,12,15,20\}$, producing 8,000 candidate benchmarks. We then search these candidates for subsets that favor different algorithms. Figure~\ref{fig:gaming-comparisons} shows that ACT, DT, and BCQ can each be placed first under both IQM and optimality gap. The selected subsets contain 12, 10, and 12 environment--dataset pairs, respectively, and are therefore not unusually small. An algorithm other than the winner on the full 57-dataset continuous-control benchmark ranks first under both IQM and optimality gap in 24.70\%, 23.80\%, 20.80\%, and 15.45\% of subsets of sizes $n=10,12,15,20$, respectively. For comparison, the continuous-control evaluation in Decision Transformer contains four environments under three dataset regimes each, yielding 12 environment--dataset pairs \citep{chen2021decision}. 

        These examples show that favorable leaderboards can be found through a moderate random search over benchmark composition. A selectively composed benchmark can therefore match the size of a complete suite used in prior work while supporting a different conclusion about which algorithm performs best.  
        
        \begin{figure}
            \centering
            \includegraphics[width=0.9\linewidth]{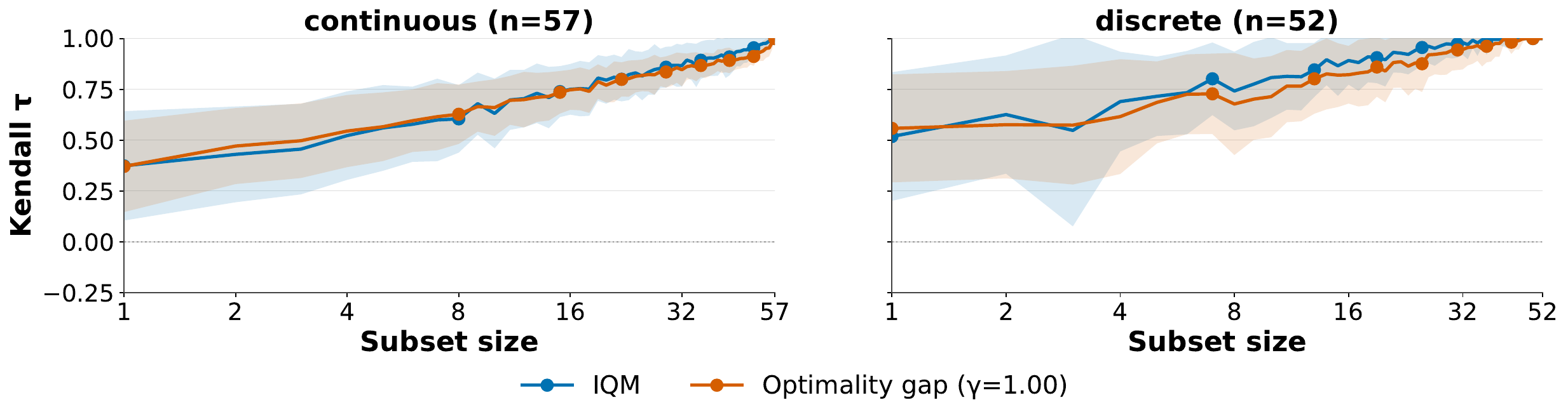}
            \caption{Agreement between subset and full-suite leaderboards as a function of subset size. Lines show the mean Kendall's $\tau$ over randomly sampled subsets, and shading shows one standard deviation. Agreement increases and variability across subsets decreases as more environment--dataset pairs are included.}
            \label{fig:subset-vs-full}
        \end{figure}

        \begin{figure}
            \centering
            \includegraphics[width=0.9\linewidth]{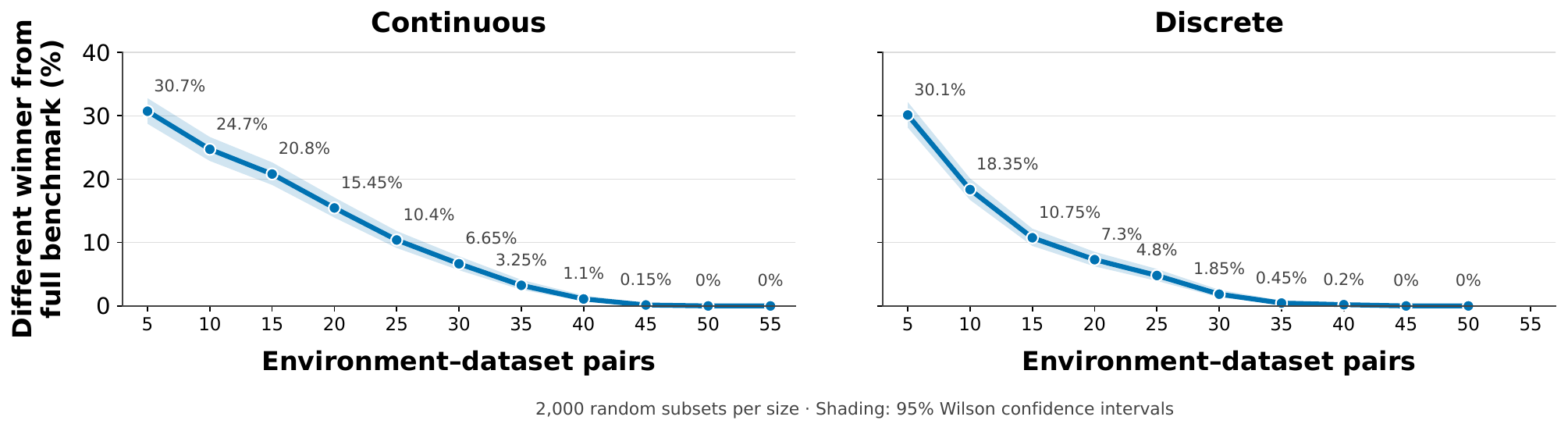}
            \caption{Probability of displaying a winner conflicting with the true top algorithm as the benchmark size increases.}
            \label{fig:winner_probability}
        \end{figure}
        
        Random subsets nevertheless become more reliable as they grow. Figure~\ref{fig:subset-vs-full} shows that their mean agreement with the full-suite leaderboard increases with subset size, while variability across subsets decreases. We show in Figure~\ref{fig:winner_probability} that the probability of producing a conflicting winner to the full-suite leaderboard decreases as more benchmarks are added.
        
        Within our suite, evaluations therefore become less sensitive to the particular pairs sampled as more pairs are included. Strongly discordant leaderboards become less likely to arise through random selection. Carefully selected subsets remain a concern, however: Figure~\ref{fig:gaming-comparisons} shows that benchmarks comparable in size to prior evaluation suites can still be composed to favor different algorithms.

        \takeawaybox{Larger benchmark suites produce more stable algorithm rankings. In our continuous-control experiments, subsets of 10--15 environment--dataset pairs, comparable to prior evaluations, yield a different winner from the full benchmark under both metrics roughly 20--25\% of the time. Increasing the subset size to 25 pairs reduces this rate to about 10\%.}

        \subsubsection{Transfer Across Expert Sources}
        
        Prior work has found that results on agent-generated datasets do not necessarily transfer to human demonstrations \citep{mandlekar2021matterslearningofflinehuman}. We examine whether this is part of a broader pattern by comparing algorithm rankings across several sources of expert data.
        
        We divide the 57 continuous-control datasets according to how their trajectories were generated:
        
        \begin{itemize}
            \item \textbf{Cloned (4 datasets).} Datasets generated wholly or partly by policies trained through behavior cloning.
            \item \textbf{Human (7 datasets).} Datasets collected through direct human control or teleoperation.
            \item \textbf{RL + Planner (6 datasets).} Datasets for which both a planner and a policy trained through reinforcement learning contribute to trajectory generation.
            \item \textbf{RL (32 datasets).} Datasets generated by expert policies trained through reinforcement learning.
            \item \textbf{Scripted (8 datasets).} Datasets generated by a predefined planner or scripted controller rather than a learned policy.
        \end{itemize}
        
        \begin{figure}
            \centering
            \includegraphics[width=\linewidth]{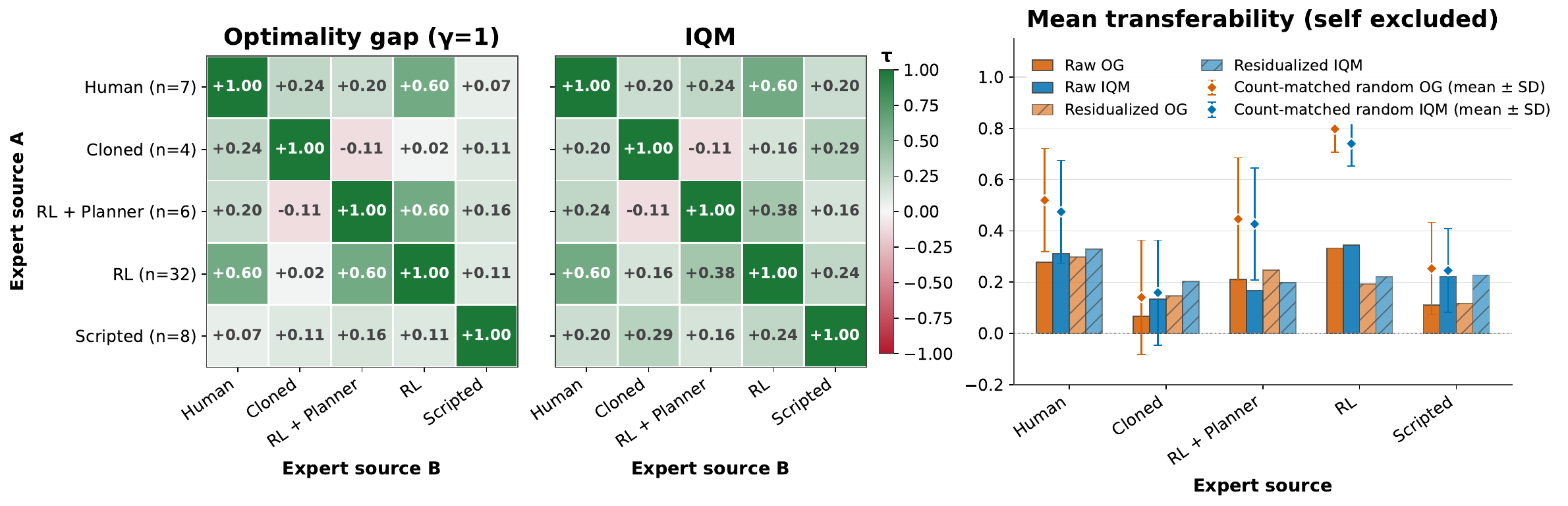}
            \caption{Agreement between algorithm rankings across expert sources. The left and center panels report pairwise Kendall's $\tau$ under optimality gap and IQM. The right panel reports mean off-diagonal agreement before and after controlling for the log number of datasets in each source. The diamond with whiskers represents the count-matched random baseline, with whiskers representing 1 standard deviation.}
            \label{fig:cross-expert-transfer}
        \end{figure}
        
        Figure~\ref{fig:cross-expert-transfer} shows that algorithm rankings transfer unevenly across expert sources. Pairwise Kendall's $\tau$ ranges from $-0.16$ to $0.60$ under optimality gap and from $-0.07$ to $0.47$ under IQM. Most pairs exhibit little or negative agreement. In particular, poor transfer is not confined to comparisons between human and machine-generated data; it also occurs between different methods of generating trajectories computationally.
        
        Averaged over the ten source pairs, Kendall's $\tau$ is $0.164$ under optimality gap and $0.173$ under IQM. Randomly constructed groups with the same sizes reach approximately $0.26$ under both metrics. Rankings transfer unusually poorly, even compared to sampling random datasets to predict the ranking.
        
        Group size explains part of the variation between sources. The raw averages make RL appear more transferable, but this category contains 32 datasets, compared with only 4--8 for each remaining source. After controlling for the log number of datasets, the apparent advantage of RL largely disappears, and every source has lower mean agreement than the random baseline. No single expert source consequently produces a ranking that is broadly representative of the others.
        
        These results extend the distinction previously observed between human and agent-generated demonstrations. Changes in algorithm ordering seem to occur not only between human and machine-generated data, but also between cloned, scripted, planned, and RL-generated datasets. 
         
        \takeawaybox{Algorithm rankings transfer poorly across expert sources and agree less than rankings obtained from size-matched random groups. This failure extends beyond human-versus-machine comparisons to different sources of machine-generated data.}

    \subsection{Leaderboard and evaluations}
    \label{subsec:leaderboard}

        \begin{figure}
            \centering
            \includegraphics[width=\linewidth]{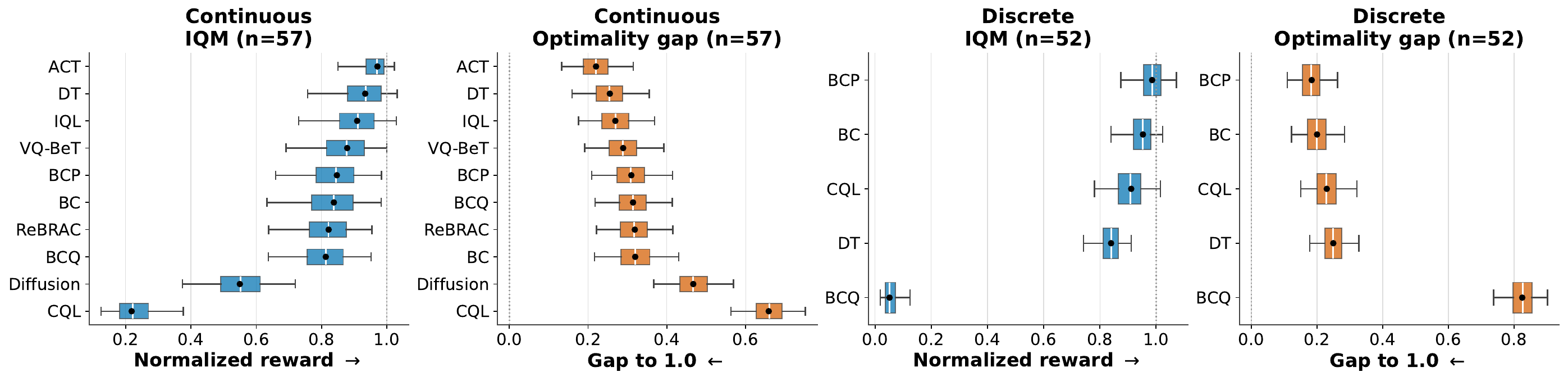}
            \caption{Overall algorithm performance on continuous- and discrete-control datasets. We report interquartile mean (IQM; higher is better) and optimality gap with threshold $\gamma=1$ (lower is better). Markers denote point estimates, while boxes and whiskers summarize bootstrap uncertainty. ACT leads the continuous-control aggregate, whereas BCP leads the discrete-control aggregate under both metrics.}
            \label{fig:leaderboard-overall}
        \end{figure}
        
        \begin{figure}
            \centering
            \includegraphics[width=\linewidth]{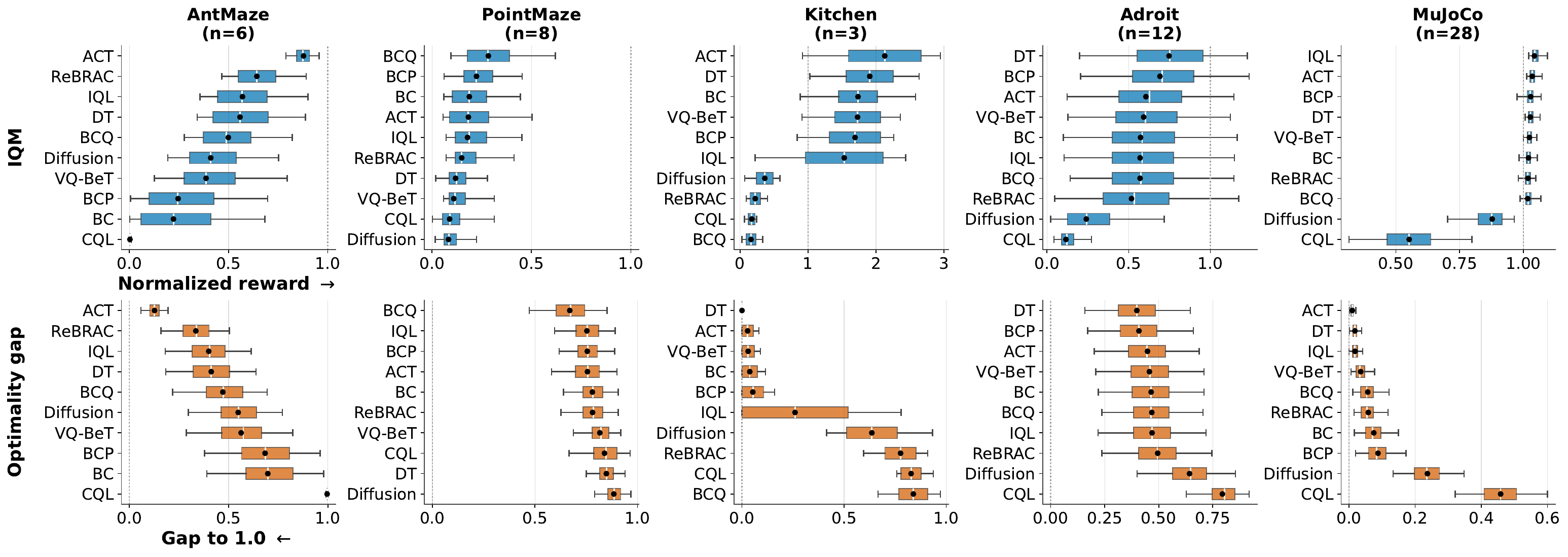}
            \caption{Continuous-control performance separated by environment collection. The leading algorithm changes across collections: ACT leads on AntMaze and Kitchen, BCQ on PointMaze, DT on Adroit, and IQL or ACT on MuJoCo depending on the aggregation metric.}
            \label{fig:leaderboard-collection}
        \end{figure}
        
        \begin{figure}
            \centering
            \includegraphics[width=\linewidth]{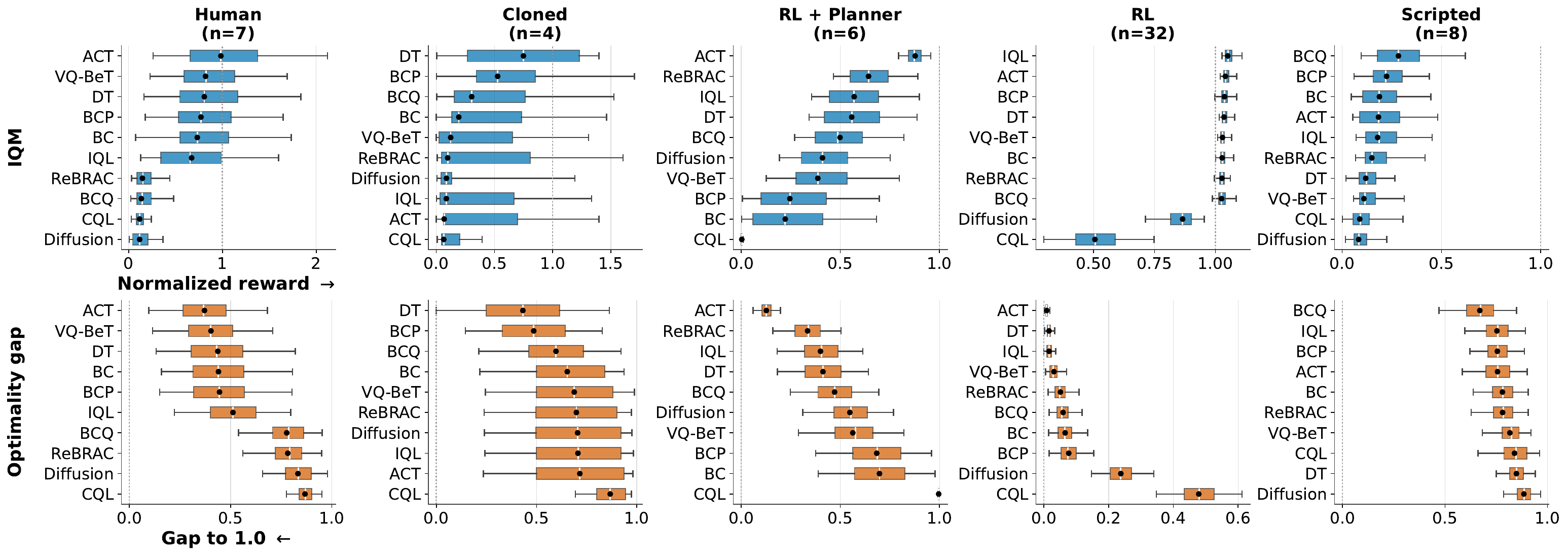}
            \caption{Continuous-control performance separated by the source of the trajectories in each dataset. The leading algorithm changes from ACT on human and RL~+~planner data, to DT on cloned data, IQL on RL data, and BCQ on scripted data.}
            \label{fig:leaderboard-expert-source}
        \end{figure}
        
        We next ask which algorithms perform well across the suite and where their relative strengths change. Figure~\ref{fig:leaderboard-overall} presents the overall continuous- and discrete-control leaderboards, while Figures~\ref{fig:leaderboard-collection} and~\ref{fig:leaderboard-expert-source} separate continuous-control results by environment collection and expert source. Full per-dataset scores are reported in Tables~\ref{tab:continuous-leaderboard} and~\ref{tab:atari-leaderboard}.
        
        ACT is the strongest algorithm in the overall continuous-control aggregate, ranking first under both IQM and optimality gap. Its advantage is broad rather than universal: ACT also leads on AntMaze, Kitchen, human demonstrations, and RL~+~planner data, but different methods take the lead elsewhere. BCQ performs best on PointMaze and scripted datasets, DT leads on Adroit and cloned data, and IQL obtains the highest IQM on MuJoCo and RL-generated data. The aggregate leaderboard therefore provides a useful summary, but it obscures substantial changes in algorithm ordering across evaluation regimes.
        
        The human-data results illustrate this distinction. ACT remains first and VQ-BeT rises to second under both metrics, consistent with VQ-BeT's design for multimodal behavior generation from demonstration data \citep{lee2024vqbet}. DT follows, while BCP places fourth under IQM. 
        
        The simple cloning baselines are especially competitive. BCP uses a small MLP and differs from ordinary behavior cloning primarily through data filtering, yet it remains close to the leading group in several continuous-control settings. Its discrete-control performance is even stronger: BCP ranks first on Atari under both IQM and optimality gap, with unfiltered BC second. Prior studies such as CORL~\citep{tarasov2023revisiting} often have behavior cloning at the bottom of their leaderboards, but in our continuous control suite they appear to be closer to the middle post-tuning. These are by far the cheapest and simplest methods we study, the performance here is thus quite impressive. 
        
        Diffusion and CQL produce the weakest continuous-control aggregates, although for different reasons. We evaluate the U-Net variant of Diffusion Policy as a standalone policy trained separately on each dataset \citep{chi2023diffusion}. It achieves competitive returns on many AntMaze, Adroit, and MuJoCo datasets but fails severely on others. These results should not be interpreted as evidence that diffusion-based action generation is flawed. In foundation VLAs such as $\pi_0$, the related flow-matching objective is used to model multimodal action chunks within a much larger pretrained system \citep{black2026pi0visionlanguageactionflowmodel}. Its action expert is also substantially smaller than the VLM backbone, and cached representations avoid repeatedly evaluating the full model during generation. This differs considerably from training Diffusion Policy from scratch as the complete policy. Action chunking itself is unlikely to explain our results because other chunked policies perform strongly under the same evaluation pipeline. Finally, extending training from 100 to 300 epochs produced little additional reward at substantially greater cost.
        
        CQL exhibits a sharp domain reversal. It ranks last in the overall continuous-control comparison but third on Atari, ahead of DT and far ahead of BCQ. Because the original CQL results were obtained on a different set of datasets \citep{kumar2020conservative}, we checked whether this discrepancy could be explained by our implementation. We adapted the original repository to load our Minari datasets and evaluated it on our suite of environments and datasets, represented in Figure~\ref{fig:header-figure}, excluding PointMaze because the original work reports neither results nor tuning guidance for that collection. Following the procedure used by CORL, we applied the authors' tuning recommendations when available and otherwise used the hyperparameters reported for the most similar dataset. When these configurations failed, we additionally evaluated a small set of existing published configurations. The original implementation produced performance comparable to our tuned models across the overlapping settings, making a simple implementation failure an unlikely explanation for CQL's continuous-control performance. Prior reproduction and benchmarking efforts similarly find that CQL is highly sensitive to hyperparameters, low-level implementation choices, and the distribution of the offline data \citep{tarasov2023revisiting,kang2023improving}. Further implementation and tuning details are provided in Appendix~\ref{app:og-baselines}.
        
        Finally, the value-function-based methods do not behave as a single coherent group. IQL is strongest on RL-generated and MuJoCo data, ReBRAC is competitive on RL~+~planner data, and BCQ leads on scripted and PointMaze datasets. Their relative performance changes sharply elsewhere: BCQ performs poorly on Atari despite leading PointMaze, while IQL and ReBRAC are much less competitive on human demonstrations.
        
        \takeawaybox{ACT is the strongest overall continuous-control method, while the simple BCP baseline leads Atari. No algorithm dominates every data regime: environment collection and expert source substantially change which method performs best, particularly beyond the top position.}
    
\section{Hyperparameter Sensitivity and Transfer}
\label{sec:hyperparam-sens}

    The previous section showed that tuning budget can materially change algorithm rankings. We now ask which hyperparameters drive those changes, whether the same hyperparameters matter across environments, and how a broad sweep can be distilled into a strong default configuration. We examine sensitivity first and then turn to transfer.

    \subsection{Hyperparameter sensitivities}
    \label{subsec:sensitivity}
    
        \begin{figure}
            \centering
            \includegraphics[width=\linewidth]{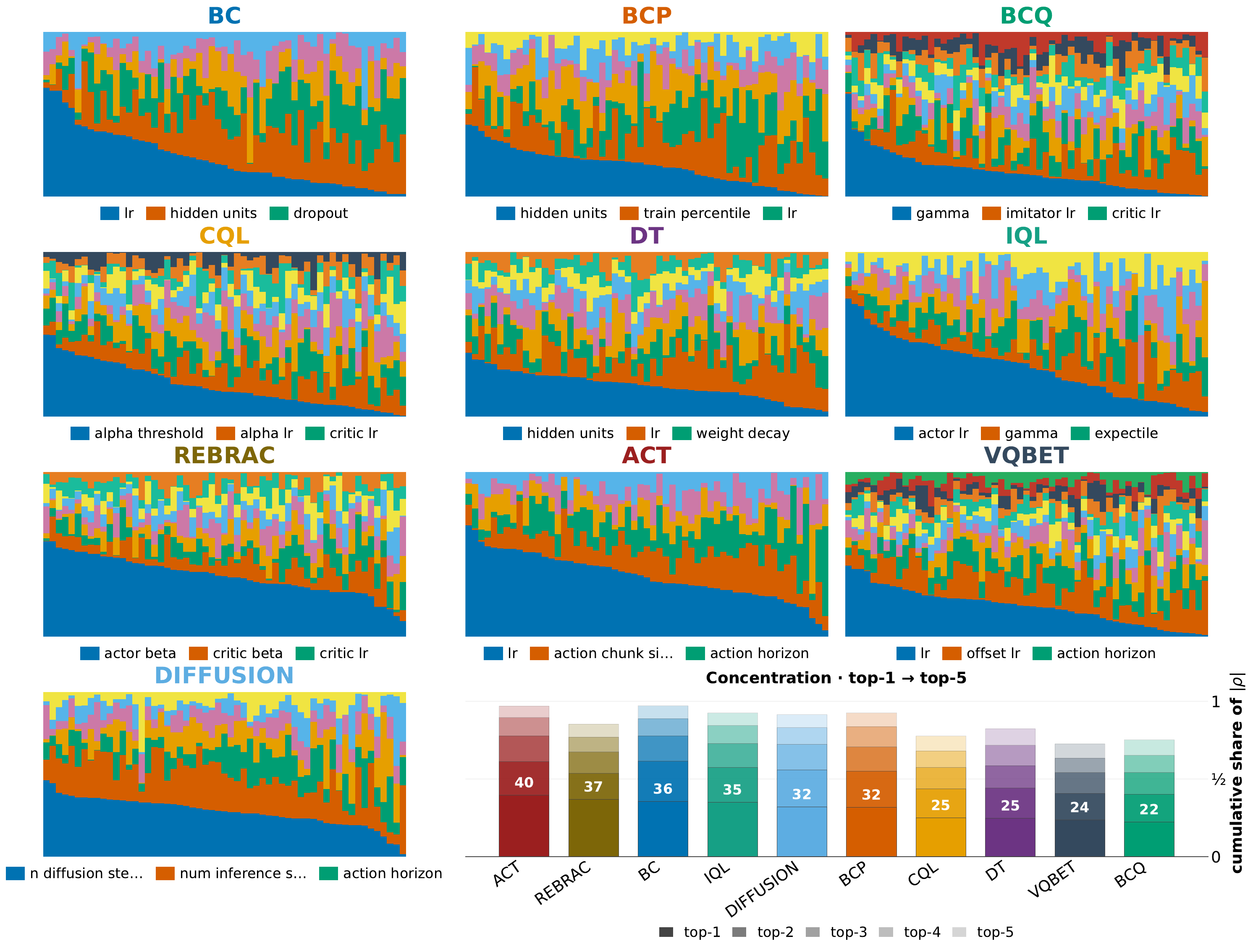}
            \caption{Hyperparameter sensitivity across the 57 continuous-control environments. For each algorithm--environment pair, we compute the absolute Spearman correlation $|\rho_h|$ between each hyperparameter and trial score and normalize these magnitudes as $s_h=|\rho_h|/\sum_j|\rho_j|$. Each vertical bar represents one environment and shows the resulting sensitivity shares. The bottom panel reports the median cumulative share of the one through five most influential hyperparameters within each environment; the displayed value is the median top-one share. Environment ordering is chosen separately within each algorithm for readability.}
            \label{fig:hp-sensitivity-continuous}
        \end{figure}
        
        \begin{figure}
            \centering
            \includegraphics[width=\linewidth]{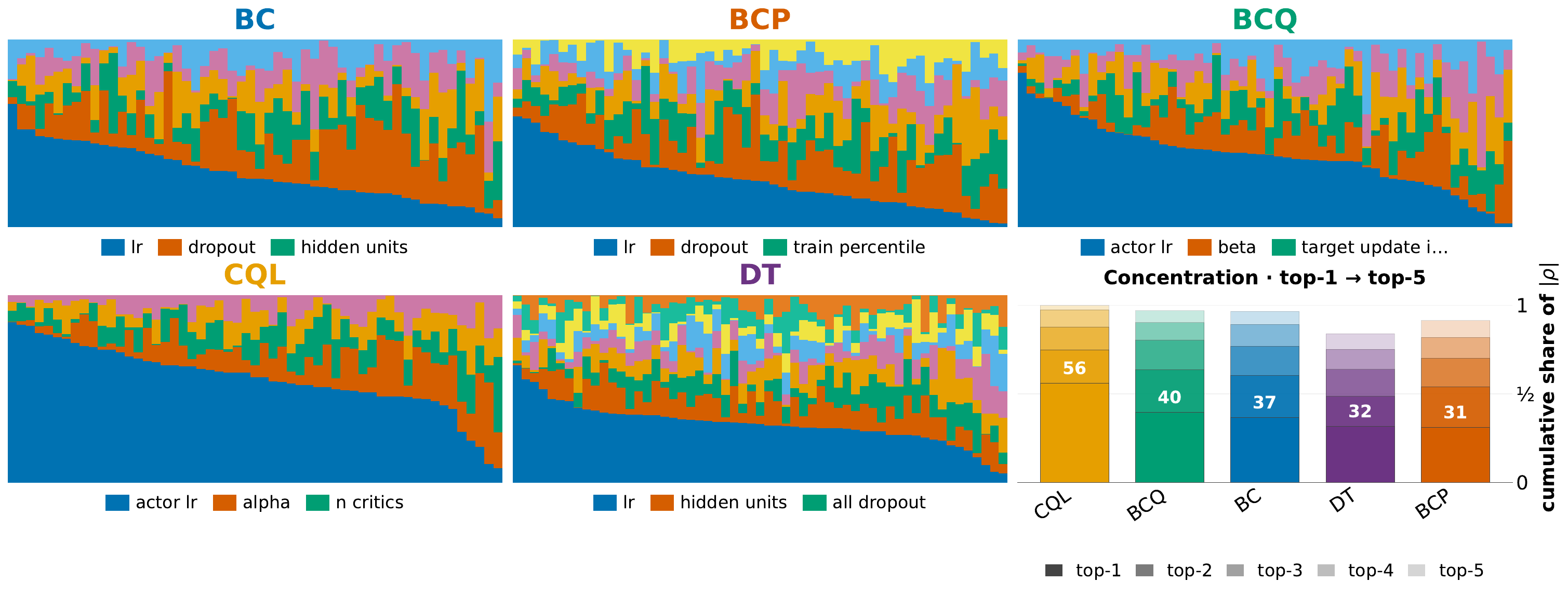}
            \caption{Hyperparameter sensitivity across the 54 Atari environments, computed as in Figure~\ref{fig:hp-sensitivity-continuous}. The sensitivity profiles are more consistent than in continuous control: learning rate is the most influential hyperparameter for every algorithm, with particularly strong concentration for CQL and BCQ.}
            \label{fig:hp-sensitivity-atari}
        \end{figure}
        
        Figures~\ref{fig:hp-sensitivity-continuous} and~\ref{fig:hp-sensitivity-atari} visualize how sensitivity is distributed across algorithms and environments. For each algorithm--environment pair, we measure the marginal association between hyperparameter $h$ and trial performance using Spearman's rank correlation $\rho_h$. Because the direction of the relationship is secondary to whether a hyperparameter affects performance, we take the absolute correlation and normalize it within the environment:
        
        \[
        s_h = \frac{|\rho_h|}{\sum_j |\rho_j|}.
        \]
        
        The resulting value measures each hyperparameter's share of the total marginal rank association within an environment. We report both the median $|\rho_h|$ and the median normalized share in Tables~\ref{tab:hp-sensitivity-continuous} and~\ref{tab:hp-sensitivity-discrete}.
        
        We find that most algorithms are primarily sensitive to one to three hyperparameters, but their identities often change across datasets. Although learning rate is the most important hyperparameter for ACT and VQ-BeT in aggregate, action horizon dominates ACT on \texttt{kitchen/complete}, \texttt{humanoid/simple}, \texttt{humanoid/medium}, \texttt{pointmaze/medium-dense}, and \texttt{door/cloned}. For VQ-BeT, it dominates on both \texttt{inverteddoublependulum} datasets, \texttt{invertedpendulum/expert}, and \texttt{antmaze/medium-diverse}. Even rarely dominant parameters can matter locally: VQ-BeT's commitment weight leads on \texttt{hammer/human}, while its secondary-code weight leads on \texttt{pen/human}. Full per-environment heatmaps are provided in Figure~\ref{fig:hp-sensitivity-breakdown} of Appendix~\ref{app:sens-breakdown}. 

        Sensitivity remains concentrated even after accounting for differences in search-space size. For VQ-BeT, the three most important hyperparameters within an environment account for a median of $74.5\%$ of total sensitivity, compared with $25\%$ under a uniform allocation across its 12 tuned parameters. The corresponding values are $70.9\%$ versus $27.3\%$ for BCQ and $85.8\%$ versus $33.3\%$ for ReBRAC. By comparison, ACT's top-three share is higher in absolute terms at $88.7\%$, but its six-parameter search space gives a uniform reference of $50\%$.
        
        The dominant hyperparameter is not always the one conventionally emphasized. Learning rate ranks first for every discrete algorithm, but the continuous-control results are more varied: $\gamma$ leads for BCQ, hidden-unit count for DT, actor regularization $\beta$ for ReBRAC, the conservative-loss threshold for CQL, and the number of diffusion steps for Diffusion Policy. Consequently, tuning only learning rate and a few standard optimization parameters can miss the principal source of variation for several algorithms.
        
        We additionally investigate whether the sensitivities are driven by suboptimal (too wide) ranges, resulting in divergent runs. We trimmed 10\% and 25\% of the runs at the boundaries, and find that the original top hyperparameter remains completely unchanged. The same leading hyperparameter remained ranked first for all algorithms under both trims.
        
        \takeawaybox{Sensitivity is usually concentrated, but its identity is not always transferable. In continuous control, the top three hyperparameters capture $54\%$--$78\%$ of total absolute rank association, yet the dominant parameters often change across environments.}

    \subsection{Better Default Hyperparameters}
    \label{subsec:transfer}

        Given the cost of running 25--50 tuning trials, we attempt to reduce the burden on practitioners by deriving strong default configurations for existing algorithms. Algorithm libraries commonly provide a single default configuration, but these defaults can leave substantial performance on the table. We therefore ask whether sweeps already completed by an algorithm's developers can be distilled into data-driven defaults that transfer to previously unseen environments without target-specific tuning.
    
        \textbf{Deriving improved defaults.} Our heuristic is motivated by the hypothesis that, although the best hyperparameter value may vary across environments, good values occupy broad, overlapping regions of the search space. Values near the center of high-performing configurations may therefore transfer more reliably than either a library default or the best configuration from a single environment.
    
        For each algorithm and training environment, we retain the top $10\%$ of trials by score. We pool these trials across environments and select each numerical hyperparameter by its median, computing the median in log-space for log-scaled parameters. For categorical parameters, we use the mode. Restricting the calculation to high-performing trials avoids clearly poor regions, while the median reduces the influence of noisy trials and environment-specific extremes. This procedure produces one fixed configuration per algorithm that can directly replace its existing library default.
    
        \begin{figure}
            \centering
            \includegraphics[width=0.7\linewidth]{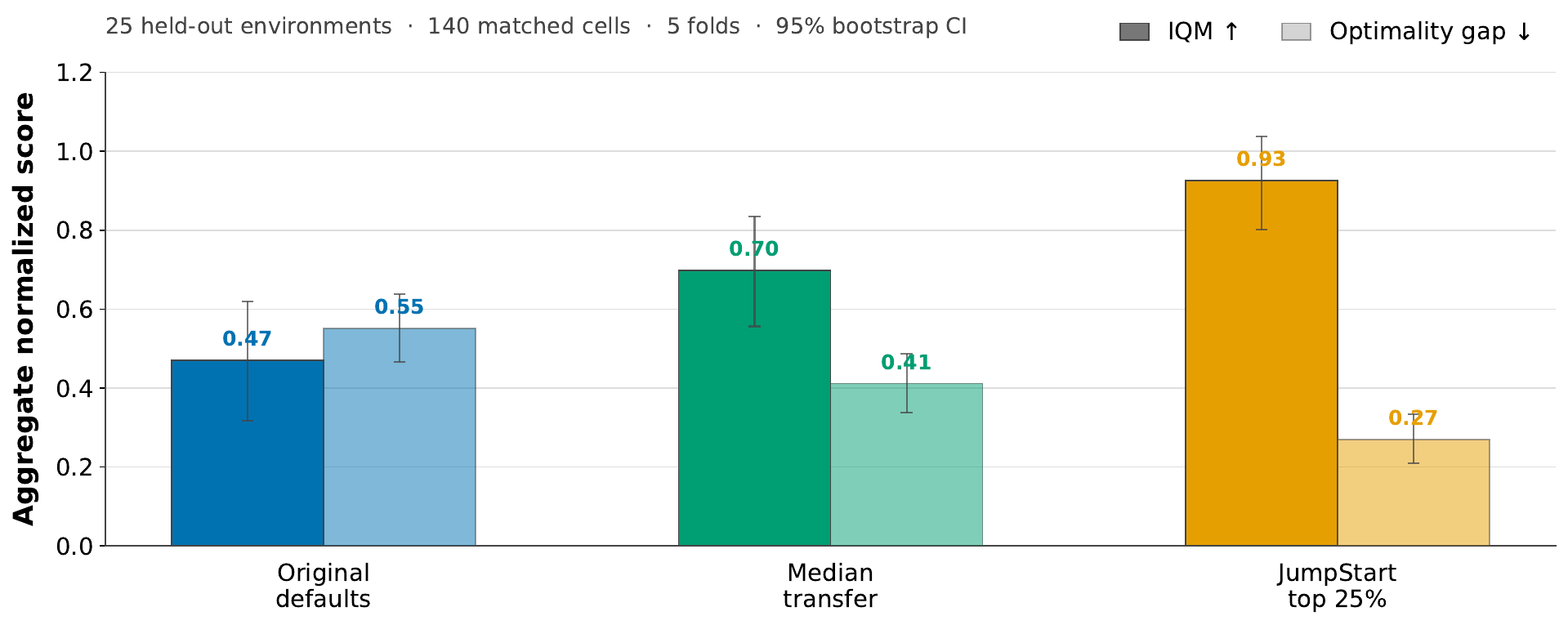}
            \caption{Performance of existing \texttt{d3rlpy} defaults, median-transfer defaults, and fully tuned JumpStart top-quartile results. The evaluation contains five folds of five held-out environments, yielding 25 environments and 140 matched algorithm--environment cells. Each fold's test environments are excluded when constructing its transferred defaults. Error bars are $95\%$ bootstrap confidence intervals.}
            \label{fig:transfer-summary}
        \end{figure}
    
        \textbf{Do the improved defaults work?} We evaluate BC, BCQ, CQL, DT, IQL, and ReBRAC against their existing \texttt{d3rlpy} defaults~\citep{d3rlpy}. The evaluation uses five folds containing five held-out environments each. For every fold, its test environments are excluded from the construction of the transferred configuration.
    
        As shown in Figure~\ref{fig:transfer-summary}, our defaults increase IQM from $0.47$ to $0.70$ and reduce optimality gap from $0.55$ to $0.41$. Fully tuned JumpStart results reach an IQM of $0.93$ and an optimality gap of $0.27$. Median transfer therefore recovers roughly half of the improvement obtained through full target-specific tuning without evaluating any configuration on the target environment. Recovering the remaining improvement without target-specific search is difficult because both the identities of the dominant hyperparameters and their best values vary across environments, preventing a single coordinate-wise default from reproducing each target's specialized joint configuration. 
    
        These aggregate results establish that the complete configurations transfer, but do not explain why. We next test whether their individual hyperparameters fall within regions associated with strong performance on unseen environments.
    
        \begin{figure}
            \centering
            \includegraphics[width=0.8\linewidth]{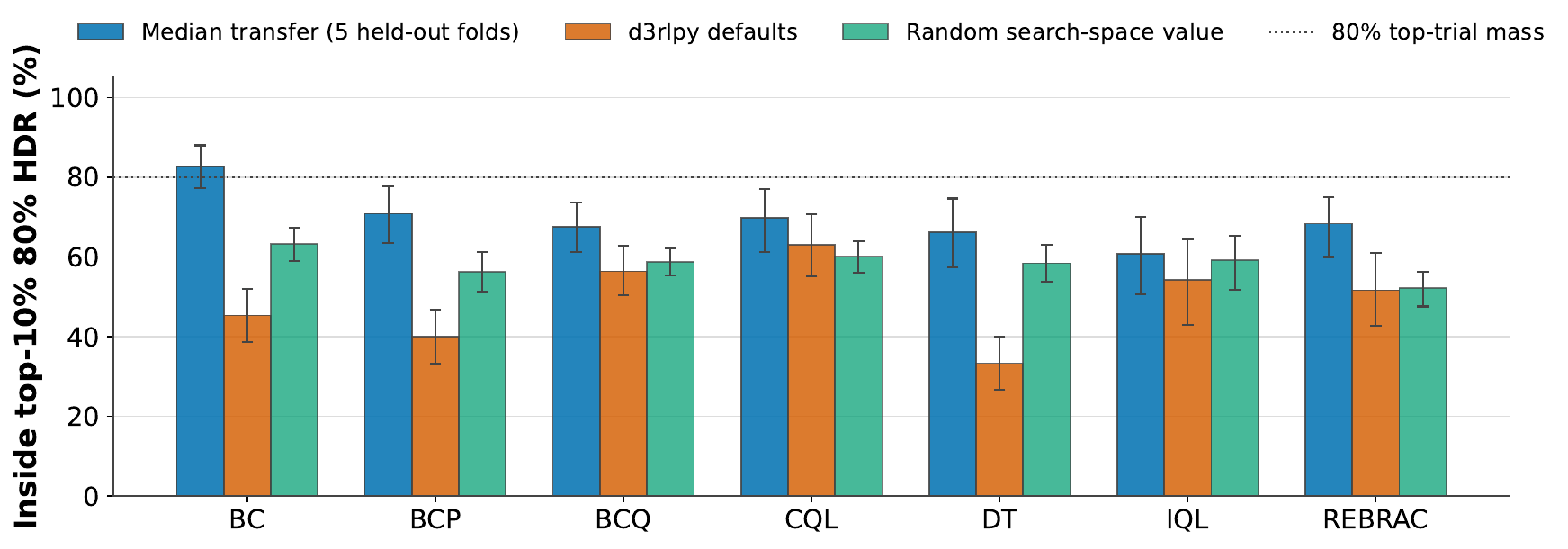}
            \caption{Marginal support for candidate defaults across five folds. Within each fold, its five final test environments are excluded, and each remaining environment is held out in turn. A value is supported when it lies inside the $80\%$ highest-density region (HDR) of hyperparameter values among the target environment's top $25\%$ of trials. We compare our median-transfer defaults with the existing \texttt{d3rlpy} defaults and the exact coverage of a value drawn from the configured search distribution. The dotted line denotes the $80\%$ probability mass contained by the HDR by construction, not an expected transfer rate. Error bars are $95\%$ environment-bootstrap confidence intervals.}
            \label{fig:transfer-support}
        \end{figure}
    
        \textbf{Why do the defaults transfer?} Within each fold, we exclude its five final test environments and hold out each remaining training environment in turn. For every algorithm, target environment, and hyperparameter, we recompute the transferred default using only the other training environments. We then estimate the distribution $\hat p_{e,h}$ of good values on the target from its top $25\%$ of trials, using a boundary-corrected kernel density estimate for numerical parameters and a lightly smoothed empirical probability mass function for categorical parameters.
    
        For numerical parameters, we evaluate $\hat p_{e,h}$ on a dense grid spanning the configured search bounds and normalize it using numerical integration. We then lower a density threshold until the grid intervals above that threshold contain $80\%$ of the estimated probability mass. Boundary correction is applied when estimating the density, before this numerical integration. For categorical parameters, we analogously select the highest-probability categories containing at least $80\%$ of the smoothed probability mass. We record whether each candidate default lies inside the resulting region. Random search-space coverage is computed directly from the probability assigned to the same region by the original tuning distribution.
    
        Figure~\ref{fig:transfer-support} shows that median-transfer coverage ranges from approximately $61\%$ to $83\%$ across algorithms, compared with $33\%$--$63\%$ for the existing defaults and $52\%$--$63\%$ for a random search-space value. The median-transfer default has the highest coverage for every evaluated algorithm. This supports the intuition behind the heuristic: high-performing configurations commonly occupy regions shared across environments rather than isolated, environment-specific optima.
    
        This density analysis is marginal and does not capture interactions between hyperparameters. Nevertheless, its agreement with the held-out evaluation suggests that the full configurations transfer because their individual values repeatedly fall within broad regions associated with strong performance. The transferred configurations should therefore be interpreted as strong starting points rather than replacements for full tuning.
    
        \takeawaybox{Taking the coordinate-wise median of high-performing configurations from prior environments produces strong transferable defaults. Across 25 held-out environments, these defaults raise IQM from $0.47$ to $0.70$ and reduce optimality gap from $0.55$ to $0.41$, recovering roughly half of the benefit of full tuning without target-specific search.}

\section{Recommendation System for Practitioners}
\label{sec:recommendation-system}

    The preceding analyses show that the best algorithm depends on the dataset. We therefore train a dataset-conditioned recommender that uses observable properties of a continuous-action dataset and its environment to rank ten offline policy-learning algorithms and return the three most promising.

    \textbf{Prediction task.} For each algorithm--dataset pair, the target is the normalized mean performance of the trials in the top $25\%$,
    \[
    y_{a,d} =
    \frac{\overline{R}_{a,d}^{\,\mathrm{top}\text{-}25\%}-R_d^{\mathrm{random}}}
         {R_d^{\mathrm{dataset}}-R_d^{\mathrm{random}}},
    \]
    where $R_d^{\mathrm{dataset}}$ is the mean trajectory return in the dataset. Each dataset is one training example, with target vector $\mathbf y_d=(y_{1,d},\ldots,y_{10,d})$. The recommender predicts this vector and returns the three algorithms with the highest predicted scores.

    Features include dataset size, episode statistics, observation and action dimensionality, task domain, expert source, reward type, and trajectory-return statistics. We exclude collection names, dataset identities, exact data-generation algorithms, and benchmark results. Missing categorical values are represented explicitly as unknown, while numerical values are imputed within each training fold.

    \textbf{Candidate regressors.} We evaluate Ridge, Elastic Net, partial least squares, $k$-nearest neighbors, RBF-SVR, random forest, histogram gradient boosting, and ExtraTrees. These models span linear, low-rank, local, kernel, and tree-based approaches suited to our small tabular dataset; we omit neural predictors because only 57 dataset-level examples are available.

    \begin{algorithm}[t]
        \caption{Evaluation of each candidate recommender}
        \label{alg:recommender-evaluation}
        \footnotesize
        \begin{algorithmic}[1]
            \ForAll{recommender models $m$}
                \ForAll{environment families $g$}
                    \State Hold out every dataset from family $g$
                    \State Tune $m$ on the remaining datasets using three-fold grouped cross-validation
                    \State Refit $m$ on all remaining datasets
                    \State Predict scores for the ten algorithms on each held-out dataset
                    \State Recommend the three algorithms with the highest predicted scores
                \EndFor
                \State Evaluate all out-of-family recommendations
            \EndFor
        \end{algorithmic}
    \end{algorithm}

    \textbf{Evaluation protocol.} We evaluate on 57 datasets from 30 environment families using leave-one-environment-family-out validation, as summarized in Algorithm~\ref{alg:recommender-evaluation}. All variants of the target environment are removed together. Model selection, preprocessing, and hyperparameter tuning occur entirely within the outer training split, so the held-out family is used only for final evaluation.

    We compare against a fold-local global-mean baseline, which recommends the three algorithms with the highest mean performance on the outer training datasets regardless of target characteristics. This represents the common strategy of selecting whichever algorithms are strongest overall.

    We report top-three overlap and best-of-three regret. Top-three overlap is the fraction of the true top three recovered by the recommendation. Best-of-three regret measures the score lost after trying all three recommendations,
    \[
    \operatorname{regret}_d =
    \max_a y_{a,d}
    -
    \max_{a\in\widehat{\mathcal A}_d^{(3)}} y_{a,d},
    \]
    where $\widehat{\mathcal A}_d^{(3)}$ is the recommended set. Regret is low whenever at least one recommended algorithm performs close to the true winner, even if the complete ordering is imperfect.
    
    \begin{figure}[t]
        \centering
        \includegraphics[width=\linewidth]{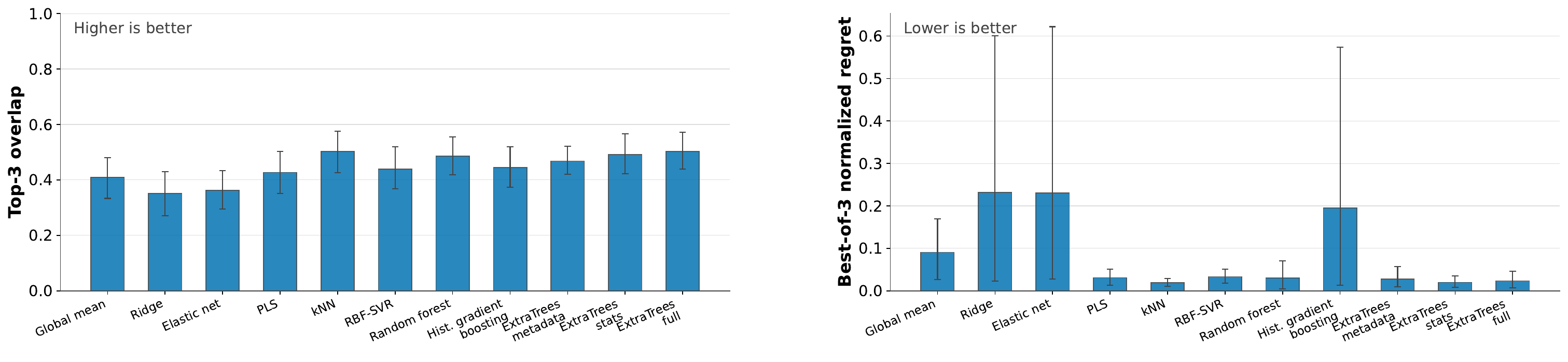}
        \caption{Algorithm-recommendation performance under leave-one-environment-family-out validation. The global-mean baseline recommends the fold-local leaderboard to every target, while the learned models condition on observable dataset features and are tuned entirely within each outer training split. Higher top-three overlap and lower best-of-three regret are better. Error bars show $95\%$ environment-clustered bootstrap confidence intervals.}
        \label{fig:recommendation-evaluation}
    \end{figure}
    
    \textbf{Main recommendation results.} Figure~\ref{fig:recommendation-evaluation} shows that both random forest and ExtraTrees improve substantially over the global leaderboard. Random forest increases top-three overlap from $0.409$ to $0.485$, includes the true winner among its recommendations on $63.2\%$ rather than $47.4\%$ of datasets, and reduces best-of-three regret from $0.091$ to $0.030$. It also improves NDCG@3 from $0.769$ to $0.847$ and Kendall's $\tau$ from $0.324$ to $0.401$.
    
    ExtraTrees obtains slightly stronger point estimates on the main evaluation, with $0.503$ overlap and $0.023$ regret, but the two tree models are statistically indistinguishable. Its overlap advantage over random forest is $0.018$, with a paired $95\%$ confidence interval of $[-0.035,0.077]$, and its regret advantage is similarly uncertain. Relative to the global baseline (ranked by mean performance across all training datasets), random forest improves overlap by $0.076$ and reduces regret by $0.061$; the paired confidence interval for the regret reduction is $[0.011,0.130]$.
    
    \begin{figure}[t]
        \centering
        \includegraphics[width=0.6\linewidth]{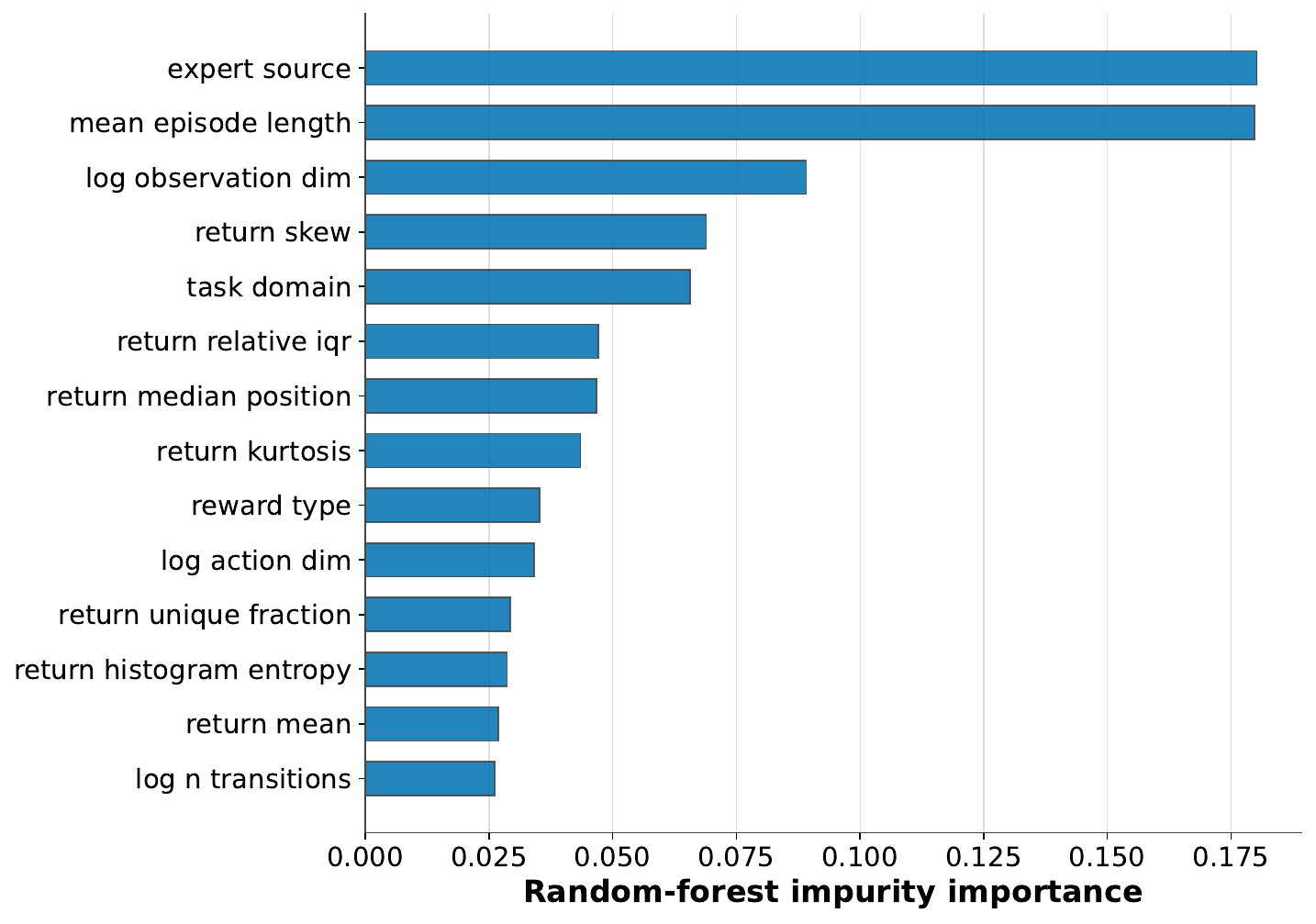}
        \caption{Impurity-based feature importance for the random-forest recommender. Expert source and mean episode length provide the strongest predictive signal, followed by observation dimensionality, task domain, and properties of the return distribution.}
        \label{fig:recommendation-features}
    \end{figure}
    
    \textbf{What information supports the recommendations?} Figure~\ref{fig:recommendation-features} shows that expert source and mean episode length are the most informative features, followed by observation dimensionality, return skew, and task domain. Several additional return-distribution statistics provide secondary signal, whereas dataset size contributes relatively little. Because expert source is the strongest feature, the source-held-out evaluation below is an important test of whether the recommender can generalize when this information has not been observed during training.
    
    \begin{figure}[t]
        \centering
        \includegraphics[width=\linewidth]{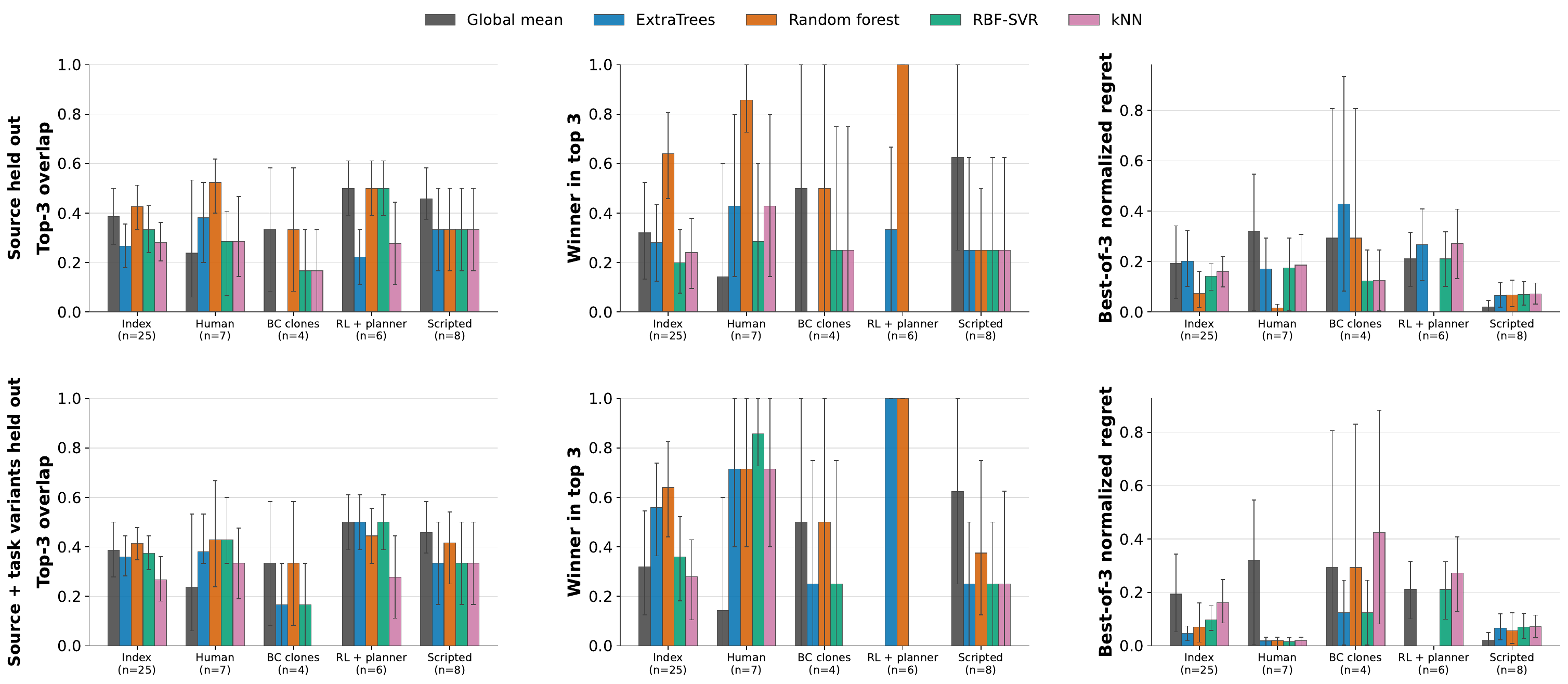}
        \caption{Transfer to minority expert-source categories. The top row removes each target source from training; the bottom row additionally removes all training datasets from environment families represented in the corresponding test set. ``Index'' aggregates the 25 datasets across Human, BC-clone, RL-plus-planner, and Scripted sources. Error bars show $95\%$ environment-clustered bootstrap confidence intervals.}
        \label{fig:expert-source-transfer}
    \end{figure}
    
    \textbf{Transfer to unseen expert sources.} We stress-test the recommender on four minority source categories: Human demonstrations, BC clones, RL-plus-planner data, and Scripted data. Each test dataset is first predicted using a model trained without any datasets from its source category. As shown in Figure~\ref{fig:expert-source-transfer}, random forest obtains $0.427$ aggregate top-three overlap, includes the winner on $64.0\%$ of datasets, and achieves regret of $0.073$. ExtraTrees obtains $0.267$ overlap, $28.0\%$ winner coverage, and $0.202$ regret, while the global baseline obtains $0.387$, $32.0\%$, and $0.194$, respectively.
    
    In the stricter test, we also remove every training dataset from environment families represented in the test set. Random forest retains $0.413$ overlap and $64.0\%$ winner coverage, compared with $0.360$ and $56.0\%$ for ExtraTrees. 
    
    We do not perform a corresponding source-held-out test for RL-generated data. These datasets account for 32 of the 57 datasets, contain all locomotion coverage, and supply much of the observed variation in trajectory and action dimensionality. Removing them would therefore confound expert-source shift with task-domain and feature-support shifts. Generalization among RL-generated datasets remains covered by the main environment-family holdout.
    
    \begin{figure}[t]
        \centering
        \includegraphics[width=0.8\linewidth]{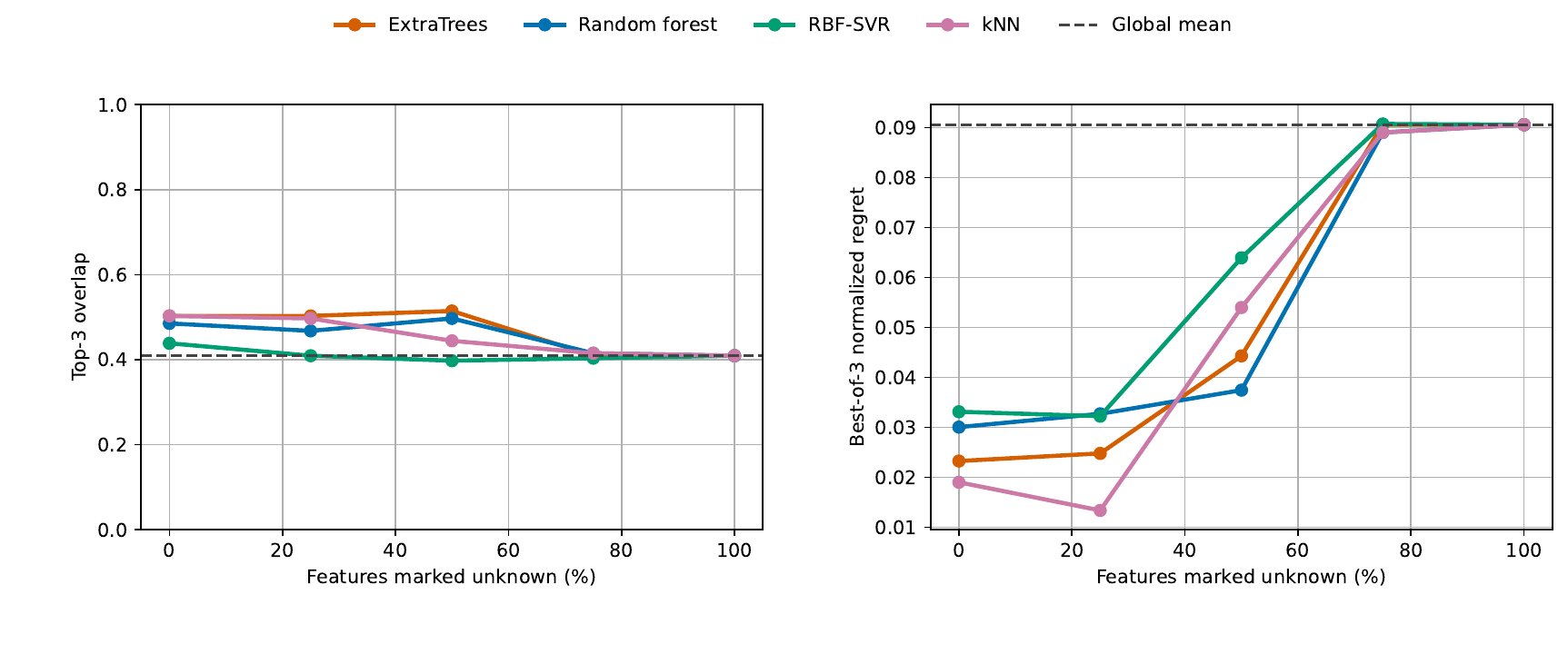}
        \caption{Recommendation quality as test features are randomly marked unknown. The tree models remain effective with half of the features missing. When fewer than $35\%$ are available, the recommender explicitly falls back to the fold-local global leaderboard, causing all methods to converge to the baseline at high missingness.}
        \label{fig:recommendation-missingness}
    \end{figure}
    
    \textbf{Incomplete dataset information.} Figure~\ref{fig:recommendation-missingness} shows that the tree recommenders remain stable under moderate feature masking. With $50\%$ of features unknown, random forest obtains $0.497$ top-three overlap and $0.037$ regret, compared with $0.515$ and $0.044$ for ExtraTrees. At higher missingness, both models converge to the explicit global-leaderboard fallback rather than making poorly supported predictions. 
    
    \begin{figure}[t]
        \centering
        \includegraphics[width=0.8\linewidth]{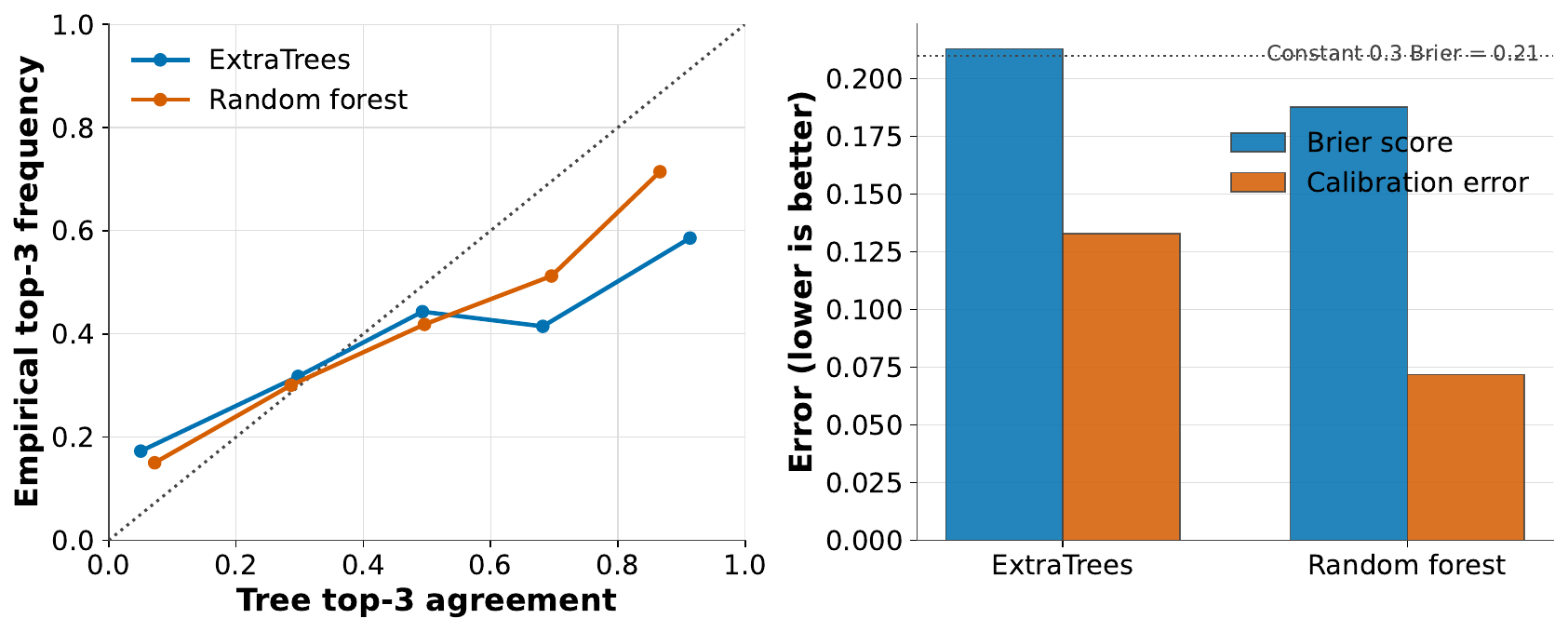}
        \caption{Reliability of ensemble agreement as a recommendation-confidence measure. Left: empirical top-three frequency against the fraction of trees placing an algorithm in the top three. Right: Brier score and calibration error. Random forest is better calibrated.}
        \label{fig:recommendation-confidence}
    \end{figure}
    
    \textbf{Interpreting confidence.} Tree agreement provides a measure of how stable a recommendation is across the ensemble. Random forest produces better-calibrated agreement scores than ExtraTrees, with a Brier score of $0.188$ rather than $0.213$ and expected calibration error of $0.072$ rather than $0.133$.
    
    We select random forest as the final recommender because it appears to be the most robust, while providing strong predictive performance.

\section{Takeaways and Discussion}
\label{sec:discussion}

    Reliable evaluation and comparison have been longstanding challenges across machine learning. A benchmark should do more than produce a single ordering of methods: it should reveal where each method succeeds or fails, and how strongly its conclusions depend on tuning, randomness, and benchmark composition. We distill our findings into practical guidance for developing, evaluating, and reporting offline policy-learning methods.

    \textbf{Pilot experiments.} Early in a project, researchers commonly compare a proposed method with a small set of baselines to decide whether further investment is justified. Such pilots are only informative when the baselines are competitive; otherwise, an apparent improvement may reflect weak configurations rather than a better method. Figure~\ref{fig:header-figure} shows that tuning can substantially improve published baselines and change their relative ordering. JumpStart provides a consistent set of strong, downloadable baselines and their configurations, allowing new methods to be compared against competitive reference points without first reproducing a large tuning sweep. We hope this makes pilot experiments more informative and allows unpromising directions to be identified earlier.

    \textbf{Hyperparameters.} Hyperparameters should be treated as part of a method's evaluation rather than as incidental implementation details. Studies should report their search spaces, search procedure, tuning budget, and model-selection rule, along with the hyperparameters to which performance is most sensitive. Repeating a fixed configuration across seeds answers a different question from exploring additional configurations and cannot compensate for an inadequately searched space. In our experiments, 25--50 trials provides a useful practical compromise, although continuous-control rankings remain sensitive to further tuning. When target-specific tuning is infeasible, the transferred defaults developed in Section~\ref{subsec:transfer} provide a stronger starting point than existing library defaults, but they should not be interpreted as a replacement for tuning when accurate comparisons are required.

    \textbf{Evaluation suites and methodology.} Broader evaluation generally produces more stable conclusions, although size alone does not guarantee a representative benchmark. For context, the main Decision Transformer comparisons contain nine D4RL locomotion datasets and four Atari games~\citep{chen2021decision}; d3rlpy evaluates 17 datasets~\citep{d3rlpy}; and the unified evaluations of \citet{kang2023improving} and \citet{tarasov2023revisiting} cover 26 and 30 tasks or datasets, respectively. Our results show that subsets of only 10--12 environment--dataset pairs can produce conflicting winners, and that even similarly sized suites can support different conclusions depending on their composition. Researchers should therefore report the complete benchmark composition, use as broad a suite as practical, and disaggregate results across important dataset properties. In particular, rankings transfer poorly across expert sources, so evaluations containing human, scripted, planned, or RL-generated data should report these regimes separately rather than relying only on a pooled aggregate.

    Evaluation protocols must also be consistent across methods. We recommend robust aggregate metrics such as IQM and optimality gap, accompanied by uncertainty estimates and per-dataset results~\citep{agarwal2021deep}. Baseline scores should be obtained using the same environment versions, wrappers, normalization procedure, checkpoint-selection rule, and evaluation budget as the proposed method. Numbers should not be copied from prior papers unless these details are known to match, particularly when environment versions differ. By releasing trained policies, their scores and hyperparameters, and the corresponding training and evaluation code, JumpStart allows baselines to be reevaluated as protocols change rather than leaving their published scores permanently tied to one implementation.

    \textbf{Choosing an algorithm.} Our results do not support a single universally best algorithm. Performance depends on observable properties of the dataset and environment, and rankings can change across task domains and expert sources. The practitioner recommender in Section~\ref{sec:recommendation-system} uses these properties to produce a shortlist of three promising algorithms rather than asserting a universal winner. This shortlist is intended to guide an initial pilot: when its recommendations are confident, it can reduce the number of methods that must be tested; when confidence is low or task information is limited, the global leaderboard remains a safer fallback.

    \textbf{Living benchmarks and leaderboards.} Static leaderboards become outdated as algorithms, datasets, and evaluation practices change. We therefore release an open-source website and extensible result database that support external submissions and self-hosting. New algorithms and datasets can be added while retaining the policies, scores, hyperparameters, and evaluation provenance underlying each entry. Because we release every trained policy, existing results can also be reevaluated as environments, metrics, and evaluation practices evolve. This allows the benchmark to develop with the field while keeping its comparisons reproducible and auditable.

\section{Conclusion}

    We presented JumpStart, an extensively tuned comparison of ten offline policy-learning algorithms across 114 datasets and more than 160,000 trained policies. Our results show that algorithm rankings depend substantially on hyperparameter tuning, benchmark composition, and dataset characteristics, limiting the conclusions that can be drawn from small or weakly tuned evaluations. We translate these findings into practical tuning guidance, stronger default configurations, and dataset-conditioned algorithm recommendations. By releasing every trained policy with its metadata and code through an extensible leaderboard, JumpStart can incorporate new results and reevaluate existing models as evaluation practices change, providing an evolving foundation for future offline policy-learning research.

\clearpage
\newcommand{\hprangetile}[2]{%
    \begin{minipage}[t]{0.32\linewidth}
        \vspace{0pt}
        \centering
        \textbf{#1}\par\vspace{1pt}
        \begin{tabular*}{\linewidth}{@{\extracolsep{\fill}}lr@{}}
            \toprule
            Parameter & Range \\
            \midrule
            #2
            \bottomrule
        \end{tabular*}
    \end{minipage}%
}

\begin{table}[t]
    \centering
    \caption{Hyperparameter search spaces for continuous-control algorithms. Fixed parameters are omitted.}
    \label{tab:hp-ranges-continuous}

    \begingroup
    \scriptsize
    \renewcommand{\arraystretch}{0.92}
    \setlength{\tabcolsep}{2pt}

    \emph{Notation:} $\mathrm{U}$ is uniform, $\mathrm{LogU}$ is log-uniform, $\mathrm{Int}$ is inclusive integer, and braces denote uniformly sampled choices.
    \vspace{4pt}

    \hprangetile{BC}{
        Learning rate & $\mathrm{LogU}[10^{-5},10^{-2}]$ \\
        Weight decay & $\mathrm{LogU}[10^{-4},10^{-1}]$ \\
        Gradient clip & $\mathrm{U}[0.1,2]$ \\
        Dropout & $\mathrm{U}[0,0.3]$ \\
        Hidden units & $\{128,256,512\}\times2$ \\
        Policy type & \{det., stoch.\} \\
    }
    \hfill
    \hprangetile{BCP}{
        Learning rate & $\mathrm{LogU}[10^{-5},10^{-2}]$ \\
        Weight decay & $\mathrm{LogU}[10^{-4},10^{-1}]$ \\
        Gradient clip & $\mathrm{U}[0.1,2]$ \\
        Dropout & $\mathrm{U}[0,0.3]$ \\
        Hidden units & $\{128,256,512\}\times2$ \\
        Train percentile & $\mathrm{U}[0,90]$ \\
        Policy type & \{det., stoch.\} \\
    }
    \hfill
    \hprangetile{BCQ}{
        Actor LR & $\mathrm{LogU}[10^{-5},10^{-2}]$ \\
        Critic LR & $\mathrm{LogU}[10^{-5},10^{-2}]$ \\
        Imitator LR & $\mathrm{LogU}[10^{-5},10^{-2}]$ \\
        Discount $\gamma$ & $\mathrm{LogU}[0.9,0.999]$ \\
        Target $\tau$ & $\mathrm{U}[0.001,0.02]$ \\
        Critics & $\mathrm{Int}[1,4]$ \\
        Actor-update interval & $\mathrm{Int}[1,4]$ \\
        $\lambda$ & $\mathrm{U}[0.5,0.95]$ \\
        Action samples & $\mathrm{Int}[20,200]$ \\
        RL start step & $\mathrm{Int}[0,10000]$ \\
        $\beta$ & $\mathrm{U}[0.1,0.9]$ \\
    }

    \par\medskip

    \hprangetile{CQL}{
        Actor LR & $\mathrm{LogU}[10^{-5},10^{-2}]$ \\
        Critic LR & $\mathrm{LogU}[10^{-5},10^{-2}]$ \\
        Temperature LR & $\mathrm{LogU}[10^{-5},10^{-2}]$ \\
        Alpha LR & $\mathrm{LogU}[10^{-5},10^{-2}]$ \\
        Discount $\gamma$ & $\mathrm{LogU}[0.9,0.999]$ \\
        Target $\tau$ & $\mathrm{U}[0.001,0.02]$ \\
        Conservative weight & $\mathrm{U}[0.1,10]$ \\
        Alpha threshold & $\mathrm{U}[-1,20]$ \\
        Action samples & $\mathrm{Int}[5,20]$ \\
        Critics & $\mathrm{Int}[1,4]$ \\
    }
    \hfill
    \hprangetile{DT}{
        Learning rate & $\mathrm{LogU}[10^{-5},10^{-2}]$ \\
        Weight decay & $\mathrm{LogU}[10^{-4},10^{-1}]$ \\
        Gradient clip & $\mathrm{U}[0.1,2]$ \\
        Context size & $\{10,20,30\}$ \\
        Hidden units & $\{128,256\}\times3$ \\
        Dropout & $\mathrm{U}[0,0.3]$ \\
        Attention heads & $\{1,2,4\}$ \\
        Layers & $\mathrm{Int}[2,5]$ \\
        Discount $\gamma$ & $\mathrm{LogU}[0.9,0.999]$ \\
        Reward scale & $\{0,1\}$ \\
    }
    \hfill
    \hprangetile{IQL}{
        Actor LR & $\mathrm{LogU}[10^{-5},10^{-2}]$ \\
        Critic LR & $\mathrm{LogU}[10^{-5},10^{-2}]$ \\
        Discount $\gamma$ & $\mathrm{LogU}[0.9,0.999]$ \\
        Target $\tau$ & $\mathrm{U}[0.001,0.02]$ \\
        Critics & $\mathrm{Int}[1,4]$ \\
        Expectile & $\mathrm{U}[0.5,0.9]$ \\
        Weight temperature & $\mathrm{U}[1,10]$ \\
    }

    \par\medskip

    \hprangetile{ReBRAC}{
        Actor LR & $\mathrm{LogU}[10^{-5},10^{-2}]$ \\
        Critic LR & $\mathrm{LogU}[10^{-5},10^{-2}]$ \\
        Actor $\beta$ & $\mathrm{LogU}[10^{-5},1]$ \\
        Critic $\beta$ & $\mathrm{LogU}[10^{-5},1]$ \\
        Discount $\gamma$ & $\mathrm{LogU}[0.9,0.999]$ \\
        Target $\tau$ & $\mathrm{U}[0.001,0.02]$ \\
        Critics & $\mathrm{Int}[1,4]$ \\
        Smoothing $\sigma$ & $\mathrm{U}[0,0.5]$ \\
        Smoothing clip & $\mathrm{U}[0,1]$ \\
    }
    \hfill
    \hprangetile{ACT}{
        Learning rate & $\mathrm{LogU}[10^{-5},10^{-2}]$ \\
        KL weight & $\mathrm{LogU}[1,10]$ \\
        Action chunk size & $\{2,8\}$ \\
        Action horizon & $\{1,4\}$ \\
        Weight decay & $\mathrm{LogU}[10^{-6},10^{-2}]$ \\
        Gradient clip & $\mathrm{U}[0.1,2]$ \\
    }
    \hfill
    \hprangetile{VQ-BeT}{
        Learning rate & $\mathrm{LogU}[10^{-5},10^{-2}]$ \\
        Offset LR & $\mathrm{LogU}[10^{-5},10^{-2}]$ \\
        Action chunk size & $\{2,8\}$ \\
        Action horizon & $\{1,4\}$ \\
        VQ-VAE step ratio & $\mathrm{U}[0.1,0.4]$ \\
        Commitment weight & $\mathrm{LogU}[0.01,1]$ \\
        Focal $\gamma$ & $\mathrm{U}[0.5,5]$ \\
        Offset-loss weight & $\mathrm{U}[0.1,5]$ \\
        Primary-code weight & $\mathrm{U}[1,10]$ \\
        Secondary-code weight & $\mathrm{U}[0.1,3]$ \\
        Weight decay & $\mathrm{LogU}[10^{-6},10^{-2}]$ \\
        Gradient clip & $\mathrm{U}[0.1,2]$ \\
    }

    \par\medskip

    \hfill
    \hprangetile{Diffusion Policy}{
        Learning rate & $\mathrm{LogU}[10^{-5},10^{-3}]$ \\
        Diffusion steps & $\{50,100\}$ \\
        Inference steps & $\mathrm{Int}[20,100]$ \\
        Action chunk size & $\{8,16\}$ \\
        Action horizon & $\{1,4\}$ \\
        Weight decay & $\mathrm{LogU}[10^{-6},10^{-2}]$ \\
        Gradient clip & $\mathrm{U}[0.1,2]$ \\
    }
    \hfill\null

    \endgroup
\end{table}

\begin{table}[t]
    \centering
    \caption{Hyperparameter search spaces for discrete-control algorithms. Fixed parameters are omitted.}
    \label{tab:hp-ranges-discrete}

    \begingroup
    \scriptsize
    \renewcommand{\arraystretch}{0.92}
    \setlength{\tabcolsep}{2pt}

    \emph{Notation:} $\mathrm{U}$ is uniform, $\mathrm{LogU}$ is log-uniform, $\mathrm{Int}$ is inclusive integer, and braces denote uniformly sampled choices.
    \vspace{4pt}

    \hprangetile{BC}{
        Learning rate & $\mathrm{LogU}[10^{-5},10^{-2}]$ \\
        Weight decay & $\mathrm{LogU}[10^{-4},10^{-1}]$ \\
        Gradient clip & $\mathrm{U}[0.1,2]$ \\
        Dropout & $\mathrm{U}[0,0.3]$ \\
        Hidden units & $\{128,256,512\}\times2$ \\
        Regularization & $\mathrm{U}[0,1]$ \\
    }
    \hfill
    \hprangetile{BCP}{
        Learning rate & $\mathrm{LogU}[10^{-5},10^{-2}]$ \\
        Weight decay & $\mathrm{LogU}[10^{-4},10^{-1}]$ \\
        Gradient clip & $\mathrm{U}[0.1,2]$ \\
        Dropout & $\mathrm{U}[0,0.3]$ \\
        Hidden units & $\{128,256,512\}\times2$ \\
        Train percentile & $\mathrm{U}[0,90]$ \\
        Regularization & $\mathrm{U}[0,1]$ \\
    }
    \hfill
    \hprangetile{BCQ}{
        Learning rate & $\mathrm{LogU}[10^{-5},10^{-2}]$ \\
        Discount $\gamma$ & $\mathrm{LogU}[0.9,0.999]$ \\
        Critics & $\mathrm{Int}[1,4]$ \\
        Target interval & $\mathrm{Int}[2000,20000]$ \\
        Action flexibility & $\mathrm{U}[0.01,0.1]$ \\
        $\beta$ & $\mathrm{U}[0.1,0.9]$ \\
    }

    \par\medskip

    \hfill
    \hprangetile{CQL}{
        Learning rate & $\mathrm{LogU}[10^{-5},10^{-2}]$ \\
        Discount $\gamma$ & $\mathrm{LogU}[0.9,0.999]$ \\
        Critics & $\mathrm{Int}[1,4]$ \\
        Target interval & $\mathrm{Int}[1000,10000]$ \\
        $\alpha$ & $\mathrm{U}[0.1,2]$ \\
    }
    \hfill
    \hprangetile{DT}{
        Learning rate & $\mathrm{LogU}[10^{-5},10^{-2}]$ \\
        Weight decay & $\mathrm{LogU}[10^{-4},10^{-1}]$ \\
        Gradient clip & $\mathrm{U}[0.1,2]$ \\
        Context size & $\{10,20,30\}$ \\
        Hidden units & $\{128,256\}\times3$ \\
        Dropout & $\mathrm{U}[0,0.3]$ \\
        Attention heads & $\{1,2,4\}$ \\
        Layers & $\mathrm{Int}[2,5]$ \\
        Discount $\gamma$ & $\mathrm{LogU}[0.9,0.999]$ \\
        Reward scale & $\{0,1\}$ \\
    }
    \hfill\null

    \endgroup
\end{table}

\begin{table}[t!]
  \centering
  \caption{Normalized scores on continuous-control datasets, grouped by benchmark collection. Entries report the point estimate $\pm$ half-width of its 95\% confidence interval. Dataset rows summarize the top 25\% of evaluated hyperparameter configurations; shaded rows report collection-level interquartile mean (IQM; $\uparrow$) and optimality gap ($\downarrow$). The best point estimate in each row is bold.}
  \label{tab:continuous-leaderboard}
  \begingroup
  \footnotesize
  \setlength{\tabcolsep}{1.55pt}
  \renewcommand{\arraystretch}{0.68}
  \begin{adjustbox}{max width=\textwidth,max totalheight=0.84\textheight,center}
  \begin{tabular}{@{}l*{10}{r}@{}}
    \toprule
    Dataset & \textbf{BC} & \textbf{BCP} & \textbf{BCQ} & \textbf{CQL} & \textbf{DT} & \textbf{IQL} & \textbf{ReBRAC} & \textbf{ACT} & \textbf{VQ-BeT} & \textbf{Diff.} \\
    \midrule
    \multicolumn{11}{@{}l}{\textbf{AntMaze} \hfill \textit{6 datasets}} \\[-0.35ex]
    \cmidrule(lr){1-11}
    large-diverse & 0.004{\tiny\,$\pm$0.005} & 0.000{\tiny\,$\pm$0.000} & 0.288{\tiny\,$\pm$0.017} & 0.000{\tiny\,$\pm$0.001} & 0.454{\tiny\,$\pm$0.024} & 0.322{\tiny\,$\pm$0.024} & 0.464{\tiny\,$\pm$0.026} & \textbf{0.804}{\tiny\,$\pm$0.019} & 0.113{\tiny\,$\pm$0.026} & 0.187{\tiny\,$\pm$0.017} \\
    large-play & 0.007{\tiny\,$\pm$0.007} & 0.007{\tiny\,$\pm$0.004} & 0.295{\tiny\,$\pm$0.013} & 0.000{\tiny\,$\pm$0.000} & 0.297{\tiny\,$\pm$0.013} & 0.397{\tiny\,$\pm$0.010} & 0.421{\tiny\,$\pm$0.019} & \textbf{0.745}{\tiny\,$\pm$0.020} & 0.100{\tiny\,$\pm$0.009} & 0.130{\tiny\,$\pm$0.022} \\
    medium-diverse & 0.000{\tiny\,$\pm$0.000} & 0.020{\tiny\,$\pm$0.013} & 0.245{\tiny\,$\pm$0.015} & 0.000{\tiny\,$\pm$0.000} & 0.449{\tiny\,$\pm$0.034} & 0.367{\tiny\,$\pm$0.015} & 0.610{\tiny\,$\pm$0.033} & \textbf{0.849}{\tiny\,$\pm$0.023} & 0.192{\tiny\,$\pm$0.030} & 0.327{\tiny\,$\pm$0.017} \\
    medium-play & 0.115{\tiny\,$\pm$0.035} & 0.194{\tiny\,$\pm$0.035} & 0.479{\tiny\,$\pm$0.018} & 0.002{\tiny\,$\pm$0.002} & 0.325{\tiny\,$\pm$0.014} & 0.545{\tiny\,$\pm$0.011} & 0.556{\tiny\,$\pm$0.016} & \textbf{0.883}{\tiny\,$\pm$0.009} & 0.446{\tiny\,$\pm$0.017} & 0.372{\tiny\,$\pm$0.021} \\
    umaze-diverse & 0.764{\tiny\,$\pm$0.031} & 0.762{\tiny\,$\pm$0.023} & 0.935{\tiny\,$\pm$0.010} & 0.012{\tiny\,$\pm$0.007} & \textbf{1.003}{\tiny\,$\pm$0.005} & 0.970{\tiny\,$\pm$0.005} & 0.940{\tiny\,$\pm$0.009} & 0.975{\tiny\,$\pm$0.005} & 0.795{\tiny\,$\pm$0.025} & 0.941{\tiny\,$\pm$0.009} \\
    umaze & 0.927{\tiny\,$\pm$0.012} & 0.916{\tiny\,$\pm$0.019} & 0.933{\tiny\,$\pm$0.025} & 0.010{\tiny\,$\pm$0.010} & \textbf{1.045}{\tiny\,$\pm$0.004} & 1.041{\tiny\,$\pm$0.004} & 1.008{\tiny\,$\pm$0.008} & 0.983{\tiny\,$\pm$0.009} & 0.976{\tiny\,$\pm$0.009} & 0.756{\tiny\,$\pm$0.032} \\
    \cmidrule(lr){1-11}
    \rowcolor[gray]{0.94}\textit{IQM} $\uparrow$ & 0.222{\tiny\,$\pm$0.340} & 0.246{\tiny\,$\pm$0.345} & 0.499{\tiny\,$\pm$0.272} & 0.003{\tiny\,$\pm$0.004} & 0.558{\tiny\,$\pm$0.272} & 0.570{\tiny\,$\pm$0.272} & 0.643{\tiny\,$\pm$0.213} & \textbf{0.878}{\tiny\,$\pm$0.083} & 0.387{\tiny\,$\pm$0.334} & 0.410{\tiny\,$\pm$0.279} \\
    \rowcolor[gray]{0.94}\textit{Opt. gap} $\downarrow$ & 0.697{\tiny\,$\pm$0.294} & 0.684{\tiny\,$\pm$0.291} & 0.471{\tiny\,$\pm$0.238} & 0.996{\tiny\,$\pm$0.004} & 0.412{\tiny\,$\pm$0.227} & 0.400{\tiny\,$\pm$0.216} & 0.335{\tiny\,$\pm$0.172} & \textbf{0.127}{\tiny\,$\pm$0.068} & 0.563{\tiny\,$\pm$0.268} & 0.548{\tiny\,$\pm$0.237} \\
    \midrule
    \multicolumn{11}{@{}l}{\textbf{Adroit} \hfill \textit{12 datasets}} \\[-0.35ex]
    \cmidrule(lr){1-11}
    door-cloned & 0.253{\tiny\,$\pm$0.029} & 0.711{\tiny\,$\pm$0.056} & 0.154{\tiny\,$\pm$0.056} & 0.044{\tiny\,$\pm$0.004} & \textbf{1.229}{\tiny\,$\pm$0.065} & 0.032{\tiny\,$\pm$0.003} & 0.045{\tiny\,$\pm$0.004} & 0.060{\tiny\,$\pm$0.021} & 0.022{\tiny\,$\pm$0.002} & 0.042{\tiny\,$\pm$0.009} \\
    door-expert & 1.036{\tiny\,$\pm$0.000} & 1.044{\tiny\,$\pm$0.001} & \textbf{1.049}{\tiny\,$\pm$0.002} & 0.198{\tiny\,$\pm$0.093} & 1.045{\tiny\,$\pm$0.000} & 1.040{\tiny\,$\pm$0.000} & 1.036{\tiny\,$\pm$0.002} & 1.037{\tiny\,$\pm$0.001} & 1.034{\tiny\,$\pm$0.001} & 0.837{\tiny\,$\pm$0.025} \\
    door-human & 0.075{\tiny\,$\pm$0.018} & 0.094{\tiny\,$\pm$0.018} & 0.013{\tiny\,$\pm$0.005} & 0.050{\tiny\,$\pm$0.021} & 0.041{\tiny\,$\pm$0.010} & 0.199{\tiny\,$\pm$0.010} & 0.018{\tiny\,$\pm$0.000} & \textbf{0.521}{\tiny\,$\pm$0.067} & 0.290{\tiny\,$\pm$0.035} & 0.011{\tiny\,$\pm$0.001} \\
    hammer-cloned & 0.134{\tiny\,$\pm$0.011} & 0.344{\tiny\,$\pm$0.043} & \textbf{0.454}{\tiny\,$\pm$0.099} & 0.084{\tiny\,$\pm$0.025} & 0.268{\tiny\,$\pm$0.018} & 0.141{\tiny\,$\pm$0.018} & 0.150{\tiny\,$\pm$0.051} & 0.075{\tiny\,$\pm$0.012} & 0.225{\tiny\,$\pm$0.028} & 0.133{\tiny\,$\pm$0.016} \\
    hammer-expert & 1.306{\tiny\,$\pm$0.001} & 1.313{\tiny\,$\pm$0.001} & \textbf{1.353}{\tiny\,$\pm$0.002} & 0.155{\tiny\,$\pm$0.074} & 1.311{\tiny\,$\pm$0.001} & 1.314{\tiny\,$\pm$0.001} & 1.315{\tiny\,$\pm$0.003} & 1.311{\tiny\,$\pm$0.001} & 1.303{\tiny\,$\pm$0.002} & 0.718{\tiny\,$\pm$0.039} \\
    hammer-human & 0.071{\tiny\,$\pm$0.015} & 0.048{\tiny\,$\pm$0.011} & \textbf{0.180}{\tiny\,$\pm$0.045} & 0.023{\tiny\,$\pm$0.009} & 0.073{\tiny\,$\pm$0.010} & 0.140{\tiny\,$\pm$0.010} & 0.047{\tiny\,$\pm$0.008} & 0.088{\tiny\,$\pm$0.016} & 0.137{\tiny\,$\pm$0.011} & 0.003{\tiny\,$\pm$0.000} \\
    pen-cloned & 1.463{\tiny\,$\pm$0.022} & \textbf{1.703}{\tiny\,$\pm$0.023} & 1.527{\tiny\,$\pm$0.022} & 0.393{\tiny\,$\pm$0.048} & 1.398{\tiny\,$\pm$0.021} & 1.335{\tiny\,$\pm$0.018} & 1.606{\tiny\,$\pm$0.041} & 1.397{\tiny\,$\pm$0.016} & 1.308{\tiny\,$\pm$0.015} & 1.190{\tiny\,$\pm$0.024} \\
    pen-expert & 2.420{\tiny\,$\pm$0.014} & 2.544{\tiny\,$\pm$0.016} & \textbf{2.592}{\tiny\,$\pm$0.020} & 2.180{\tiny\,$\pm$0.079} & 2.429{\tiny\,$\pm$0.013} & 2.536{\tiny\,$\pm$0.015} & 2.534{\tiny\,$\pm$0.025} & 2.398{\tiny\,$\pm$0.014} & 2.396{\tiny\,$\pm$0.020} & 1.941{\tiny\,$\pm$0.030} \\
    pen-human & 0.888{\tiny\,$\pm$0.016} & \textbf{0.900}{\tiny\,$\pm$0.017} & 0.884{\tiny\,$\pm$0.010} & 0.336{\tiny\,$\pm$0.030} & 0.828{\tiny\,$\pm$0.015} & 0.857{\tiny\,$\pm$0.015} & 0.785{\tiny\,$\pm$0.017} & 0.873{\tiny\,$\pm$0.019} & 0.842{\tiny\,$\pm$0.013} & 0.066{\tiny\,$\pm$0.004} \\
    relocate-cloned & 0.000{\tiny\,$\pm$0.000} & 0.002{\tiny\,$\pm$0.001} & 0.004{\tiny\,$\pm$0.001} & 0.009{\tiny\,$\pm$0.000} & 0.004{\tiny\,$\pm$0.000} & 0.001{\tiny\,$\pm$0.000} & \textbf{0.009}{\tiny\,$\pm$0.000} & 0.000{\tiny\,$\pm$0.000} & 0.000{\tiny\,$\pm$0.000} & 0.005{\tiny\,$\pm$0.000} \\
    relocate-expert & 1.049{\tiny\,$\pm$0.002} & \textbf{1.055}{\tiny\,$\pm$0.002} & 0.697{\tiny\,$\pm$0.024} & 0.166{\tiny\,$\pm$0.045} & 1.038{\tiny\,$\pm$0.002} & 1.042{\tiny\,$\pm$0.002} & 1.040{\tiny\,$\pm$0.007} & 1.039{\tiny\,$\pm$0.001} & 1.023{\tiny\,$\pm$0.002} & 0.475{\tiny\,$\pm$0.046} \\
    relocate-human & 0.000{\tiny\,$\pm$0.000} & -0.001{\tiny\,$\pm$0.000} & 0.000{\tiny\,$\pm$0.000} & 0.002{\tiny\,$\pm$0.000} & -0.001{\tiny\,$\pm$0.000} & 0.001{\tiny\,$\pm$0.000} & 0.002{\tiny\,$\pm$0.000} & \textbf{0.004}{\tiny\,$\pm$0.001} & 0.000{\tiny\,$\pm$0.000} & -0.001{\tiny\,$\pm$0.000} \\
    \cmidrule(lr){1-11}
    \rowcolor[gray]{0.94}\textit{IQM} $\uparrow$ & 0.573{\tiny\,$\pm$0.532} & 0.691{\tiny\,$\pm$0.516} & 0.570{\tiny\,$\pm$0.501} & 0.116{\tiny\,$\pm$0.115} & \textbf{0.747}{\tiny\,$\pm$0.514} & 0.570{\tiny\,$\pm$0.521} & 0.517{\tiny\,$\pm$0.563} & 0.605{\tiny\,$\pm$0.510} & 0.592{\tiny\,$\pm$0.497} & 0.241{\tiny\,$\pm$0.347} \\
    \rowcolor[gray]{0.94}\textit{Opt. gap} $\downarrow$ & 0.465{\tiny\,$\pm$0.244} & 0.409{\tiny\,$\pm$0.245} & 0.468{\tiny\,$\pm$0.234} & 0.795{\tiny\,$\pm$0.146} & \textbf{0.399}{\tiny\,$\pm$0.244} & 0.469{\tiny\,$\pm$0.250} & 0.495{\tiny\,$\pm$0.255} & 0.448{\tiny\,$\pm$0.243} & 0.457{\tiny\,$\pm$0.250} & 0.643{\tiny\,$\pm$0.228} \\
    \midrule
    \multicolumn{11}{@{}l}{\textbf{Kitchen} \hfill \textit{3 datasets}} \\[-0.35ex]
    \cmidrule(lr){1-11}
    complete & 0.885{\tiny\,$\pm$0.054} & 0.841{\tiny\,$\pm$0.062} & 0.028{\tiny\,$\pm$0.011} & 0.064{\tiny\,$\pm$0.052} & \textbf{1.028}{\tiny\,$\pm$0.048} & 0.221{\tiny\,$\pm$0.021} & 0.175{\tiny\,$\pm$0.075} & 0.917{\tiny\,$\pm$0.060} & 0.910{\tiny\,$\pm$0.046} & 0.067{\tiny\,$\pm$0.009} \\
    mixed & 1.740{\tiny\,$\pm$0.095} & 1.974{\tiny\,$\pm$0.068} & 0.120{\tiny\,$\pm$0.053} & 0.211{\tiny\,$\pm$0.119} & 2.065{\tiny\,$\pm$0.056} & 1.942{\tiny\,$\pm$0.056} & 0.091{\tiny\,$\pm$0.076} & \textbf{2.521}{\tiny\,$\pm$0.054} & 1.921{\tiny\,$\pm$0.127} & 0.440{\tiny\,$\pm$0.083} \\
    partial & 2.581{\tiny\,$\pm$0.079} & 2.262{\tiny\,$\pm$0.104} & 0.334{\tiny\,$\pm$0.083} & 0.242{\tiny\,$\pm$0.154} & 2.634{\tiny\,$\pm$0.106} & 2.436{\tiny\,$\pm$0.063} & 0.405{\tiny\,$\pm$0.172} & \textbf{2.949}{\tiny\,$\pm$0.081} & 2.357{\tiny\,$\pm$0.096} & 0.586{\tiny\,$\pm$0.069} \\
    \cmidrule(lr){1-11}
    \rowcolor[gray]{0.94}\textit{IQM} $\uparrow$ & 1.735{\tiny\,$\pm$0.848} & 1.692{\tiny\,$\pm$0.710} & 0.161{\tiny\,$\pm$0.153} & 0.172{\tiny\,$\pm$0.089} & 1.909{\tiny\,$\pm$0.803} & 1.533{\tiny\,$\pm$1.107} & 0.224{\tiny\,$\pm$0.157} & \textbf{2.129}{\tiny\,$\pm$1.016} & 1.729{\tiny\,$\pm$0.724} & 0.364{\tiny\,$\pm$0.260} \\
    \rowcolor[gray]{0.94}\textit{Opt. gap} $\downarrow$ & 0.038{\tiny\,$\pm$0.057} & 0.053{\tiny\,$\pm$0.080} & 0.839{\tiny\,$\pm$0.153} & 0.828{\tiny\,$\pm$0.089} & \textbf{0.000}{\tiny\,$\pm$0.000} & 0.260{\tiny\,$\pm$0.389} & 0.776{\tiny\,$\pm$0.157} & 0.028{\tiny\,$\pm$0.041} & 0.030{\tiny\,$\pm$0.045} & 0.636{\tiny\,$\pm$0.260} \\
    \midrule
    \multicolumn{11}{@{}l}{\textbf{PointMaze} \hfill \textit{8 datasets}} \\[-0.35ex]
    \cmidrule(lr){1-11}
    large-dense & 0.104{\tiny\,$\pm$0.011} & \textbf{0.202}{\tiny\,$\pm$0.019} & 0.156{\tiny\,$\pm$0.026} & 0.125{\tiny\,$\pm$0.021} & 0.156{\tiny\,$\pm$0.030} & 0.079{\tiny\,$\pm$0.018} & 0.116{\tiny\,$\pm$0.023} & 0.111{\tiny\,$\pm$0.007} & 0.082{\tiny\,$\pm$0.009} & 0.004{\tiny\,$\pm$0.011} \\
    large & 0.024{\tiny\,$\pm$0.003} & 0.022{\tiny\,$\pm$0.002} & 0.020{\tiny\,$\pm$0.002} & 0.003{\tiny\,$\pm$0.002} & 0.023{\tiny\,$\pm$0.003} & 0.025{\tiny\,$\pm$0.002} & 0.022{\tiny\,$\pm$0.004} & 0.024{\tiny\,$\pm$0.002} & 0.055{\tiny\,$\pm$0.001} & \textbf{0.131}{\tiny\,$\pm$0.030} \\
    medium-dense & 0.109{\tiny\,$\pm$0.017} & 0.215{\tiny\,$\pm$0.025} & \textbf{0.252}{\tiny\,$\pm$0.032} & 0.235{\tiny\,$\pm$0.023} & 0.140{\tiny\,$\pm$0.028} & 0.112{\tiny\,$\pm$0.018} & 0.126{\tiny\,$\pm$0.026} & 0.083{\tiny\,$\pm$0.018} & 0.056{\tiny\,$\pm$0.011} & 0.051{\tiny\,$\pm$0.011} \\
    medium & 0.024{\tiny\,$\pm$0.004} & 0.021{\tiny\,$\pm$0.003} & 0.054{\tiny\,$\pm$0.004} & -0.003{\tiny\,$\pm$0.005} & 0.014{\tiny\,$\pm$0.003} & 0.065{\tiny\,$\pm$0.005} & 0.033{\tiny\,$\pm$0.005} & \textbf{0.077}{\tiny\,$\pm$0.015} & 0.058{\tiny\,$\pm$0.005} & 0.053{\tiny\,$\pm$0.006} \\
    open-dense & 0.452{\tiny\,$\pm$0.013} & 0.403{\tiny\,$\pm$0.020} & \textbf{0.576}{\tiny\,$\pm$0.027} & 0.180{\tiny\,$\pm$0.042} & 0.154{\tiny\,$\pm$0.043} & 0.356{\tiny\,$\pm$0.027} & 0.243{\tiny\,$\pm$0.042} & 0.546{\tiny\,$\pm$0.056} & 0.174{\tiny\,$\pm$0.044} & -0.045{\tiny\,$\pm$0.024} \\
    open & 0.095{\tiny\,$\pm$0.012} & 0.073{\tiny\,$\pm$0.012} & 0.151{\tiny\,$\pm$0.013} & -0.009{\tiny\,$\pm$0.013} & -0.006{\tiny\,$\pm$0.008} & \textbf{0.166}{\tiny\,$\pm$0.011} & 0.112{\tiny\,$\pm$0.012} & 0.043{\tiny\,$\pm$0.013} & 0.123{\tiny\,$\pm$0.003} & 0.095{\tiny\,$\pm$0.006} \\
    umaze-dense & 0.441{\tiny\,$\pm$0.026} & 0.504{\tiny\,$\pm$0.013} & \textbf{0.770}{\tiny\,$\pm$0.049} & 0.711{\tiny\,$\pm$0.051} & 0.375{\tiny\,$\pm$0.055} & 0.548{\tiny\,$\pm$0.024} & 0.587{\tiny\,$\pm$0.049} & 0.459{\tiny\,$\pm$0.006} & 0.359{\tiny\,$\pm$0.027} & 0.257{\tiny\,$\pm$0.015} \\
    umaze & 0.510{\tiny\,$\pm$0.033} & 0.517{\tiny\,$\pm$0.036} & \textbf{0.664}{\tiny\,$\pm$0.003} & 0.045{\tiny\,$\pm$0.041} & 0.358{\tiny\,$\pm$0.033} & 0.622{\tiny\,$\pm$0.006} & 0.514{\tiny\,$\pm$0.039} & 0.609{\tiny\,$\pm$0.003} & 0.561{\tiny\,$\pm$0.024} & 0.377{\tiny\,$\pm$0.015} \\
    \cmidrule(lr){1-11}
    \rowcolor[gray]{0.94}\textit{IQM} $\uparrow$ & 0.188{\tiny\,$\pm$0.192} & 0.223{\tiny\,$\pm$0.197} & \textbf{0.283}{\tiny\,$\pm$0.262} & 0.088{\tiny\,$\pm$0.156} & 0.118{\tiny\,$\pm$0.130} & 0.178{\tiny\,$\pm$0.191} & 0.149{\tiny\,$\pm$0.171} & 0.183{\tiny\,$\pm$0.224} & 0.109{\tiny\,$\pm$0.128} & 0.083{\tiny\,$\pm$0.104} \\
    \rowcolor[gray]{0.94}\textit{Opt. gap} $\downarrow$ & 0.780{\tiny\,$\pm$0.134} & 0.755{\tiny\,$\pm$0.136} & \textbf{0.670}{\tiny\,$\pm$0.190} & 0.839{\tiny\,$\pm$0.148} & 0.848{\tiny\,$\pm$0.094} & 0.753{\tiny\,$\pm$0.148} & 0.781{\tiny\,$\pm$0.140} & 0.756{\tiny\,$\pm$0.159} & 0.816{\tiny\,$\pm$0.116} & 0.885{\tiny\,$\pm$0.088} \\
    \midrule
    \multicolumn{11}{@{}l}{\textbf{MuJoCo} \hfill \textit{28 datasets}} \\[-0.35ex]
    \cmidrule(lr){1-11}
    ant-expert & 1.005{\tiny\,$\pm$0.002} & 1.003{\tiny\,$\pm$0.002} & 0.997{\tiny\,$\pm$0.002} & 0.319{\tiny\,$\pm$0.079} & 1.005{\tiny\,$\pm$0.001} & \textbf{1.012}{\tiny\,$\pm$0.001} & 0.992{\tiny\,$\pm$0.003} & 1.008{\tiny\,$\pm$0.002} & 0.974{\tiny\,$\pm$0.015} & 0.929{\tiny\,$\pm$0.019} \\
    ant-medium & 1.042{\tiny\,$\pm$0.004} & 1.029{\tiny\,$\pm$0.005} & 1.035{\tiny\,$\pm$0.004} & 0.413{\tiny\,$\pm$0.068} & 1.060{\tiny\,$\pm$0.003} & \textbf{1.069}{\tiny\,$\pm$0.002} & 0.989{\tiny\,$\pm$0.013} & 1.057{\tiny\,$\pm$0.002} & 0.935{\tiny\,$\pm$0.042} & 0.497{\tiny\,$\pm$0.066} \\
    ant-simple & 1.100{\tiny\,$\pm$0.004} & 1.094{\tiny\,$\pm$0.006} & \textbf{1.153}{\tiny\,$\pm$0.008} & 0.464{\tiny\,$\pm$0.087} & 1.109{\tiny\,$\pm$0.003} & 1.100{\tiny\,$\pm$0.003} & 1.079{\tiny\,$\pm$0.012} & 1.070{\tiny\,$\pm$0.005} & 1.034{\tiny\,$\pm$0.014} & 0.739{\tiny\,$\pm$0.038} \\
    halfcheetah-expert & 0.960{\tiny\,$\pm$0.016} & 0.862{\tiny\,$\pm$0.038} & 0.910{\tiny\,$\pm$0.034} & 0.040{\tiny\,$\pm$0.010} & 1.005{\tiny\,$\pm$0.007} & 1.051{\tiny\,$\pm$0.001} & 0.893{\tiny\,$\pm$0.051} & 1.062{\tiny\,$\pm$0.002} & \textbf{1.073}{\tiny\,$\pm$0.002} & 0.347{\tiny\,$\pm$0.020} \\
    halfcheetah-medium & 1.187{\tiny\,$\pm$0.012} & 1.144{\tiny\,$\pm$0.025} & 0.954{\tiny\,$\pm$0.021} & 0.074{\tiny\,$\pm$0.034} & 1.098{\tiny\,$\pm$0.011} & \textbf{1.259}{\tiny\,$\pm$0.002} & 0.951{\tiny\,$\pm$0.043} & 1.161{\tiny\,$\pm$0.018} & 1.203{\tiny\,$\pm$0.030} & 0.197{\tiny\,$\pm$0.015} \\
    halfcheetah-simple & 1.040{\tiny\,$\pm$0.001} & 1.040{\tiny\,$\pm$0.001} & \textbf{1.049}{\tiny\,$\pm$0.003} & 0.348{\tiny\,$\pm$0.107} & 1.033{\tiny\,$\pm$0.001} & 1.045{\tiny\,$\pm$0.000} & 1.048{\tiny\,$\pm$0.005} & 1.044{\tiny\,$\pm$0.001} & 1.035{\tiny\,$\pm$0.001} & 0.968{\tiny\,$\pm$0.006} \\
    hopper-expert & 0.935{\tiny\,$\pm$0.016} & 0.816{\tiny\,$\pm$0.025} & 0.644{\tiny\,$\pm$0.034} & 0.525{\tiny\,$\pm$0.034} & 0.831{\tiny\,$\pm$0.016} & \textbf{1.070}{\tiny\,$\pm$0.001} & 0.776{\tiny\,$\pm$0.036} & 0.950{\tiny\,$\pm$0.011} & 1.045{\tiny\,$\pm$0.003} & 0.810{\tiny\,$\pm$0.025} \\
    hopper-medium & 1.249{\tiny\,$\pm$0.007} & 1.255{\tiny\,$\pm$0.005} & 1.226{\tiny\,$\pm$0.017} & 1.238{\tiny\,$\pm$0.010} & 1.199{\tiny\,$\pm$0.015} & 1.267{\tiny\,$\pm$0.001} & 1.046{\tiny\,$\pm$0.063} & \textbf{1.269}{\tiny\,$\pm$0.001} & 1.243{\tiny\,$\pm$0.007} & 1.068{\tiny\,$\pm$0.024} \\
    hopper-simple & 1.935{\tiny\,$\pm$0.002} & \textbf{1.940}{\tiny\,$\pm$0.002} & 1.819{\tiny\,$\pm$0.057} & 1.215{\tiny\,$\pm$0.188} & 1.929{\tiny\,$\pm$0.001} & 1.936{\tiny\,$\pm$0.001} & 1.921{\tiny\,$\pm$0.008} & 1.927{\tiny\,$\pm$0.002} & 1.860{\tiny\,$\pm$0.047} & 1.154{\tiny\,$\pm$0.063} \\
    humanoid-expert & 1.149{\tiny\,$\pm$0.012} & 1.120{\tiny\,$\pm$0.016} & 1.004{\tiny\,$\pm$0.056} & 0.030{\tiny\,$\pm$0.008} & 1.229{\tiny\,$\pm$0.002} & \textbf{1.242}{\tiny\,$\pm$0.001} & 1.185{\tiny\,$\pm$0.011} & 1.089{\tiny\,$\pm$0.014} & 1.056{\tiny\,$\pm$0.022} & 0.237{\tiny\,$\pm$0.029} \\
    humanoid-medium & 1.035{\tiny\,$\pm$0.017} & 1.010{\tiny\,$\pm$0.020} & 1.117{\tiny\,$\pm$0.007} & 0.059{\tiny\,$\pm$0.009} & 1.138{\tiny\,$\pm$0.005} & \textbf{1.155}{\tiny\,$\pm$0.003} & 1.084{\tiny\,$\pm$0.014} & 1.149{\tiny\,$\pm$0.002} & 1.019{\tiny\,$\pm$0.056} & 0.414{\tiny\,$\pm$0.064} \\
    humanoid-simple & 0.574{\tiny\,$\pm$0.024} & 0.497{\tiny\,$\pm$0.032} & 0.803{\tiny\,$\pm$0.020} & 0.126{\tiny\,$\pm$0.034} & 0.895{\tiny\,$\pm$0.007} & 0.930{\tiny\,$\pm$0.008} & 0.774{\tiny\,$\pm$0.024} & \textbf{1.018}{\tiny\,$\pm$0.005} & 0.764{\tiny\,$\pm$0.067} & 0.602{\tiny\,$\pm$0.038} \\
    humanoidstandup-expert & 0.191{\tiny\,$\pm$0.008} & 0.207{\tiny\,$\pm$0.014} & 0.273{\tiny\,$\pm$0.020} & 0.107{\tiny\,$\pm$0.007} & 0.825{\tiny\,$\pm$0.014} & 0.785{\tiny\,$\pm$0.023} & 0.268{\tiny\,$\pm$0.014} & \textbf{0.956}{\tiny\,$\pm$0.008} & 0.499{\tiny\,$\pm$0.046} & 0.678{\tiny\,$\pm$0.035} \\
    humanoidstandup-medium & 0.495{\tiny\,$\pm$0.042} & 0.316{\tiny\,$\pm$0.031} & 0.887{\tiny\,$\pm$0.022} & 0.192{\tiny\,$\pm$0.020} & 0.977{\tiny\,$\pm$0.004} & \textbf{1.008}{\tiny\,$\pm$0.002} & 0.792{\tiny\,$\pm$0.034} & 0.958{\tiny\,$\pm$0.005} & 0.955{\tiny\,$\pm$0.028} & 0.948{\tiny\,$\pm$0.006} \\
    humanoidstandup-simple & 0.872{\tiny\,$\pm$0.011} & 0.875{\tiny\,$\pm$0.015} & 0.956{\tiny\,$\pm$0.008} & 0.823{\tiny\,$\pm$0.028} & 0.994{\tiny\,$\pm$0.003} & \textbf{1.007}{\tiny\,$\pm$0.003} & 0.952{\tiny\,$\pm$0.011} & 0.988{\tiny\,$\pm$0.004} & 0.970{\tiny\,$\pm$0.005} & 0.916{\tiny\,$\pm$0.009} \\
    inverteddoublependulum-expert & \textbf{1.000}{\tiny\,$\pm$0.000} & 1.000{\tiny\,$\pm$0.000} & 1.000{\tiny\,$\pm$0.000} & 0.591{\tiny\,$\pm$0.129} & 1.000{\tiny\,$\pm$0.000} & 1.000{\tiny\,$\pm$0.000} & 1.000{\tiny\,$\pm$0.000} & 1.000{\tiny\,$\pm$0.000} & 0.999{\tiny\,$\pm$0.001} & 0.079{\tiny\,$\pm$0.009} \\
    inverteddoublependulum-medium & 0.875{\tiny\,$\pm$0.021} & \textbf{1.241}{\tiny\,$\pm$0.000} & 1.235{\tiny\,$\pm$0.002} & 0.912{\tiny\,$\pm$0.054} & 0.977{\tiny\,$\pm$0.016} & 0.778{\tiny\,$\pm$0.023} & 1.086{\tiny\,$\pm$0.039} & 0.870{\tiny\,$\pm$0.026} & 0.914{\tiny\,$\pm$0.020} & 0.305{\tiny\,$\pm$0.042} \\
    invertedpendulum-expert & \textbf{1.000}{\tiny\,$\pm$0.000} & \textbf{1.000}{\tiny\,$\pm$0.000} & \textbf{1.000}{\tiny\,$\pm$0.000} & 0.036{\tiny\,$\pm$0.035} & \textbf{1.000}{\tiny\,$\pm$0.000} & \textbf{1.000}{\tiny\,$\pm$0.000} & \textbf{1.000}{\tiny\,$\pm$0.000} & \textbf{1.000}{\tiny\,$\pm$0.000} & \textbf{1.000}{\tiny\,$\pm$0.000} & 0.962{\tiny\,$\pm$0.008} \\
    invertedpendulum-medium & 1.587{\tiny\,$\pm$0.008} & 1.689{\tiny\,$\pm$0.018} & \textbf{11.130}{\tiny\,$\pm$0.000} & \textbf{11.130}{\tiny\,$\pm$0.000} & 1.457{\tiny\,$\pm$0.042} & \textbf{11.130}{\tiny\,$\pm$0.000} & \textbf{11.130}{\tiny\,$\pm$0.000} & 1.504{\tiny\,$\pm$0.018} & 1.181{\tiny\,$\pm$0.027} & 0.922{\tiny\,$\pm$0.037} \\
    pusher-expert & 1.015{\tiny\,$\pm$0.001} & 1.009{\tiny\,$\pm$0.002} & 1.003{\tiny\,$\pm$0.001} & 0.847{\tiny\,$\pm$0.006} & 1.019{\tiny\,$\pm$0.001} & \textbf{1.024}{\tiny\,$\pm$0.000} & 1.009{\tiny\,$\pm$0.003} & 1.023{\tiny\,$\pm$0.000} & 1.022{\tiny\,$\pm$0.001} & 0.968{\tiny\,$\pm$0.006} \\
    pusher-medium & 1.023{\tiny\,$\pm$0.001} & 1.023{\tiny\,$\pm$0.001} & 1.023{\tiny\,$\pm$0.001} & 0.969{\tiny\,$\pm$0.006} & \textbf{1.025}{\tiny\,$\pm$0.001} & 1.024{\tiny\,$\pm$0.001} & 1.017{\tiny\,$\pm$0.001} & 1.024{\tiny\,$\pm$0.001} & 1.020{\tiny\,$\pm$0.001} & 0.987{\tiny\,$\pm$0.003} \\
    reacher-expert & 1.029{\tiny\,$\pm$0.001} & 1.028{\tiny\,$\pm$0.001} & 1.029{\tiny\,$\pm$0.001} & 1.029{\tiny\,$\pm$0.001} & 1.026{\tiny\,$\pm$0.001} & \textbf{1.032}{\tiny\,$\pm$0.001} & 1.019{\tiny\,$\pm$0.001} & 1.030{\tiny\,$\pm$0.000} & 1.028{\tiny\,$\pm$0.001} & 1.006{\tiny\,$\pm$0.001} \\
    reacher-medium & 1.029{\tiny\,$\pm$0.001} & 1.028{\tiny\,$\pm$0.001} & \textbf{1.040}{\tiny\,$\pm$0.001} & 1.036{\tiny\,$\pm$0.001} & 1.031{\tiny\,$\pm$0.001} & 1.030{\tiny\,$\pm$0.001} & 1.034{\tiny\,$\pm$0.001} & 1.030{\tiny\,$\pm$0.001} & 1.027{\tiny\,$\pm$0.001} & 1.006{\tiny\,$\pm$0.001} \\
    swimmer-expert & \textbf{1.025}{\tiny\,$\pm$0.000} & 1.025{\tiny\,$\pm$0.000} & 1.020{\tiny\,$\pm$0.001} & 0.835{\tiny\,$\pm$0.067} & 1.015{\tiny\,$\pm$0.001} & 1.024{\tiny\,$\pm$0.000} & 1.023{\tiny\,$\pm$0.000} & 1.022{\tiny\,$\pm$0.000} & 1.022{\tiny\,$\pm$0.000} & 1.000{\tiny\,$\pm$0.001} \\
    swimmer-medium & 0.999{\tiny\,$\pm$0.003} & 1.094{\tiny\,$\pm$0.005} & 1.187{\tiny\,$\pm$0.017} & \textbf{1.198}{\tiny\,$\pm$0.019} & 1.059{\tiny\,$\pm$0.003} & 1.131{\tiny\,$\pm$0.006} & 1.121{\tiny\,$\pm$0.016} & 1.170{\tiny\,$\pm$0.012} & 1.140{\tiny\,$\pm$0.008} & 1.017{\tiny\,$\pm$0.009} \\
    walker2d-expert & 1.007{\tiny\,$\pm$0.002} & 0.973{\tiny\,$\pm$0.011} & 0.986{\tiny\,$\pm$0.004} & 0.369{\tiny\,$\pm$0.071} & 1.001{\tiny\,$\pm$0.002} & \textbf{1.015}{\tiny\,$\pm$0.000} & 0.983{\tiny\,$\pm$0.010} & 1.002{\tiny\,$\pm$0.001} & 1.007{\tiny\,$\pm$0.003} & 0.868{\tiny\,$\pm$0.023} \\
    walker2d-medium & 1.070{\tiny\,$\pm$0.002} & 1.075{\tiny\,$\pm$0.002} & 1.018{\tiny\,$\pm$0.006} & 0.180{\tiny\,$\pm$0.062} & 1.064{\tiny\,$\pm$0.002} & \textbf{1.082}{\tiny\,$\pm$0.001} & 1.060{\tiny\,$\pm$0.003} & 1.077{\tiny\,$\pm$0.002} & 1.071{\tiny\,$\pm$0.003} & 0.999{\tiny\,$\pm$0.006} \\
    walker2d-simple & 1.031{\tiny\,$\pm$0.001} & 1.033{\tiny\,$\pm$0.001} & 1.035{\tiny\,$\pm$0.001} & 0.908{\tiny\,$\pm$0.036} & \textbf{1.035}{\tiny\,$\pm$0.002} & 1.029{\tiny\,$\pm$0.001} & 1.032{\tiny\,$\pm$0.002} & 1.019{\tiny\,$\pm$0.001} & 1.019{\tiny\,$\pm$0.001} & 0.990{\tiny\,$\pm$0.005} \\
    \cmidrule(lr){1-11}
    \rowcolor[gray]{0.94}\textit{IQM} $\uparrow$ & 1.020{\tiny\,$\pm$0.036} & 1.028{\tiny\,$\pm$0.047} & 1.017{\tiny\,$\pm$0.041} & 0.552{\tiny\,$\pm$0.241} & 1.027{\tiny\,$\pm$0.029} & \textbf{1.043}{\tiny\,$\pm$0.037} & 1.018{\tiny\,$\pm$0.034} & 1.035{\tiny\,$\pm$0.030} & 1.024{\tiny\,$\pm$0.026} & 0.878{\tiny\,$\pm$0.131} \\
    \rowcolor[gray]{0.94}\textit{Opt. gap} $\downarrow$ & 0.075{\tiny\,$\pm$0.067} & 0.088{\tiny\,$\pm$0.076} & 0.057{\tiny\,$\pm$0.055} & 0.458{\tiny\,$\pm$0.139} & 0.018{\tiny\,$\pm$0.018} & 0.018{\tiny\,$\pm$0.021} & 0.058{\tiny\,$\pm$0.052} & \textbf{0.010}{\tiny\,$\pm$0.010} & 0.035{\tiny\,$\pm$0.036} & 0.237{\tiny\,$\pm$0.107} \\
    \bottomrule
  \end{tabular}
  \end{adjustbox}
  \endgroup
\end{table}

\begin{table}[t!]
  \centering
  \caption{Normalized scores on Atari environments. Entries report the point estimate $\pm$ half-width of its 95\% confidence interval. Environment rows summarize the top 25\% of evaluated hyperparameter configurations; shaded rows report the collection-level interquartile mean (IQM; $\uparrow$) and optimality gap ($\downarrow$). The best point estimate in each row is bold.}
  \label{tab:atari-leaderboard}
  \begingroup
  \small
  \setlength{\tabcolsep}{3.0pt}
  \renewcommand{\arraystretch}{0.80}
  \begin{adjustbox}{max width=\textwidth,max totalheight=0.84\textheight,center}
  \begin{tabular}{@{}l*{5}{r}@{}}
    \toprule
    Environment & \textbf{BC} & \textbf{BCP} & \textbf{BCQ} & \textbf{CQL} & \textbf{DT} \\
    \midrule
    \multicolumn{6}{@{}l}{\textbf{Atari} \hfill \textit{52 environments}} \\[-0.35ex]
    \cmidrule(lr){1-6}
    Alien & 1.201{\scriptsize\,$\pm$0.120} & 1.062{\scriptsize\,$\pm$0.132} & 0.261{\scriptsize\,$\pm$0.048} & \textbf{1.210}{\scriptsize\,$\pm$0.173} & 0.828{\scriptsize\,$\pm$0.010} \\
    Amidar & 0.938{\scriptsize\,$\pm$0.009} & \textbf{1.230}{\scriptsize\,$\pm$0.036} & 0.010{\scriptsize\,$\pm$0.004} & 1.156{\scriptsize\,$\pm$0.054} & 0.962{\scriptsize\,$\pm$0.016} \\
    Assault & 0.244{\scriptsize\,$\pm$0.006} & 0.229{\scriptsize\,$\pm$0.008} & 0.006{\scriptsize\,$\pm$0.001} & 0.165{\scriptsize\,$\pm$0.010} & \textbf{0.249}{\scriptsize\,$\pm$0.005} \\
    Asterix & 1.087{\scriptsize\,$\pm$0.107} & \textbf{2.580}{\scriptsize\,$\pm$0.128} & 1.133{\scriptsize\,$\pm$0.080} & 1.335{\scriptsize\,$\pm$0.115} & 0.760{\scriptsize\,$\pm$0.049} \\
    Asteroids & 0.763{\scriptsize\,$\pm$0.079} & 0.838{\scriptsize\,$\pm$0.065} & 0.271{\scriptsize\,$\pm$0.068} & 0.867{\scriptsize\,$\pm$0.075} & \textbf{1.075}{\scriptsize\,$\pm$0.009} \\
    Atlantis & 1.096{\scriptsize\,$\pm$0.001} & \textbf{1.101}{\scriptsize\,$\pm$0.003} & 0.011{\scriptsize\,$\pm$0.002} & 1.093{\scriptsize\,$\pm$0.002} & 1.089{\scriptsize\,$\pm$0.001} \\
    Bank Heist & 0.838{\scriptsize\,$\pm$0.053} & \textbf{1.309}{\scriptsize\,$\pm$0.058} & -0.009{\scriptsize\,$\pm$0.008} & 1.185{\scriptsize\,$\pm$0.044} & 1.077{\scriptsize\,$\pm$0.014} \\
    Battle Zone & 1.087{\scriptsize\,$\pm$0.054} & \textbf{1.146}{\scriptsize\,$\pm$0.084} & 0.123{\scriptsize\,$\pm$0.019} & 0.965{\scriptsize\,$\pm$0.065} & 0.798{\scriptsize\,$\pm$0.005} \\
    Beam Rider & 0.152{\scriptsize\,$\pm$0.021} & 0.112{\scriptsize\,$\pm$0.021} & 0.026{\scriptsize\,$\pm$0.005} & 0.086{\scriptsize\,$\pm$0.006} & \textbf{0.157}{\scriptsize\,$\pm$0.009} \\
    Berzerk & 0.268{\scriptsize\,$\pm$0.025} & 0.243{\scriptsize\,$\pm$0.022} & 0.039{\scriptsize\,$\pm$0.015} & \textbf{0.476}{\scriptsize\,$\pm$0.072} & 0.425{\scriptsize\,$\pm$0.007} \\
    Bowling & 1.444{\scriptsize\,$\pm$0.025} & \textbf{1.631}{\scriptsize\,$\pm$0.036} & -0.113{\scriptsize\,$\pm$0.173} & 1.502{\scriptsize\,$\pm$0.052} & 1.059{\scriptsize\,$\pm$0.008} \\
    Boxing & \textbf{1.049}{\scriptsize\,$\pm$0.003} & 1.049{\scriptsize\,$\pm$0.003} & 0.018{\scriptsize\,$\pm$0.007} & 1.031{\scriptsize\,$\pm$0.010} & 0.960{\scriptsize\,$\pm$0.002} \\
    Breakout & 0.180{\scriptsize\,$\pm$0.064} & \textbf{0.328}{\scriptsize\,$\pm$0.107} & 0.002{\scriptsize\,$\pm$0.001} & 0.182{\scriptsize\,$\pm$0.099} & 0.315{\scriptsize\,$\pm$0.029} \\
    Centipede & 1.013{\scriptsize\,$\pm$0.071} & \textbf{1.193}{\scriptsize\,$\pm$0.237} & 0.455{\scriptsize\,$\pm$0.131} & 1.131{\scriptsize\,$\pm$0.109} & 0.713{\scriptsize\,$\pm$0.025} \\
    Chopper Command & \textbf{1.046}{\scriptsize\,$\pm$0.086} & 1.022{\scriptsize\,$\pm$0.087} & 0.003{\scriptsize\,$\pm$0.005} & 0.881{\scriptsize\,$\pm$0.123} & 0.668{\scriptsize\,$\pm$0.017} \\
    Crazy Climber & \textbf{1.236}{\scriptsize\,$\pm$0.036} & 1.234{\scriptsize\,$\pm$0.028} & -0.064{\scriptsize\,$\pm$0.004} & 1.149{\scriptsize\,$\pm$0.029} & 1.097{\scriptsize\,$\pm$0.006} \\
    Defender & \textbf{0.917}{\scriptsize\,$\pm$0.025} & 0.856{\scriptsize\,$\pm$0.023} & 0.046{\scriptsize\,$\pm$0.014} & 0.813{\scriptsize\,$\pm$0.031} & 0.720{\scriptsize\,$\pm$0.008} \\
    Demon Attack & 0.048{\scriptsize\,$\pm$0.006} & 0.083{\scriptsize\,$\pm$0.041} & 0.001{\scriptsize\,$\pm$0.000} & \textbf{0.315}{\scriptsize\,$\pm$0.074} & 0.038{\scriptsize\,$\pm$0.002} \\
    Double Dunk & 1.004{\scriptsize\,$\pm$0.005} & 1.004{\scriptsize\,$\pm$0.005} & 1.000{\scriptsize\,$\pm$0.000} & \textbf{1.017}{\scriptsize\,$\pm$0.011} & 0.849{\scriptsize\,$\pm$0.013} \\
    Fishing Derby & \textbf{1.019}{\scriptsize\,$\pm$0.000} & \textbf{1.019}{\scriptsize\,$\pm$0.000} & 0.043{\scriptsize\,$\pm$0.007} & \textbf{1.019}{\scriptsize\,$\pm$0.000} & 0.902{\scriptsize\,$\pm$0.006} \\
    Freeway & \textbf{1.013}{\scriptsize\,$\pm$0.000} & \textbf{1.013}{\scriptsize\,$\pm$0.000} & 0.677{\scriptsize\,$\pm$0.021} & \textbf{1.013}{\scriptsize\,$\pm$0.000} & 0.992{\scriptsize\,$\pm$0.003} \\
    Frostbite & 2.127{\scriptsize\,$\pm$0.495} & \textbf{2.629}{\scriptsize\,$\pm$0.555} & 0.633{\scriptsize\,$\pm$0.023} & 1.034{\scriptsize\,$\pm$0.024} & 1.037{\scriptsize\,$\pm$0.010} \\
    Gopher & 0.975{\scriptsize\,$\pm$0.097} & 1.064{\scriptsize\,$\pm$0.090} & 0.064{\scriptsize\,$\pm$0.054} & \textbf{1.221}{\scriptsize\,$\pm$0.102} & 0.938{\scriptsize\,$\pm$0.009} \\
    Gravitar & 1.287{\scriptsize\,$\pm$0.207} & 1.265{\scriptsize\,$\pm$0.186} & 0.372{\scriptsize\,$\pm$0.133} & \textbf{1.369}{\scriptsize\,$\pm$0.217} & 0.940{\scriptsize\,$\pm$0.013} \\
    H.E.R.O. & 1.137{\scriptsize\,$\pm$0.003} & \textbf{1.141}{\scriptsize\,$\pm$0.001} & 0.021{\scriptsize\,$\pm$0.013} & 1.134{\scriptsize\,$\pm$0.002} & 1.030{\scriptsize\,$\pm$0.005} \\
    Ice Hockey & \textbf{0.801}{\scriptsize\,$\pm$0.021} & 0.674{\scriptsize\,$\pm$0.017} & 0.128{\scriptsize\,$\pm$0.061} & 0.748{\scriptsize\,$\pm$0.029} & 0.661{\scriptsize\,$\pm$0.008} \\
    James Bond & 0.643{\scriptsize\,$\pm$0.133} & 0.472{\scriptsize\,$\pm$0.100} & 0.015{\scriptsize\,$\pm$0.004} & 0.529{\scriptsize\,$\pm$0.113} & \textbf{0.659}{\scriptsize\,$\pm$0.019} \\
    Kangaroo & 0.801{\scriptsize\,$\pm$0.048} & \textbf{0.814}{\scriptsize\,$\pm$0.053} & 0.026{\scriptsize\,$\pm$0.008} & 0.456{\scriptsize\,$\pm$0.096} & 0.805{\scriptsize\,$\pm$0.015} \\
    Krull & 1.197{\scriptsize\,$\pm$0.012} & \textbf{1.205}{\scriptsize\,$\pm$0.012} & 0.239{\scriptsize\,$\pm$0.050} & 1.099{\scriptsize\,$\pm$0.017} & 1.025{\scriptsize\,$\pm$0.002} \\
    Kung-Fu Master & 0.990{\scriptsize\,$\pm$0.039} & 1.097{\scriptsize\,$\pm$0.040} & 0.019{\scriptsize\,$\pm$0.006} & \textbf{1.127}{\scriptsize\,$\pm$0.039} & 0.888{\scriptsize\,$\pm$0.005} \\
    Ms. Pac-Man & 0.935{\scriptsize\,$\pm$0.068} & 0.943{\scriptsize\,$\pm$0.077} & 0.132{\scriptsize\,$\pm$0.028} & \textbf{1.159}{\scriptsize\,$\pm$0.083} & 0.781{\scriptsize\,$\pm$0.012} \\
    Name This Game & 1.235{\scriptsize\,$\pm$0.032} & \textbf{1.285}{\scriptsize\,$\pm$0.046} & 0.096{\scriptsize\,$\pm$0.026} & 1.146{\scriptsize\,$\pm$0.047} & 0.898{\scriptsize\,$\pm$0.006} \\
    Phoenix & \textbf{0.193}{\scriptsize\,$\pm$0.015} & 0.157{\scriptsize\,$\pm$0.012} & 0.008{\scriptsize\,$\pm$0.004} & 0.170{\scriptsize\,$\pm$0.011} & 0.160{\scriptsize\,$\pm$0.006} \\
    Pitfall! & \textbf{1.023}{\scriptsize\,$\pm$0.000} & \textbf{1.023}{\scriptsize\,$\pm$0.000} & \textbf{1.023}{\scriptsize\,$\pm$0.000} & \textbf{1.023}{\scriptsize\,$\pm$0.000} & 0.992{\scriptsize\,$\pm$0.004} \\
    Pong & \textbf{1.873}{\scriptsize\,$\pm$0.001} & 1.872{\scriptsize\,$\pm$0.000} & -0.010{\scriptsize\,$\pm$0.009} & 0.305{\scriptsize\,$\pm$0.180} & 1.549{\scriptsize\,$\pm$0.036} \\
    Private Eye & 1.000{\scriptsize\,$\pm$0.000} & 1.000{\scriptsize\,$\pm$0.000} & \textbf{5.669}{\scriptsize\,$\pm$3.896} & 1.000{\scriptsize\,$\pm$0.000} & 2.505{\scriptsize\,$\pm$1.654} \\
    Q*bert & \textbf{1.074}{\scriptsize\,$\pm$0.018} & 1.074{\scriptsize\,$\pm$0.020} & 0.005{\scriptsize\,$\pm$0.002} & 0.984{\scriptsize\,$\pm$0.027} & 0.892{\scriptsize\,$\pm$0.005} \\
    River Raid & 0.852{\scriptsize\,$\pm$0.018} & 0.835{\scriptsize\,$\pm$0.026} & -0.003{\scriptsize\,$\pm$0.010} & \textbf{0.869}{\scriptsize\,$\pm$0.017} & 0.822{\scriptsize\,$\pm$0.006} \\
    Road Runner & \textbf{1.029}{\scriptsize\,$\pm$0.025} & 0.936{\scriptsize\,$\pm$0.025} & 0.002{\scriptsize\,$\pm$0.003} & 0.846{\scriptsize\,$\pm$0.038} & 0.924{\scriptsize\,$\pm$0.005} \\
    Robot Tank & \textbf{1.081}{\scriptsize\,$\pm$0.028} & 0.983{\scriptsize\,$\pm$0.028} & 0.130{\scriptsize\,$\pm$0.036} & 0.729{\scriptsize\,$\pm$0.024} & 0.811{\scriptsize\,$\pm$0.013} \\
    Seaquest & \textbf{1.028}{\scriptsize\,$\pm$0.000} & \textbf{1.028}{\scriptsize\,$\pm$0.000} & 0.085{\scriptsize\,$\pm$0.010} & 1.024{\scriptsize\,$\pm$0.002} & 1.014{\scriptsize\,$\pm$0.001} \\
    Space Invaders & 1.042{\scriptsize\,$\pm$0.107} & \textbf{1.362}{\scriptsize\,$\pm$0.095} & 0.330{\scriptsize\,$\pm$0.053} & 1.115{\scriptsize\,$\pm$0.080} & 0.894{\scriptsize\,$\pm$0.022} \\
    Star Gunner & 0.233{\scriptsize\,$\pm$0.019} & \textbf{0.706}{\scriptsize\,$\pm$0.131} & 0.005{\scriptsize\,$\pm$0.002} & 0.490{\scriptsize\,$\pm$0.081} & 0.327{\scriptsize\,$\pm$0.029} \\
    Surround & 0.861{\scriptsize\,$\pm$0.020} & \textbf{0.881}{\scriptsize\,$\pm$0.021} & 0.000{\scriptsize\,$\pm$0.001} & 0.564{\scriptsize\,$\pm$0.024} & 0.513{\scriptsize\,$\pm$0.022} \\
    Tennis & \textbf{0.644}{\scriptsize\,$\pm$0.003} & 0.642{\scriptsize\,$\pm$0.000} & 0.642{\scriptsize\,$\pm$0.000} & 0.643{\scriptsize\,$\pm$0.001} & 0.200{\scriptsize\,$\pm$0.061} \\
    Time Pilot & 0.416{\scriptsize\,$\pm$0.034} & \textbf{0.418}{\scriptsize\,$\pm$0.040} & 0.034{\scriptsize\,$\pm$0.007} & 0.231{\scriptsize\,$\pm$0.010} & 0.359{\scriptsize\,$\pm$0.011} \\
    Tutankham & 0.515{\scriptsize\,$\pm$0.096} & \textbf{1.051}{\scriptsize\,$\pm$0.107} & -0.087{\scriptsize\,$\pm$0.050} & 0.626{\scriptsize\,$\pm$0.107} & 0.991{\scriptsize\,$\pm$0.028} \\
    Up 'n Down & \textbf{0.981}{\scriptsize\,$\pm$0.018} & 0.939{\scriptsize\,$\pm$0.023} & 0.006{\scriptsize\,$\pm$0.003} & 0.714{\scriptsize\,$\pm$0.034} & 0.717{\scriptsize\,$\pm$0.014} \\
    Video Pinball & 0.154{\scriptsize\,$\pm$0.097} & 0.136{\scriptsize\,$\pm$0.081} & -0.039{\scriptsize\,$\pm$0.002} & -0.012{\scriptsize\,$\pm$0.016} & \textbf{0.189}{\scriptsize\,$\pm$0.017} \\
    Wizard of Wor & 3.104{\scriptsize\,$\pm$1.238} & 2.202{\scriptsize\,$\pm$1.061} & 0.433{\scriptsize\,$\pm$1.123} & \textbf{3.723}{\scriptsize\,$\pm$1.124} & 1.297{\scriptsize\,$\pm$0.089} \\
    Yars' Revenge & 0.709{\scriptsize\,$\pm$0.036} & 0.713{\scriptsize\,$\pm$0.045} & 0.009{\scriptsize\,$\pm$0.003} & 0.698{\scriptsize\,$\pm$0.039} & \textbf{0.719}{\scriptsize\,$\pm$0.009} \\
    Zaxxon & 0.545{\scriptsize\,$\pm$0.044} & 0.509{\scriptsize\,$\pm$0.048} & 0.002{\scriptsize\,$\pm$0.003} & \textbf{0.715}{\scriptsize\,$\pm$0.045} & 0.612{\scriptsize\,$\pm$0.006} \\
    \cmidrule(lr){1-6}
    \rowcolor[gray]{0.94}\textit{IQM} $\uparrow$ & 0.953{\scriptsize\,$\pm$0.092} & \textbf{0.986}{\scriptsize\,$\pm$0.099} & 0.052{\scriptsize\,$\pm$0.053} & 0.911{\scriptsize\,$\pm$0.117} & 0.840{\scriptsize\,$\pm$0.085} \\
    \rowcolor[gray]{0.94}\textit{Opt. gap} $\downarrow$ & 0.200{\scriptsize\,$\pm$0.080} & \textbf{0.184}{\scriptsize\,$\pm$0.076} & 0.825{\scriptsize\,$\pm$0.082} & 0.230{\scriptsize\,$\pm$0.085} & 0.249{\scriptsize\,$\pm$0.074} \\
    \bottomrule
  \end{tabular}
  \end{adjustbox}
  \endgroup
\end{table}


\begin{table}[t!]
\centering
\caption{Hyperparameter sensitivity in continuous control. Within each environment, $\rho$ is the Spearman correlation between the hyperparameter and trial score, and Share is its L1-normalized magnitude, $|\rho_h|/\sum_j|\rho_j|$. Entries report the median magnitude and share across 57 environments; rows are ordered by median $|\rho|$. Because each statistic is aggregated independently, the displayed shares need not sum to 100\%.}
\label{tab:hp-sensitivity-continuous}
\begingroup
\scriptsize
\setlength{\tabcolsep}{2.5pt}
\renewcommand{\arraystretch}{0.88}
\begin{minipage}[t]{0.485\linewidth}
\centering
\textbf{BC}\par\vspace{0.15em}
\begin{tabularx}{\linewidth}{@{}>{\raggedright\arraybackslash}X D{.}{.}{3} D{.}{.}{1}@{}}
\toprule
Hyperparameter & \multicolumn{1}{c}{$\widetilde{|\rho|}$} & \multicolumn{1}{c}{Share (\%)} \\
\midrule
Hidden units & 0.206 & 22.5 \\
Learning rate & 0.190 & 19.2 \\
Dropout & 0.171 & 19.8 \\
Weight decay & 0.127 & 12.1 \\
Policy type & 0.094 & 10.7 \\
Grad-clip norm & 0.092 & 8.7 \\
\bottomrule
\end{tabularx}
\end{minipage}
\hfill
\begin{minipage}[t]{0.485\linewidth}
\centering
\textbf{BCP}\par\vspace{0.15em}
\begin{tabularx}{\linewidth}{@{}>{\raggedright\arraybackslash}X D{.}{.}{3} D{.}{.}{1}@{}}
\toprule
Hyperparameter & \multicolumn{1}{c}{$\widetilde{|\rho|}$} & \multicolumn{1}{c}{Share (\%)} \\
\midrule
Train percentile & 0.245 & 17.1 \\
Hidden units & 0.233 & 19.4 \\
Learning rate & 0.202 & 13.8 \\
Dropout & 0.194 & 17.9 \\
Policy type & 0.138 & 9.2 \\
Weight decay & 0.104 & 10.2 \\
Grad-clip norm & 0.099 & 7.2 \\
\bottomrule
\end{tabularx}
\end{minipage}

\vspace{0.65em}

\begin{minipage}[t]{0.485\linewidth}
\centering
\textbf{BCQ}\par\vspace{0.15em}
\begin{tabularx}{\linewidth}{@{}>{\raggedright\arraybackslash}X D{.}{.}{3} D{.}{.}{1}@{}}
\toprule
Hyperparameter & \multicolumn{1}{c}{$\widetilde{|\rho|}$} & \multicolumn{1}{c}{Share (\%)} \\
\midrule
$\gamma$ & 0.173 & 13.7 \\
Critic LR & 0.151 & 11.2 \\
Imitator LR & 0.131 & 10.1 \\
$\lambda$ & 0.111 & 8.6 \\
Actor update interval & 0.096 & 7.3 \\
No. critics & 0.090 & 7.6 \\
Action samples & 0.087 & 6.3 \\
$\beta$ & 0.086 & 7.3 \\
Actor LR & 0.080 & 6.4 \\
$\tau$ & 0.079 & 6.4 \\
RL start step & 0.072 & 5.0 \\
\bottomrule
\end{tabularx}
\end{minipage}
\hfill
\begin{minipage}[t]{0.485\linewidth}
\centering
\textbf{CQL}\par\vspace{0.15em}
\begin{tabularx}{\linewidth}{@{}>{\raggedright\arraybackslash}X D{.}{.}{3} D{.}{.}{1}@{}}
\toprule
Hyperparameter & \multicolumn{1}{c}{$\widetilde{|\rho|}$} & \multicolumn{1}{c}{Share (\%)} \\
\midrule
$\alpha$ threshold & 0.204 & 15.3 \\
No. critics & 0.119 & 9.6 \\
$\alpha$ LR & 0.117 & 10.5 \\
Critic LR & 0.115 & 10.5 \\
$\gamma$ & 0.111 & 9.6 \\
Actor LR & 0.104 & 8.5 \\
Action samples & 0.073 & 6.8 \\
Conserv. weight & 0.068 & 6.2 \\
$\tau$ & 0.055 & 5.7 \\
Temp. LR & 0.050 & 4.9 \\
\bottomrule
\end{tabularx}
\end{minipage}

\vspace{0.65em}

\begin{minipage}[t]{0.485\linewidth}
\centering
\textbf{DT}\par\vspace{0.15em}
\begin{tabularx}{\linewidth}{@{}>{\raggedright\arraybackslash}X D{.}{.}{3} D{.}{.}{1}@{}}
\toprule
Hyperparameter & \multicolumn{1}{c}{$\widetilde{|\rho|}$} & \multicolumn{1}{c}{Share (\%)} \\
\midrule
Hidden units & 0.269 & 19.0 \\
Learning rate & 0.223 & 16.7 \\
Dropout (all) & 0.146 & 11.1 \\
Weight decay & 0.146 & 10.5 \\
Layers & 0.141 & 10.5 \\
$\gamma$ & 0.128 & 8.0 \\
Context size & 0.112 & 8.0 \\
Heads & 0.097 & 6.5 \\
Grad-clip norm & 0.076 & 5.1 \\
\bottomrule
\end{tabularx}
\end{minipage}
\hfill
\begin{minipage}[t]{0.485\linewidth}
\centering
\textbf{IQL}\par\vspace{0.15em}
\begin{tabularx}{\linewidth}{@{}>{\raggedright\arraybackslash}X D{.}{.}{3} D{.}{.}{1}@{}}
\toprule
Hyperparameter & \multicolumn{1}{c}{$\widetilde{|\rho|}$} & \multicolumn{1}{c}{Share (\%)} \\
\midrule
Actor LR & 0.256 & 32.5 \\
Expectile & 0.085 & 10.4 \\
No. critics & 0.082 & 10.2 \\
$\tau$ & 0.082 & 10.6 \\
Critic LR & 0.081 & 9.6 \\
Weight temp. & 0.081 & 11.3 \\
$\gamma$ & 0.072 & 9.3 \\
\bottomrule
\end{tabularx}
\end{minipage}

\vspace{0.65em}

\begin{minipage}[t]{0.485\linewidth}
\centering
\textbf{ReBRAC}\par\vspace{0.15em}
\begin{tabularx}{\linewidth}{@{}>{\raggedright\arraybackslash}X D{.}{.}{3} D{.}{.}{1}@{}}
\toprule
Hyperparameter & \multicolumn{1}{c}{$\widetilde{|\rho|}$} & \multicolumn{1}{c}{Share (\%)} \\
\midrule
Actor $\beta$ & 0.703 & 36.8 \\
$\gamma$ & 0.142 & 8.2 \\
Critic $\beta$ & 0.141 & 8.7 \\
No. critics & 0.134 & 8.0 \\
Critic LR & 0.134 & 8.1 \\
$\tau$ & 0.115 & 6.8 \\
Target smooth. $\sigma$ & 0.114 & 7.0 \\
Actor LR & 0.091 & 5.8 \\
Target smooth. clip & 0.091 & 5.8 \\
\bottomrule
\end{tabularx}
\end{minipage}
\hfill
\begin{minipage}[t]{0.485\linewidth}
\centering
\textbf{ACT}\par\vspace{0.15em}
\begin{tabularx}{\linewidth}{@{}>{\raggedright\arraybackslash}X D{.}{.}{3} D{.}{.}{1}@{}}
\toprule
Hyperparameter & \multicolumn{1}{c}{$\widetilde{|\rho|}$} & \multicolumn{1}{c}{Share (\%)} \\
\midrule
Learning rate & 0.525 & 36.1 \\
Chunk size & 0.238 & 17.0 \\
Action horizon & 0.174 & 15.0 \\
Grad-clip norm & 0.122 & 8.6 \\
KL weight & 0.113 & 9.2 \\
Weight decay & 0.108 & 8.9 \\
\bottomrule
\end{tabularx}
\end{minipage}

\vspace{0.65em}

\begin{minipage}[t]{0.485\linewidth}
\centering
\textbf{VQ-BeT}\par\vspace{0.15em}
\begin{tabularx}{\linewidth}{@{}>{\raggedright\arraybackslash}X D{.}{.}{3} D{.}{.}{1}@{}}
\toprule
Hyperparameter & \multicolumn{1}{c}{$\widetilde{|\rho|}$} & \multicolumn{1}{c}{Share (\%)} \\
\midrule
Learning rate & 0.380 & 18.6 \\
Offset LR & 0.265 & 14.5 \\
Action horizon & 0.256 & 12.0 \\
Chunk size & 0.188 & 8.4 \\
Focal $\gamma$ & 0.123 & 5.7 \\
VQ-VAE step ratio & 0.117 & 6.0 \\
Commitment weight & 0.107 & 5.6 \\
Weight decay & 0.104 & 4.9 \\
Offset loss weight & 0.101 & 4.6 \\
Grad-clip norm & 0.091 & 4.6 \\
Secondary code wt. & 0.087 & 4.4 \\
Primary code wt. & 0.081 & 5.1 \\
\bottomrule
\end{tabularx}
\end{minipage}
\hfill
\begin{minipage}[t]{0.485\linewidth}
\centering
\textbf{Diffusion}\par\vspace{0.15em}
\begin{tabularx}{\linewidth}{@{}>{\raggedright\arraybackslash}X D{.}{.}{3} D{.}{.}{1}@{}}
\toprule
Hyperparameter & \multicolumn{1}{c}{$\widetilde{|\rho|}$} & \multicolumn{1}{c}{Share (\%)} \\
\midrule
Diffusion steps & 0.489 & 27.6 \\
Inference steps & 0.376 & 21.3 \\
Learning rate & 0.178 & 10.5 \\
Action horizon & 0.155 & 10.8 \\
Chunk size & 0.129 & 9.9 \\
Grad-clip norm & 0.096 & 6.6 \\
Weight decay & 0.093 & 6.9 \\
\bottomrule
\end{tabularx}
\end{minipage}
\endgroup
\end{table}

\begin{table}[t!]
\centering
\caption{Hyperparameter sensitivity in discrete control (Atari). Within each environment, $\rho$ is the Spearman correlation between the hyperparameter and trial score, and Share is its L1-normalized magnitude, $|\rho_h|/\sum_j|\rho_j|$. Entries report the median magnitude and share across 54 environments; rows are ordered by median $|\rho|$. Because each statistic is aggregated independently, the displayed shares need not sum to 100\%.}
\label{tab:hp-sensitivity-discrete}
\begingroup
\scriptsize
\setlength{\tabcolsep}{2.5pt}
\renewcommand{\arraystretch}{0.88}
\begin{minipage}[t]{0.485\linewidth}
\centering
\textbf{BC}\par\vspace{0.15em}
\begin{tabularx}{\linewidth}{@{}>{\raggedright\arraybackslash}X D{.}{.}{3} D{.}{.}{1}@{}}
\toprule
Hyperparameter & \multicolumn{1}{c}{$\widetilde{|\rho|}$} & \multicolumn{1}{c}{Share (\%)} \\
\midrule
Learning rate & 0.233 & 25.7 \\
Dropout & 0.161 & 21.1 \\
Hidden units & 0.097 & 12.0 \\
Reg. factor & 0.092 & 11.0 \\
Weight decay & 0.084 & 10.4 \\
Grad-clip norm & 0.076 & 9.7 \\
\bottomrule
\end{tabularx}
\end{minipage}
\hfill
\begin{minipage}[t]{0.485\linewidth}
\centering
\textbf{BCP}\par\vspace{0.15em}
\begin{tabularx}{\linewidth}{@{}>{\raggedright\arraybackslash}X D{.}{.}{3} D{.}{.}{1}@{}}
\toprule
Hyperparameter & \multicolumn{1}{c}{$\widetilde{|\rho|}$} & \multicolumn{1}{c}{Share (\%)} \\
\midrule
Learning rate & 0.233 & 24.6 \\
Dropout & 0.156 & 17.3 \\
Train percentile & 0.127 & 13.3 \\
Reg. factor & 0.101 & 11.9 \\
Grad-clip norm & 0.087 & 9.9 \\
Hidden units & 0.079 & 10.2 \\
Weight decay & 0.077 & 8.9 \\
\bottomrule
\end{tabularx}
\end{minipage}

\vspace{0.65em}

\begin{minipage}[t]{0.485\linewidth}
\centering
\textbf{BCQ}\par\vspace{0.15em}
\begin{tabularx}{\linewidth}{@{}>{\raggedright\arraybackslash}X D{.}{.}{3} D{.}{.}{1}@{}}
\toprule
Hyperparameter & \multicolumn{1}{c}{$\widetilde{|\rho|}$} & \multicolumn{1}{c}{Share (\%)} \\
\midrule
Actor LR & 0.234 & 38.7 \\
Target update interval & 0.079 & 11.4 \\
$\beta$ & 0.079 & 12.4 \\
$\gamma$ & 0.078 & 9.4 \\
Action flexibility & 0.073 & 9.4 \\
No. critics & 0.066 & 9.6 \\
\bottomrule
\end{tabularx}
\end{minipage}
\hfill
\begin{minipage}[t]{0.485\linewidth}
\centering
\textbf{CQL}\par\vspace{0.15em}
\begin{tabularx}{\linewidth}{@{}>{\raggedright\arraybackslash}X D{.}{.}{3} D{.}{.}{1}@{}}
\toprule
Hyperparameter & \multicolumn{1}{c}{$\widetilde{|\rho|}$} & \multicolumn{1}{c}{Share (\%)} \\
\midrule
Actor LR & 0.655 & 56.0 \\
$\alpha$ & 0.143 & 12.2 \\
No. critics & 0.109 & 10.6 \\
Target update interval & 0.102 & 10.2 \\
$\gamma$ & 0.096 & 7.3 \\
\bottomrule
\end{tabularx}
\end{minipage}

\vspace{0.65em}

\makebox[\linewidth][c]{%
\begin{minipage}[t]{0.485\linewidth}
\centering
\textbf{DT}\par\vspace{0.15em}
\begin{tabularx}{\linewidth}{@{}>{\raggedright\arraybackslash}X D{.}{.}{3} D{.}{.}{1}@{}}
\toprule
Hyperparameter & \multicolumn{1}{c}{$\widetilde{|\rho|}$} & \multicolumn{1}{c}{Share (\%)} \\
\midrule
Learning rate & 0.465 & 31.2 \\
Hidden units & 0.191 & 12.5 \\
Dropout (all) & 0.143 & 9.6 \\
Context size & 0.113 & 8.3 \\
Heads & 0.101 & 7.3 \\
Layers & 0.092 & 7.8 \\
Grad-clip norm & 0.092 & 7.5 \\
$\gamma$ & 0.088 & 7.0 \\
Weight decay & 0.087 & 6.0 \\
\bottomrule
\end{tabularx}
\end{minipage}
}
\endgroup
\end{table}

\clearpage

\bibliographystyle{unsrtnat}
\bibliography{refs}

@misc{fu2021d4rldatasetsdeepdatadriven,
      title={D4RL: Datasets for Deep Data-Driven Reinforcement Learning}, 
      author={Justin Fu and Aviral Kumar and Ofir Nachum and George Tucker and Sergey Levine},
      year={2021},
      eprint={2004.07219},
      archivePrefix={arXiv},
      primaryClass={cs.LG},
      url={https://arxiv.org/abs/2004.07219}, 
}

@misc{fujimoto2019offpolicydeepreinforcementlearning,
      title={Off-Policy Deep Reinforcement Learning without Exploration}, 
      author={Scott Fujimoto and David Meger and Doina Precup},
      year={2019},
      eprint={1812.02900},
      archivePrefix={arXiv},
      primaryClass={cs.LG},
      url={https://arxiv.org/abs/1812.02900}, 
}

@misc{kumar2019stabilizingoffpolicyqlearningbootstrapping,
      title={Stabilizing Off-Policy Q-Learning via Bootstrapping Error Reduction}, 
      author={Aviral Kumar and Justin Fu and George Tucker and Sergey Levine},
      year={2019},
      eprint={1906.00949},
      archivePrefix={arXiv},
      primaryClass={cs.LG},
      url={https://arxiv.org/abs/1906.00949}, 
}

@misc{wu2019behaviorregularizedofflinereinforcement,
      title={Behavior Regularized Offline Reinforcement Learning}, 
      author={Yifan Wu and George Tucker and Ofir Nachum},
      year={2019},
      eprint={1911.11361},
      archivePrefix={arXiv},
      primaryClass={cs.LG},
      url={https://arxiv.org/abs/1911.11361}, 
}

@misc{mandlekar2021matterslearningofflinehuman,
      title={What Matters in Learning from Offline Human Demonstrations for Robot Manipulation}, 
      author={Ajay Mandlekar and Danfei Xu and Josiah Wong and Soroush Nasiriany and Chen Wang and Rohun Kulkarni and Li Fei-Fei and Silvio Savarese and Yuke Zhu and Roberto Martín-Martín},
      year={2021},
      eprint={2108.03298},
      archivePrefix={arXiv},
      primaryClass={cs.RO},
      url={https://arxiv.org/abs/2108.03298}, 
}

@misc{minari,
	author = {Younis, Omar G. and Perez-Vicente, Rodrigo and Balis, John U. and Dudley, Will and Davey, Alex and Terry, Jordan K},
	doi = {10.5281/zenodo.13767625},
	month = sep,
	publisher = {Zenodo},
	title = {Minari},
	url = {https://doi.org/10.5281/zenodo.13767625},
	version = {0.5.0},
	year = 2024
}

@article{lu2023challenges,
  title={Challenges and Opportunities in Offline Reinforcement Learning from Visual Observations}, 
  author={Cong Lu and Philip J. Ball and Tim G. J. Rudner and Jack Parker-Holder and Michael A. Osborne and Yee Whye Teh},
  year={2023},
  eprint={2206.04779},
  archivePrefix={arXiv},
  journal = {Transactions on Machine Learning Research},
  primaryClass={cs.LG},
  url={https://arxiv.org/abs/2206.04779}, 
}

@inproceedings{liu2023libero,
  title={{LIBERO}: Benchmarking Knowledge Transfer for Lifelong Robot Learning},
  author={Liu, Bo and Zhu, Yifeng and Gao, Chongkai and Feng, Yihao and Liu, Qiang and Zhu, Yuke and Stone, Peter},
  booktitle={Advances in Neural Information Processing Systems (NeurIPS) Datasets and Benchmarks Track},
  year={2023},
  note={arXiv:2306.03310}
}

@article{rafailov2024d5rl,
  title={{D5RL}: Diverse Datasets for Data-Driven Deep Reinforcement Learning},
  author={Rafailov, Rafael and Hatch, Kyle Beltran and Singh, Anikait and Smith, Laura and Kumar, Aviral and Kostrikov, Ilya and Hansen-Estruch, Philippe and Kolev, Victor and Ball, Philip J. and Wu, Jiajun and Finn, Chelsea and Levine, Sergey},
  journal={Reinforcement Learning Journal},
  volume={5},
  pages={2178--2197},
  year={2024},
  note={arXiv:2408.08441}
}

@inproceedings{park2025ogbench,
  title={OGBench: Benchmarking Offline Goal-Conditioned RL}, 
  author={Seohong Park and Kevin Frans and Benjamin Eysenbach and Sergey Levine},
  year={2025},
  eprint={2410.20092},
  booktitle = {International Conference on Learning Representations (ICLR)},
  archivePrefix={arXiv},
  primaryClass={cs.LG},
  url={https://arxiv.org/abs/2410.20092}, 
}

@inproceedings{rajeswaran2018learning,
  title={Learning Complex Dexterous Manipulation with Deep Reinforcement Learning and Demonstrations}, 
  author={Aravind Rajeswaran and Vikash Kumar and Abhishek Gupta and Giulia Vezzani and John Schulman and Emanuel Todorov and Sergey Levine},
  year={2018},
  eprint={1709.10087},
  archivePrefix={arXiv},
  primaryClass={cs.LG},
  url={https://arxiv.org/abs/1709.10087}, 
  booktitle = {Proceedings of Robotics: Science and Systems (RSS)},
}

@inproceedings{gupta2019relay,
  title={Relay Policy Learning: Solving Long-Horizon Tasks via Imitation and Reinforcement Learning},
  author={Gupta, Abhishek and Kumar, Vikash and Lynch, Corey and Levine, Sergey and Hausman, Karol},
  booktitle={Conference on Robot Learning (CoRL)},
  year={2019},
  note={arXiv:1910.11956}
}

@inproceedings{wu2017flow,
  title={Flow: Architecture and Benchmarking for Reinforcement Learning in Traffic Control},
  author={Wu, Cathy and Kreidieh, Aboudy and Parvate, Kanaad and Vinitsky, Eugene and Bayen, Alexandre M.},
  booktitle={arXiv preprint arXiv:1710.05465},
  year={2017}
}

@inproceedings{dosovitskiy2017carla,
  title={{CARLA}: An Open Urban Driving Simulator},
  author={Dosovitskiy, Alexey and Ros, German and Codevilla, Felipe and L{\'o}pez, Antonio and Koltun, Vladlen},
  booktitle={Conference on Robot Learning (CoRL)},
  year={2017}
}

@article{brockman2016openaigym,
  title={{OpenAI} {Gym}},
  author={Brockman, Greg and Cheung, Vicki and Pettersson, Ludwig and Schneider, Jonas and Schulman, John and Tang, Jie and Zaremba, Wojciech},
  journal={arXiv preprint arXiv:1606.01540},
  year={2016}
}

@inproceedings{todorov2012mujoco,
  title={{MuJoCo}: A Physics Engine for Model-Based Control},
  author={Todorov, Emanuel and Erez, Tom and Tassa, Yuval},
  booktitle={IEEE/RSJ International Conference on Intelligent Robots and Systems (IROS)},
  pages={5026--5033},
  year={2012}
}

@inproceedings{zhu2020robosuite,
  title={robosuite: A Modular Simulation Framework and Benchmark for Robot Learning},
  author={Zhu, Yuke and Wong, Josiah and Mandlekar, Ajay and Mart{\'i}n-Mart{\'i}n, Roberto and Joshi, Abhishek and Nasiriany, Soroush and Zhu, Yifeng},
  booktitle={arXiv preprint arXiv:2009.12293},
  year={2020}
}

@article{cetin2024simple,
  title={Simple Ingredients for Offline Reinforcement Learning},
  author={Cetin, Edoardo and Tirinzoni, Andrea and Pirotta, Matteo and Lazaric, Alessandro and Ollivier, Yann and Touati, Ahmed},
  journal={arXiv preprint arXiv:2403.13097},
  year={2024}
}

@article{kang2023improving,
  title={Improving and Benchmarking Offline Reinforcement Learning Algorithms},
  author={Kang, Bingyi and Ma, Xiao and Wang, Yirui and Yue, Yang and Yan, Shuicheng},
  journal={arXiv preprint arXiv:2306.00972},
  year={2023}
}

@article{bhargava2023when,
  title={When should we prefer Decision Transformers for Offline Reinforcement Learning?},
  author={Bhargava, Prajjwal and Chitnis, Rohan and Geramifard, Alborz and Sodhani, Shagun and Zhang, Amy},
  journal={arXiv preprint arXiv:2305.14550},
  year={2023}
}

@article{schweighofer2021dataset,
  title={A Dataset Perspective on Offline Reinforcement Learning},
  author={Schweighofer, Kajetan and Radler, Andreas and Dinu, Marius-Constantin and Hofmarcher, Markus and Patil, Vihang and Bitto-Nemling, Angela and Eghbal-zadeh, Hamid and Hochreiter, Sepp},
  journal={arXiv preprint arXiv:2111.04714},
  year={2021}
}

@article{fu2020d4rl,
  title={D4RL: Datasets for Deep Data-Driven Reinforcement Learning},
  author={Fu, Justin and Kumar, Aviral and Nachum, Ofir and Tucker, George and Levine, Sergey},
  journal={arXiv preprint arXiv:2004.07219},
  year={2020}
}

@inproceedings{agarwal2021deep,
  title={Deep Reinforcement Learning at the Edge of the Statistical Precipice},
  author={Agarwal, Rishabh and Schwarzer, Max and Castro, Pablo Samuel and Courville, Aaron and Bellemare, Marc G.},
  booktitle={Advances in Neural Information Processing Systems (NeurIPS)},
  year={2021},
  note={arXiv:2108.13264}
}

@inproceedings{chen2021decision,
  title={Decision Transformer: Reinforcement Learning via Sequence Modeling},
  author={Chen, Lili and Lu, Kevin and Rajeswaran, Aravind and Lee, Kimin and Grover, Aditya and Laskin, Michael and Abbeel, Pieter and Srinivas, Aravind and Mordatch, Igor},
  booktitle={Advances in Neural Information Processing Systems (NeurIPS)},
  year={2021}
}

@article{zhao2023learning,
  title={Learning Fine-Grained Bimanual Manipulation with Low-Cost Hardware},
  author={Zhao, Tony Z. and Kumar, Vikash and Levine, Sergey and Finn, Chelsea},
  journal={CoRR},
  volume={abs/2304.13705},
  year={2023}
}

@inproceedings{lee2024vqbet,
  title={VQ-BeT: Behavior Generation with Latent Actions},
  author={Lee, Seungjae and Wang, Yibin and Etukuru, Haritheja and Kim, H. Jin and Shafiullah, Nur Muhammad Mahi and Pinto, Lerrel},
  booktitle={International Conference on Machine Learning (ICML)},
  year={2024}
}

@article{chi2023diffusion,
  title={Diffusion Policy: Visuomotor Policy Learning via Action Diffusion},
  author={Chi, Cheng and Feng, Siyuan and Du, Yilun and Xu, Zhenjia and Cousineau, Eric and Burchfiel, Benjamin and Song, Shuran},
  journal={CoRR},
  volume={abs/2303.04137},
  year={2023}
}

@article{kostrikov2021offline,
  title={Offline Reinforcement Learning with Implicit Q-Learning},
  author={Kostrikov, Ilya and Nair, Ashvin and Levine, Sergey},
  journal={CoRR},
  volume={abs/2110.06169},
  year={2021}
}

@inproceedings{kumar2020conservative,
  title={Conservative Q-Learning for Offline Reinforcement Learning},
  author={Kumar, Aviral and Zhou, Aurick and Tucker, George and Levine, Sergey},
  booktitle={Advances in Neural Information Processing Systems (NeurIPS)},
  year={2020}
}

@article{tarasov2023revisiting,
  title={Revisiting the Minimalist Approach to Offline Reinforcement Learning},
  author={Tarasov, Denis and Kurenkov, Vladislav and Nikulin, Alexander and Kolesnikov, Sergey},
  journal={CoRR},
  volume={abs/2305.09836},
  year={2023}
}

@inproceedings{kaiser2019modelbased,
  title={Model-Based Reinforcement Learning for Atari},
  author={Kaiser, {
  \L}ukasz and Babaeizadeh, Mohammad and Mi{\l}o{\'s}, Piotr and Osi{\'n}ski, B{\l}a{\.z}ej and Campbell, Roy H. and Czechowski, Konrad and Erhan, Dumitru and Finn, Chelsea and Kozakowski, Piotr and Levine, Sergey and Mohiuddin, Afroz and Sepassi, Ryan and Tucker, George and Michalewski, Henryk},
  booktitle={International Conference on Learning Representations (ICLR)},
  year={2020},
  note={arXiv:1903.00374}
}

@article{bergstra2012random,
  title={Random Search for Hyper-Parameter Optimization},
  author={Bergstra, James and Bengio, Yoshua},
  journal={Journal of Machine Learning Research},
  volume={13},
  pages={281--305},
  year={2012}
}

@inproceedings{bergstra2011algorithms,
  title={Algorithms for Hyper-Parameter Optimization},
  author={Bergstra, James and Bardenet, R{\'e}mi and Bengio, Yoshua and K{\'e}gl, Bal{\'a}zs},
  booktitle={Advances in Neural Information Processing Systems 24 (NeurIPS)},
  pages={2546--2554},
  year={2011}
}

@inproceedings{akiba2019optuna,
  title={Optuna: A Next-generation Hyperparameter Optimization Framework},
  author={Akiba, Takuya and Sano, Shotaro and Yanase, Toshihiko and Ohta, Takeru and Koyama, Masanori},
  booktitle={Proceedings of the 25th ACM SIGKDD International Conference on Knowledge Discovery and Data Mining},
  year={2019}
}

@misc{orsini2021mattersadversarialimitationlearning,
      title={What Matters for Adversarial Imitation Learning?}, 
      author={Manu Orsini and Anton Raichuk and Léonard Hussenot and Damien Vincent and Robert Dadashi and Sertan Girgin and Matthieu Geist and Olivier Bachem and Olivier Pietquin and Marcin Andrychowicz},
      year={2021},
      eprint={2106.00672},
      archivePrefix={arXiv},
      primaryClass={cs.LG},
      url={https://arxiv.org/abs/2106.00672}, 
}

@misc{andrychowicz2020mattersonpolicyreinforcementlearning,
      title={What Matters In On-Policy Reinforcement Learning? A Large-Scale Empirical Study}, 
      author={Marcin Andrychowicz and Anton Raichuk and Piotr Stańczyk and Manu Orsini and Sertan Girgin and Raphael Marinier and Léonard Hussenot and Matthieu Geist and Olivier Pietquin and Marcin Michalski and Sylvain Gelly and Olivier Bachem},
      year={2020},
      eprint={2006.05990},
      archivePrefix={arXiv},
      primaryClass={cs.LG},
      url={https://arxiv.org/abs/2006.05990}, 
}

@inproceedings{agarwal2020optimistic,
    title     = {An Optimistic Perspective on Offline Reinforcement Learning},
    author    = {Agarwal, Rishabh and Schuurmans, Dale and Norouzi, Mohammad},
    booktitle = {Proceedings of the 37th International Conference on Machine Learning},
    series    = {Proceedings of Machine Learning Research},
    volume    = {119},
    pages     = {104--114},
    year      = {2020},
    publisher = {PMLR},
    url       = {https://proceedings.mlr.press/v119/agarwal20c.html}
}

@article{d3rlpy,
    title   = {d3rlpy: An Offline Deep Reinforcement Learning Library},
    author  = {Seno, Takuma and Imai, Michita},
    journal = {Journal of Machine Learning Research},
    volume  = {23},
    number  = {315},
    pages   = {1--20},
    year    = {2022},
    url     = {https://www.jmlr.org/papers/v23/22-0017.html}
}

@misc{paine2020hyperparameterselectionofflinereinforcement,
      title={Hyperparameter Selection for Offline Reinforcement Learning}, 
      author={Tom Le Paine and Cosmin Paduraru and Andrea Michi and Caglar Gulcehre and Konrad Zolna and Alexander Novikov and Ziyu Wang and Nando de Freitas},
      year={2020},
      eprint={2007.09055},
      archivePrefix={arXiv},
      primaryClass={cs.LG},
      url={https://arxiv.org/abs/2007.09055}, 
}

@misc{liaw2018tuneresearchplatformdistributed,
      title={Tune: A Research Platform for Distributed Model Selection and Training}, 
      author={Richard Liaw and Eric Liang and Robert Nishihara and Philipp Moritz and Joseph E. Gonzalez and Ion Stoica},
      year={2018},
      eprint={1807.05118},
      archivePrefix={arXiv},
      primaryClass={cs.LG},
      url={https://arxiv.org/abs/1807.05118}, 
}

@InProceedings{pmlr-v119-kuznetsov20a,
  title = 	 {Controlling Overestimation Bias with Truncated Mixture of Continuous Distributional Quantile Critics},
  author =       {Kuznetsov, Arsenii and Shvechikov, Pavel and Grishin, Alexander and Vetrov, Dmitry},
  booktitle = 	 {Proceedings of the 37th International Conference on Machine Learning},
  pages = 	 {5556--5566},
  year = 	 {2020},
  editor = 	 {III, Hal Daumé and Singh, Aarti},
  volume = 	 {119},
  series = 	 {Proceedings of Machine Learning Research},
  month = 	 {13--18 Jul},
  publisher =    {PMLR},
  url = 	 {https://proceedings.mlr.press/v119/kuznetsov20a.html}
}

@article{stable-baselines3,
  author  = {Antonin Raffin and Ashley Hill and Adam Gleave and Anssi Kanervisto and Maximilian Ernestus and Noah Dormann},
  title   = {Stable-Baselines3: Reliable Reinforcement Learning Implementations},
  journal = {Journal of Machine Learning Research},
  year    = {2021},
  volume  = {22},
  number  = {268},
  pages   = {1-8},
  url     = {http://jmlr.org/papers/v22/20-1364.html}
}

@article{Bellemare_2013,
   title={The Arcade Learning Environment: An Evaluation Platform for General Agents},
   volume={47},
   ISSN={1076-9757},
   url={http://dx.doi.org/10.1613/jair.3912},
   DOI={10.1613/jair.3912},
   journal={Journal of Artificial Intelligence Research},
   publisher={AI Access Foundation},
   author={Bellemare, M. G. and Naddaf, Y. and Veness, J. and Bowling, M.},
   year={2013},
   month={June}, pages={253–279} }

@misc{huang2021cleanrlhighqualitysinglefileimplementations,
      title={CleanRL: High-quality Single-file Implementations of Deep Reinforcement Learning Algorithms}, 
      author={Shengyi Huang and Rousslan Fernand Julien Dossa and Chang Ye and Jeff Braga},
      year={2021},
      eprint={2111.08819},
      archivePrefix={arXiv},
      primaryClass={cs.LG},
      url={https://arxiv.org/abs/2111.08819}, 
}

@misc{espeholt2018impalascalabledistributeddeeprl,
      title={IMPALA: Scalable Distributed Deep-RL with Importance Weighted Actor-Learner Architectures}, 
      author={Lasse Espeholt and Hubert Soyer and Remi Munos and Karen Simonyan and Volodymir Mnih and Tom Ward and Yotam Doron and Vlad Firoiu and Tim Harley and Iain Dunning and Shane Legg and Koray Kavukcuoglu},
      year={2018},
      eprint={1802.01561},
      archivePrefix={arXiv},
      primaryClass={cs.LG},
      url={https://arxiv.org/abs/1802.01561}, 
}

@misc{haarnoja2018softactorcriticoffpolicymaximum,
      title={Soft Actor-Critic: Off-Policy Maximum Entropy Deep Reinforcement Learning with a Stochastic Actor}, 
      author={Tuomas Haarnoja and Aurick Zhou and Pieter Abbeel and Sergey Levine},
      year={2018},
      eprint={1801.01290},
      archivePrefix={arXiv},
      primaryClass={cs.LG},
      url={https://arxiv.org/abs/1801.01290}, 
}

@misc{schulman2017proximalpolicyoptimizationalgorithms,
      title={Proximal Policy Optimization Algorithms}, 
      author={John Schulman and Filip Wolski and Prafulla Dhariwal and Alec Radford and Oleg Klimov},
      year={2017},
      eprint={1707.06347},
      archivePrefix={arXiv},
      primaryClass={cs.LG},
      url={https://arxiv.org/abs/1707.06347}, 
}

@misc{black2026pi0visionlanguageactionflowmodel,
      title={$\pi_0$: A Vision-Language-Action Flow Model for General Robot Control}, 
      author={Kevin Black and Noah Brown and Danny Driess and Adnan Esmail and Michael Equi and Chelsea Finn and Niccolo Fusai and Lachy Groom and Karol Hausman and Brian Ichter and Szymon Jakubczak and Tim Jones and Liyiming Ke and Sergey Levine and Adrian Li-Bell and Mohith Mothukuri and Suraj Nair and Karl Pertsch and Lucy Xiaoyang Shi and James Tanner and Quan Vuong and Anna Walling and Haohuan Wang and Ury Zhilinsky},
      year={2026},
      eprint={2410.24164},
      archivePrefix={arXiv},
      primaryClass={cs.LG},
      url={https://arxiv.org/abs/2410.24164}, 
}

\newpage
\appendix

\section{Datasets}
\label{app:datasets}

    We describe the collection process for each dataset used in our analysis. We did not create or contribute these datasets, but include these details as they may be relevant to readers. Dataset identifiers below follow the Minari registry. Our benchmark contains 114 environment--dataset pairs: 29 from the Minari release of D4RL (commit 6c51941431a72f66721e5b4ed92c7ce6f116c816), 28 from the Minari MuJoCo collection (commit 8e62dc7f7fcb4a19f8f869c65402d4bb60049117), and 57 from our larger reproduction of Minari's Atari collection, the 100,000-transition \texttt{omi-n/atari-medium} collection.

    \textbf{AntMaze.} The six AntMaze datasets require a MuJoCo Ant to navigate a maze under sparse reward. They were collected with a hierarchical controller: a Q-iteration planner selected waypoints and a goal-conditioned SAC policy supplied the low-level actions \citep{fu2020d4rl,haarnoja2018softactorcriticoffpolicymaximum}. \texttt{umaze-v1} uses a fixed start and goal, whereas \texttt{umaze-diverse-v1} varies both over a set of valid maze cells. The medium and large layouts each have a \texttt{play} dataset, built from hand-selected start--goal pairs, and a \texttt{diverse} dataset, which samples both starts and goals more broadly. These give the four datasets \texttt{medium-play-v1}, \texttt{medium-diverse-v1}, \texttt{large-play-v1}, and \texttt{large-diverse-v1}.

    \textbf{Adroit Door.} Door uses a 24-degree-of-freedom Adroit hand to unlatch and open a door. \texttt{human-v2} contains 25 teleoperated human demonstrations. \texttt{expert-v2} contains rollouts of a policy trained by demo-augmented policy gradient (DAPG) \citep{rajeswaran2018learning}. \texttt{cloned-v2} is deliberately imperfect: a behavioral-cloning policy was trained on the union of the human and expert data, and its rollouts were mixed approximately equally with the source demonstrations \citep{fu2020d4rl}.

    \textbf{Adroit Hammer.} Hammer requires the Adroit hand to pick up a hammer and drive a nail. Its \texttt{human-v2}, \texttt{expert-v2}, and \texttt{cloned-v2} datasets were collected in the same three regimes as Door: 25 human teleoperation demonstrations, DAPG-policy rollouts, and an approximately equal mixture of source demonstrations and rollouts from a behavioral clone, respectively \citep{rajeswaran2018learning,fu2020d4rl}.

    \textbf{Adroit Pen.} Pen requires reorienting a pen to a target pose. \texttt{human-v2} contains 25 human teleoperation demonstrations, \texttt{expert-v2} contains DAPG-policy rollouts, and \texttt{cloned-v2} combines the source demonstrations with rollouts of a behavioral clone trained on the human and expert data \citep{rajeswaran2018learning,fu2020d4rl}.

    \textbf{Adroit Relocate.} Relocate requires the Adroit hand to move an object to a target position. As in the other Adroit environments, \texttt{human-v2} contains 25 teleoperated demonstrations, \texttt{expert-v2} contains DAPG-policy trajectories, and \texttt{cloned-v2} mixes the source demonstrations approximately equally with trajectories from a behavioral-cloning policy \citep{rajeswaran2018learning,fu2020d4rl}.

    \textbf{Kitchen.} The three Franka Kitchen datasets contain human demonstrations collected through virtual-reality teleoperation \citep{gupta2019relay,fu2020d4rl}. Each trajectory may include interactions with several appliances or fixtures. In \texttt{complete-v2}, all four target subtasks are completed in order. \texttt{partial-v2} contains mixed-task demonstrations with sub-trajectories that complete the target subtasks in sequence, while \texttt{mixed-v2} contains broader subtask combinations in which the four target subtasks are never all completed in sequence. Unlike the locomotion datasets below, these variants therefore differ in task coverage rather than in the training checkpoint of a single behavior policy.

    \textbf{PointMaze---Open.} PointMaze controls a point mass rather than a legged robot. A hand-coded proportional--derivative controller followed waypoints produced by a Q-iteration planner toward sampled goals \citep{fu2020d4rl,minari}. \texttt{open-v2} records the sparse-reward version of the open layout, and \texttt{open-dense-v2} uses dense distance-based reward; the behavior policy is otherwise the same.

    \textbf{PointMaze---U-Maze.} \texttt{umaze-v2} and \texttt{umaze-dense-v2} were collected in the U-shaped maze with the same scripted controller and Q-iteration planner. The former uses sparse goal reward and the latter dense shaped reward.

    \textbf{PointMaze---Medium.} \texttt{medium-v2} and \texttt{medium-dense-v2} increase the size and path complexity of the maze. Both were generated by the scripted planner--controller pair; they differ only in whether the recorded reward is sparse or dense.

    \textbf{PointMaze---Large.} \texttt{large-v2} and \texttt{large-dense-v2} use the largest PointMaze layout in the benchmark. As above, a Q-iteration planner supplied waypoints to a proportional--derivative controller, and the two datasets use sparse and dense reward, respectively.

    \textbf{MuJoCo collection.} The MuJoCo datasets were generated by rolling out checkpoints of policies trained in the corresponding Gymnasium environments \citep{todorov2012mujoco,minari}. The behavior policies were implemented with Stable-Baselines3 and use SAC, PPO, or TQC \citep{stable-baselines3,haarnoja2018softactorcriticoffpolicymaximum,schulman2017proximalpolicyoptimizationalgorithms,pmlr-v119-kuznetsov20a}. The labels \texttt{simple}, \texttt{medium}, and \texttt{expert} denote increasing training budgets, not mixtures of trajectories from policies of different quality. Except where noted below, the variants for an environment are checkpoints from one training run.

    \textbf{Ant.} Ant is a quadrupedal locomotion task. \texttt{simple-v0} and \texttt{medium-v0} were collected from SAC checkpoints after 1 and 10 million training steps. \texttt{expert-v0} was generated separately from an externally trained SAC checkpoint selected as the best pretrained model.

    \textbf{HalfCheetah.} HalfCheetah asks a planar articulated agent to run forward. Its three datasets use truncated quantile critics (TQC), a distributional variant of SAC: \texttt{simple-v0}, \texttt{medium-v0}, and \texttt{expert-v0} correspond to checkpoints after 1, 6, and 25 million training steps.

    \textbf{Hopper.} Hopper is one-legged forward locomotion. The \texttt{simple-v0}, \texttt{medium-v0}, and \texttt{expert-v0} datasets were generated from SAC checkpoints after 1, 4, and 25 million training steps, respectively.

    \textbf{Walker2d.} Walker2d is planar bipedal locomotion. Its \texttt{simple-v0}, \texttt{medium-v0}, and \texttt{expert-v0} datasets use SAC checkpoints after 1.5, 6, and 25 million training steps.

    \textbf{Swimmer.} Swimmer propels a three-link body through fluid by actuating its joints. It has no \texttt{simple} variant. \texttt{medium-v0} and \texttt{expert-v0} were collected from PPO checkpoints after 300 thousand and 100 million training steps. PPO was trained with an undiscounted objective because SAC-style discounting was ineffective for this environment.

    \textbf{Reacher.} Reacher asks a two-link arm to move its end effector to a target. It has two SAC-generated datasets: \texttt{medium-v0} uses a checkpoint after 200 thousand training steps and \texttt{expert-v0} a checkpoint after 10 million steps.

    \textbf{Pusher.} Pusher requires a robotic arm to push an object toward a target. \texttt{medium-v0} and \texttt{expert-v0} were generated from SAC checkpoints after 1 and 10 million training steps; no \texttt{simple} variant is provided.

    \textbf{Humanoid.} Humanoid is high-dimensional bipedal locomotion. Its trajectories were collected from TQC policies: \texttt{simple-v0}, \texttt{medium-v0}, and \texttt{expert-v0} use checkpoints after 2, 5, and 20 million training steps.

    \textbf{HumanoidStandup.} HumanoidStandup requires a fallen humanoid to rise to an upright posture. The behavior policy was SAC trained with a reduced observation representation. \texttt{simple-v0}, \texttt{medium-v0}, and \texttt{expert-v0} were collected after 2, 10, and 50 million training steps.

    \textbf{InvertedPendulum.} InvertedPendulum requires balancing a pole on a moving cart. \texttt{medium-v0} and \texttt{expert-v0} contain rollouts of SAC checkpoints after 10 thousand and 1 million training steps, respectively.

    \textbf{InvertedDoublePendulum.} InvertedDoublePendulum extends the balancing task to two hinged links. Its \texttt{medium-v0} and \texttt{expert-v0} datasets were collected from SAC checkpoints after 50 thousand and 10 million training steps.

    \textbf{Atari.} The discrete-control portion of the benchmark contains one \texttt{expert-v0} dataset for each of 57 Atari games in the Arcade Learning Environment \citep{Bellemare_2013}: Alien, Amidar, Assault, Asterix, Asteroids, Atlantis, Bank Heist, Battle Zone, Beam Rider, Berzerk, Bowling, Boxing, Breakout, Centipede, Chopper Command, Crazy Climber, Defender, Demon Attack, Double Dunk, Enduro, Fishing Derby, Freeway, Frostbite, Gopher, Gravitar, H.E.R.O., Ice Hockey, James Bond, Kangaroo, Krull, Kung-Fu Master, Montezuma's Revenge, Ms.~Pac-Man, Name This Game, Phoenix, Pitfall!, Pong, Private Eye, Q*bert, River Raid, Road Runner, Robot Tank, Seaquest, Skiing, Solaris, Space Invaders, Star Gunner, Surround, Tennis, Time Pilot, Tutankham, Up~'n Down, Venture, Video Pinball, Wizard of Wor, Yars' Revenge, and Zaxxon. For each game, the same pretrained CleanBA PPO policy with an IMPALA-style convolutional encoder that Minari used was rolled out in the corresponding ALE environment using the original Minari data generation scripts until a target of 100,000 transitions was reached \citep{huang2021cleanrlhighqualitysinglefileimplementations,schulman2017proximalpolicyoptimizationalgorithms,espeholt2018impalascalabledistributeddeeprl}. Actions were sampled stochastically from the policy logits rather than chosen greedily, and sticky actions were disabled. The policy received resized grayscale stacks of four frames, but collection occurred before those wrappers, so the datasets retain raw image observations. The \texttt{atari-medium} label describes dataset size; all 57 datasets use the same expert-policy collection procedure and do not represent medium-quality behavior.

\section{Original Paper Baselines}
\label{app:og-baselines}

    In Figure~\ref{fig:header-figure}, we compare our trained models with the original authors' implementations, each trained across three random seeds. We use the official repositories for BCQ, CQL, DT, IQL, and ReBRAC.\footnote{Repositories: \href{https://github.com/sfujim/BCQ}{BCQ}, \href{https://github.com/aviralkumar2907/CQL}{CQL}, \href{https://github.com/kzl/decision-transformer}{DT}, \href{https://github.com/ikostrikov/implicit_q_learning}{IQL}, and \href{https://github.com/DT6A/ReBRAC}{ReBRAC}.}

    To preserve the original training implementations, we restrict our modifications to data-ingestion adapters and environment construction. For CQL, we load the Minari HDF5 data and insert its transitions into \texttt{rlkit}'s \texttt{EnvReplayBuffer}. For DT, we convert each Minari dataset to the expected pickle format. For BCQ, we reconstruct episode buffers from Minari's stored episodes and load the observations, actions, rewards, terminal indicators, and next observations into BCQ's \texttt{ReplayBuffer}. We apply the same principle to the remaining methods, converting each dataset into the format expected by the original framework while leaving the algorithm and training procedure unchanged. This avoids introducing unintended differences in preprocessing, numerical precision, or trajectory construction.

    To ensure that all baselines are evaluated consistently, we adapt each repository's evaluation procedure to match our framework: we evaluate only the final checkpoint over 100 episodes. We reconstruct the exact environment and version associated with each Minari dataset and add any omitted constructor arguments or wrappers according to Minari's data-generation scripts. Because most original implementations use the Gym API~\citep{brockman2016openaigym}, these changes are limited primarily to environment construction and evaluation, leaving the underlying policies and training procedures unchanged.

\section{Hyperparameter Sensitivity Breakdown}
\label{app:sens-breakdown}

We provide detailed heatmaps of the hyperparameter sensitivities from Section~\ref{sec:hyperparam-sens} in Figure~\ref{fig:hp-sensitivity-breakdown}.

\begin{figure*}[p]
    \centering
    \includegraphics[width=\textwidth]{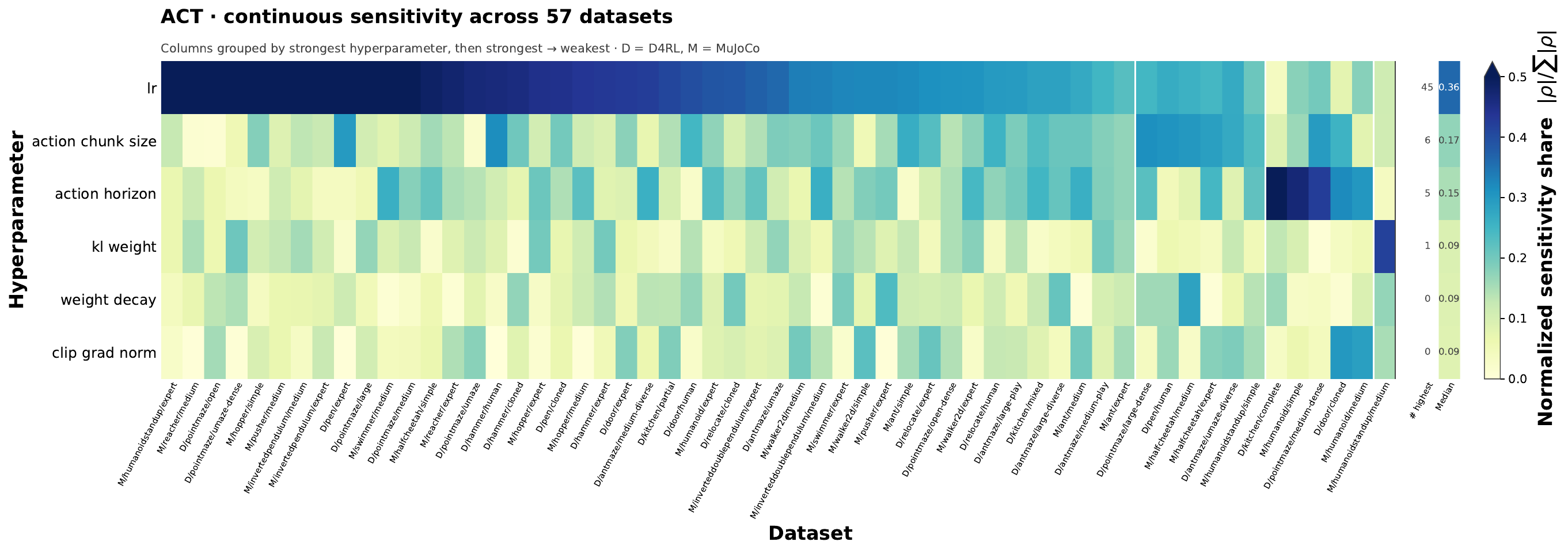}
    \includegraphics[width=\textwidth]{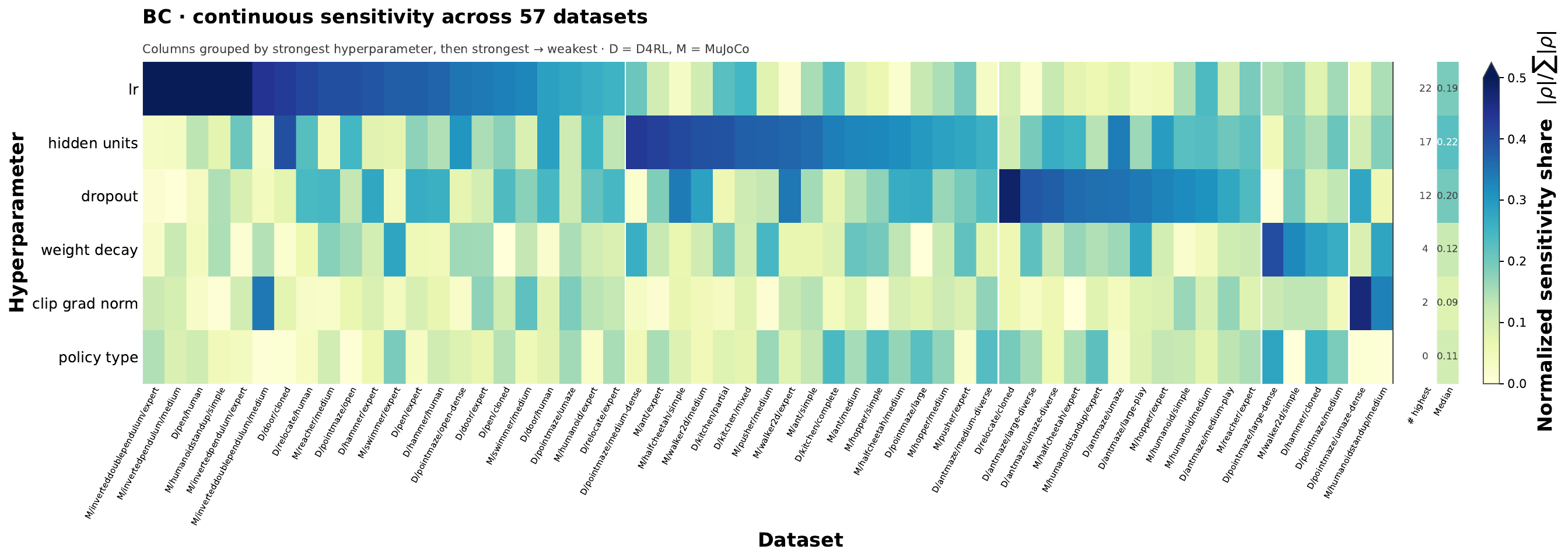}
    \caption{Per-environment hyperparameter-sensitivity breakdowns for the continuous-control algorithms. Each heatmap reports relative absolute Spearman associations between hyperparameters and trial performance. Zoom in for environment and hyperparameter labels.}
    \label{fig:hp-sensitivity-breakdown}
\end{figure*}

\begin{figure*}[p]
    \ContinuedFloat
    \centering
    \includegraphics[width=\textwidth]{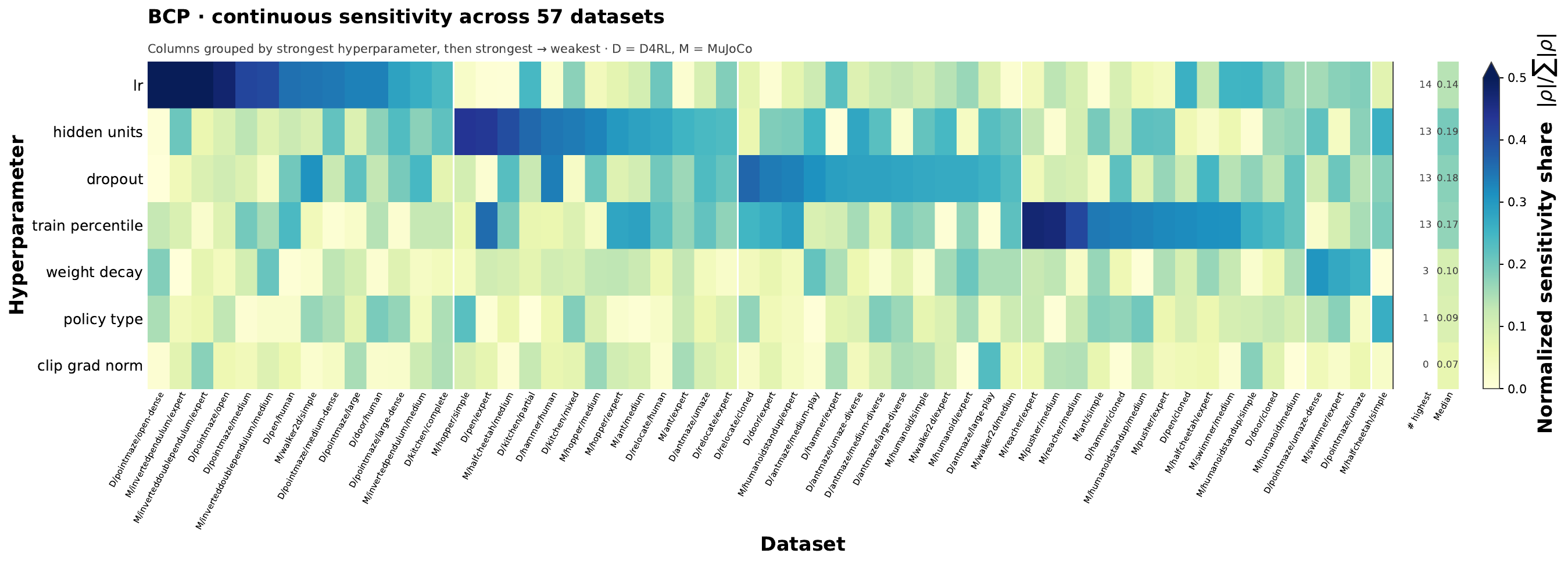}
    \includegraphics[width=\textwidth]{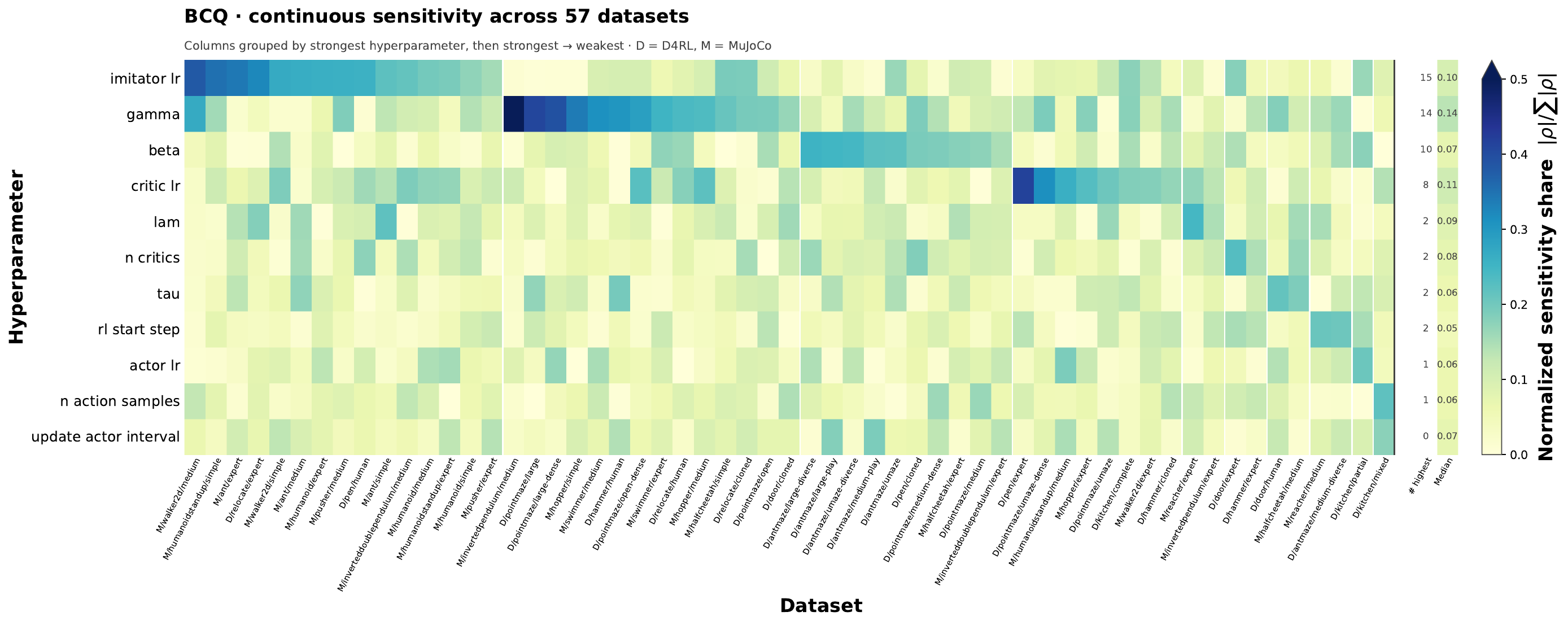}
    \caption[]{Hyperparameter-sensitivity breakdowns (continued).}
\end{figure*}

\begin{figure*}[p]
    \ContinuedFloat
    \centering
    \includegraphics[width=\textwidth]{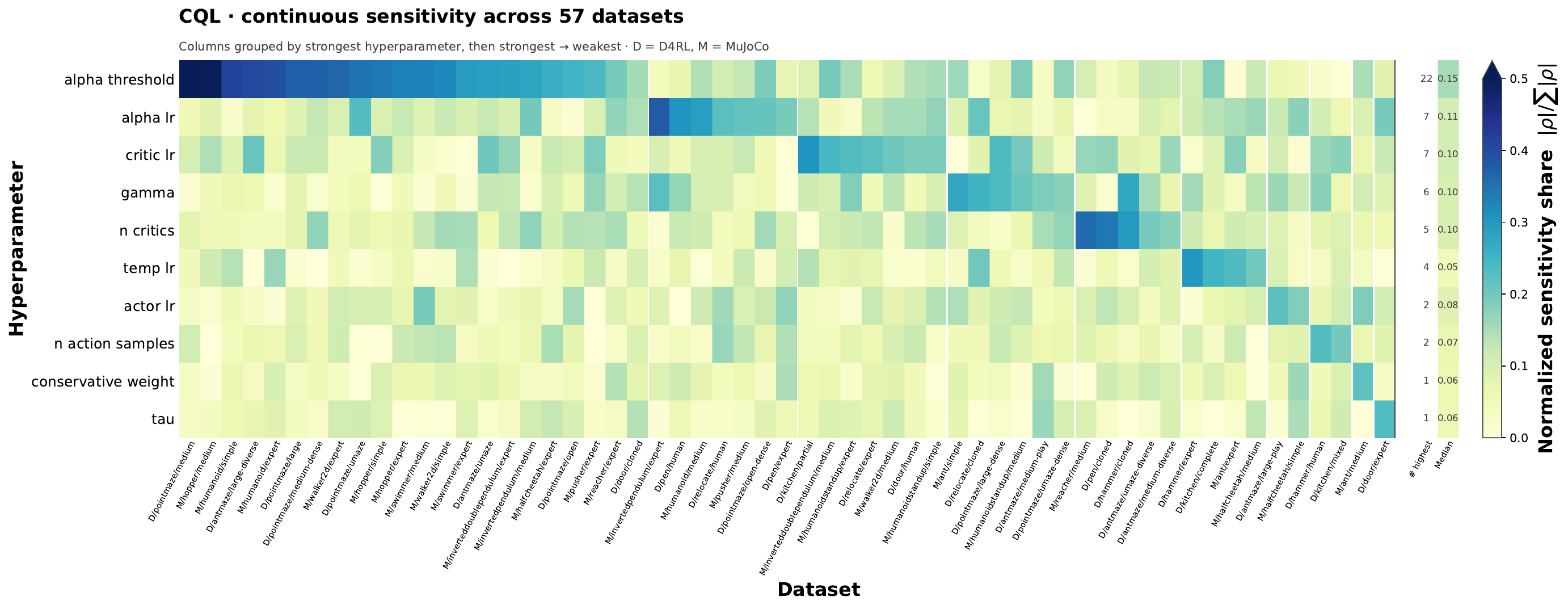}
    \includegraphics[width=\textwidth]{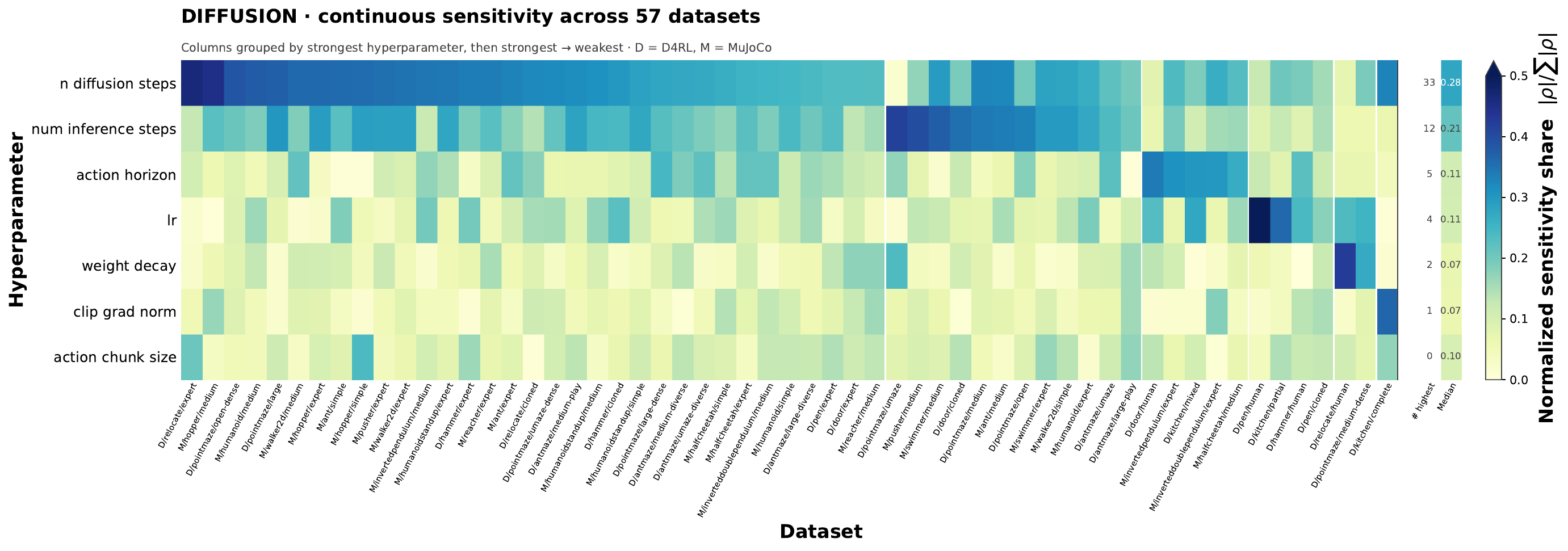}
    \caption[]{Hyperparameter-sensitivity breakdowns (continued).}
\end{figure*}

\begin{figure*}[p]
    \ContinuedFloat
    \centering
    \includegraphics[width=\textwidth]{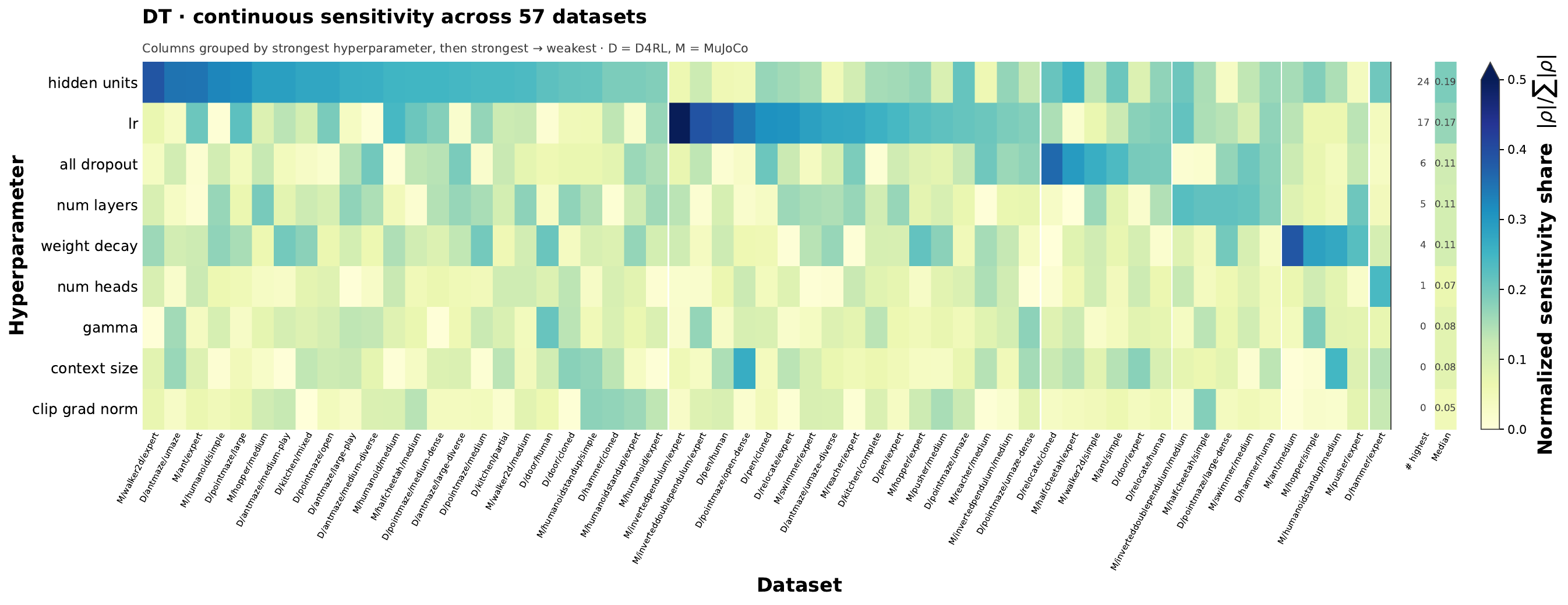}
    \includegraphics[width=\textwidth]{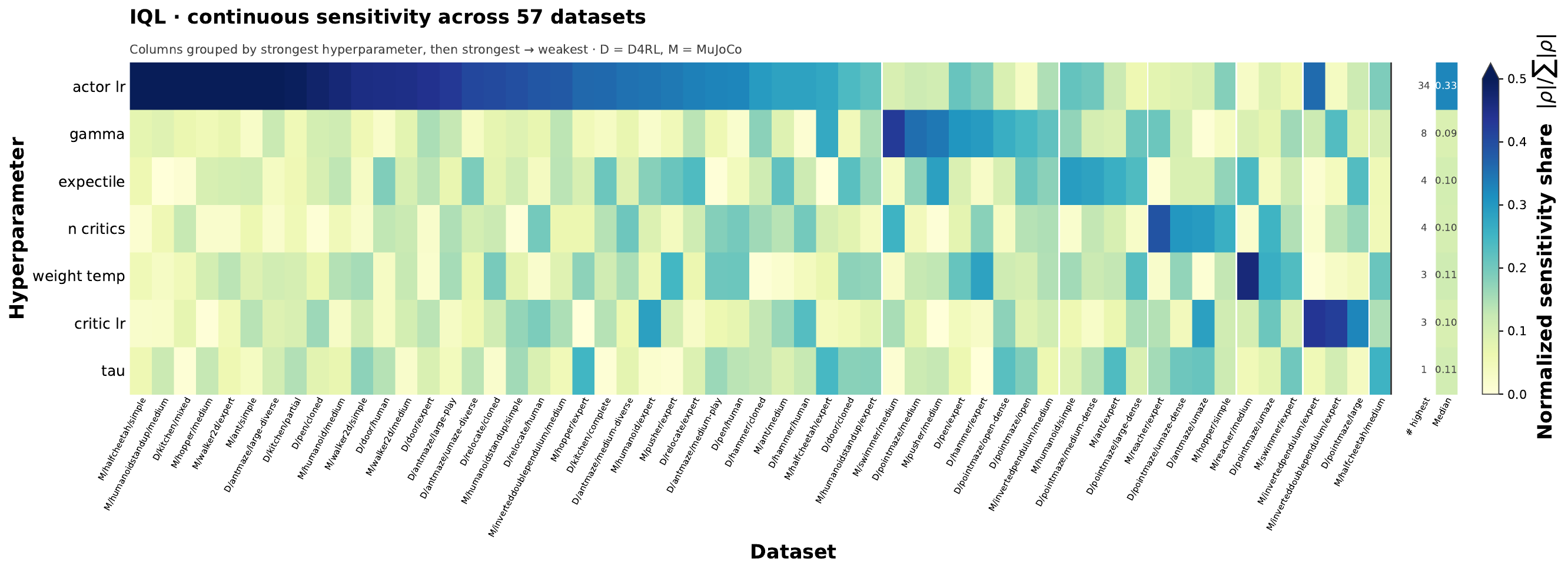}
    \caption[]{Hyperparameter-sensitivity breakdowns (continued).}
\end{figure*}

\begin{figure*}[p]
    \ContinuedFloat
    \centering
    \includegraphics[width=\textwidth]{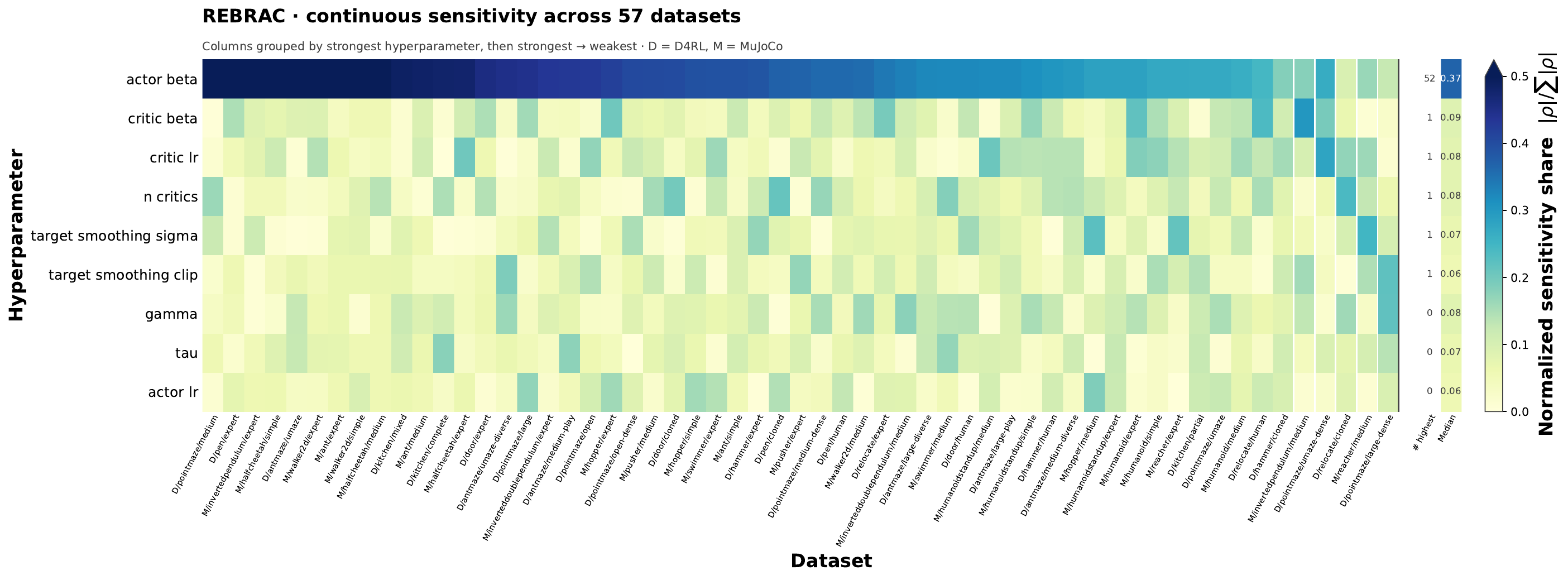}
    \includegraphics[width=\textwidth]{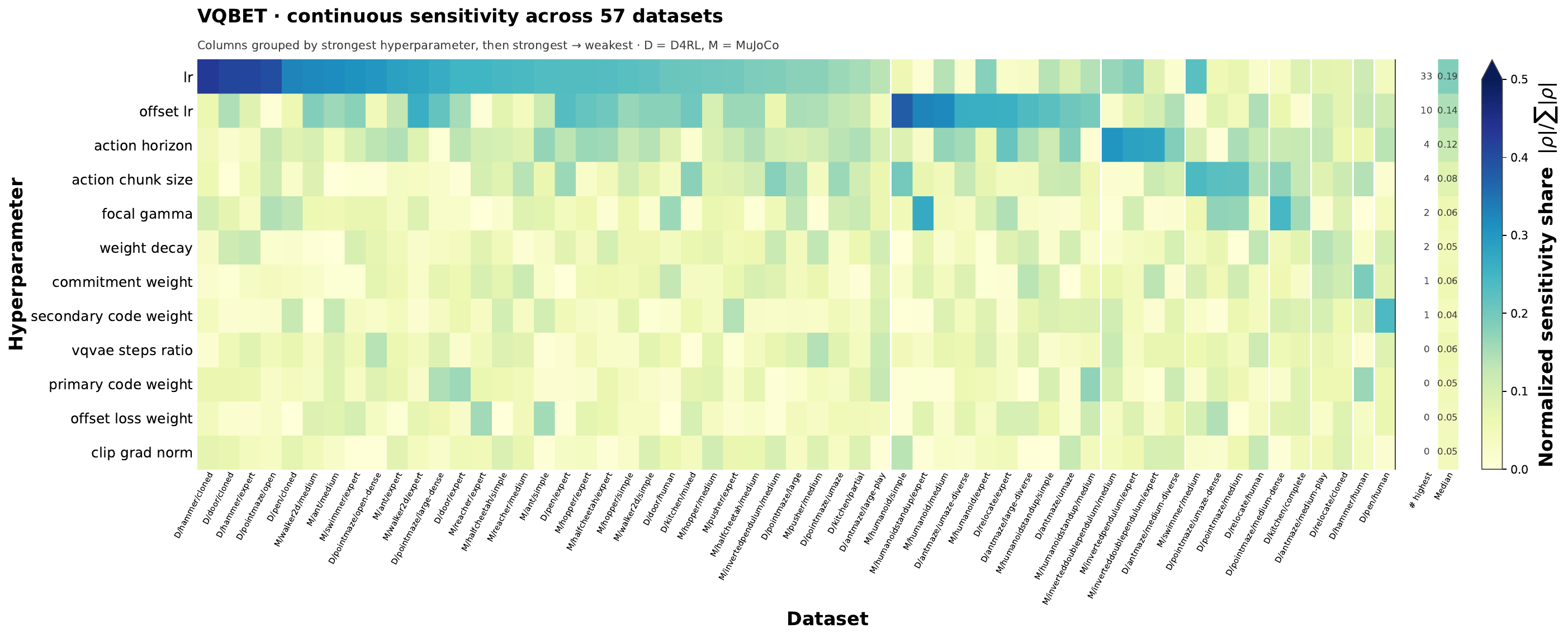}
    \caption[]{Hyperparameter-sensitivity breakdowns (continued).}
\end{figure*}

\end{document}